%% file: paper_main_review.tex
\documentclass[11pt]{article}

\usepackage[a4paper, margin=1in]{geometry}  
\usepackage[utf8]{inputenc} 
\usepackage[T1]{fontenc}    
\usepackage{hyperref}       
\usepackage{url}            
\usepackage{booktabs}       
\usepackage{amsfonts}       
\usepackage{nicefrac}       
\usepackage{microtype}      
\usepackage{xcolor}         
\usepackage{graphicx}       
\usepackage[ruled,vlined]{algorithm2e}  
\usepackage{amssymb, amsthm, amsmath, xfrac, cases}
\usepackage{empheq}
\usepackage{enumerate}
\usepackage[shortlabels]{enumitem}
\usepackage{lmodern}
\usepackage{comment}
\usepackage{subcaption}
\usepackage{bbm} 
\usepackage{graphicx}
\usepackage{booktabs} 
\usepackage{caption}
\usepackage{array}
\usepackage{amsmath}
\allowdisplaybreaks

\hypersetup{colorlinks, linkcolor={red!70!black}, citecolor={blue!70!black},
urlcolor={red!70!black} }

\theoremstyle{plain}

\theoremstyle{definition}

\theoremstyle{remark}

\usepackage{enumitem}

\usepackage[numbers,sort&compress]{natbib}
\usepackage{custom}

\input{math_commands}

\newcommand{\eps}{\varepsilon}

\title{The Rules-and-Facts Model for Simultaneous  \\ Generalization and Memorization in Neural Networks}

\author{Gabriele Farné$^1$, Fabrizio Boncoraglio$^1$, Lenka Zdeborová$^1$ }
\date{
\small
$^1$ Statistical Physics of Computation Laboratory,
\\
École Polytechnique Fédérale de Lausanne (EPFL)
CH-1015 Lausanne
}

\begin{document}
\maketitle

\begin{abstract}
A key capability of modern neural networks is their capacity to simultaneously learn underlying rules and memorize specific facts or exceptions. Yet, theoretical understanding of this dual capability remains limited. 
We introduce the Rules-and-Facts (RAF) model, a minimal solvable setting that enables precise characterization of this phenomenon by bridging two classical lines of work in the statistical physics of learning: the teacher–student framework for generalization and Gardner-style capacity analysis for memorization.
In the RAF model, a fraction 
$1 - \eps$ of training labels is generated by a structured teacher rule, while a fraction $\eps$ consists of unstructured facts with random labels.
We characterize when the learner can simultaneously recover the underlying rule—allowing generalization to new data—and memorize the unstructured examples.
Our results quantify how overparameterization enables the simultaneous realization of these two objectives: sufficient excess capacity supports memorization, while regularization and the choice of kernel or nonlinearity control the allocation of capacity between rule learning and memorization. 
Experiments on a CIFAR10-based RAF task show that the qualitative trade-off persists for fixed kernels, whereas representations learned by deeper neural networks substantially mitigate it, preserving factual recall while improving generalization.
The RAF model provides a theoretical foundation for understanding how modern neural networks can infer structure while storing rare or non-compressible information.
\end{abstract}

\section{Introduction} 
\label{sec: intro}

A hallmark of modern neural networks is their ability to simultaneously infer structured patterns and store specific exceptions or facts \cite{petroni2019language,heinzerling2021language,allen2023physics}. For example, it is desirable to learn a rule for the addition of two numbers rather than storing in the network’s weights the full list of possible sums. At the same time, certain facts—such as the name of the capital of France—likely need to be memorized. 
A language model, for instance, can generalize grammatical rules such as English verb conjugation (“jump” becomes “jumped”) while also memorizing irregular forms (“go” becomes “went”). 
Whether these behaviors require distinct rule-based and lexical mechanisms or can emerge from a single distributed system has been at the center of the long-standing ``past-tense debate'' in linguistics and cognitive science \cite{pinker2002past,mcclelland2002rules}; see also Ref.~\cite{boleda2025llms} for a recent perspective.
It is perhaps surprising that, unlike biological systems in which reasoning and memory are often attributed to distinct mechanisms or specialized brain regions, see e.g.~\cite{opitz2004brain}, modern neural networks—such as transformers—perform both functions within a single architecture. 
Understanding how and when modern neural networks are capable of generalizing and memorizing simultaneously remains a foundational open question in current machine learning, and is the main motivation for the present work. 
Modern AI systems also exhibit regimes where they produce plausible but incorrect facts, a phenomenon well documented in quantitative evaluations \cite{lin2022truthfulqa,min2023factscore,li2023halueval}; see Appendix~\ref{app:chatgpt} for an illustrative example, including a hallucinated reference to a non-existent paper by Mei, Montanari, and Nguyen.
Understanding how generalization can coexist with reliable memorization is, therefore, not only of theoretical interest but also of practical relevance.

\paragraph{Memorization vs Generalization: A Long-Standing Tension}
In classical learning theory, generalization and memorization have typically been viewed as competing or even mutually exclusive phenomena -- contributing to the long-held view that “generalization begins where memorization ends”. The classical Rademacher‑complexity bounds formalize this view by showing that models with insufficient capacity to memorize arbitrary labelings are precisely those that are guaranteed to generalize \cite{bartlett2002rademacher,abbaras2020rademacher}. 

A drastic shift in understanding came with recent efforts to explain why modern neural networks can generalize even in highly overparameterized regimes—despite also being able to fit purely random or partially randomized labels \cite{zhang2017understanding}. In that work, memorization was primarily used as a stress test for model capacity rather than as a desirable learning objective.
An influential line of work on benign overfitting subsequently established that an interpolating solution can fit all training labels, including their noisy component, without suffering poor test performance \cite{bartlett2020benign,hastie2022surprises,gerace2020generalisation,mallinar2022benign}. This does not, however, imply that the interpolating solution achieves the best generalization attainable within the same model family: positive regularization can further improve test performance, at the cost of no longer interpolating the training data. Distinguishing architectures for which interpolation is compatible with optimal generalization from those for which the two objectives remain in tension will be central to the present work.
This phenomenon is closely connected to double descent, in which test error initially worsens and then improves again as the number of model parameters increases beyond the so-called interpolation threshold \cite{belkin2019reconciling,advani2020high,spigler2019jamming}. Certain works explore this interplay even further by showing that memorization of a small subset of rare or atypical examples can, in fact, enhance generalization performance \cite{feldman2020does,zhou2025memorizing}.

Despite this shift in perspective, much of the recent literature still treats memorization as something to be tolerated -- benign at best, but not intrinsically desirable. In these works, the emphasis remains on understanding how generalization can persist despite the model’s capacity to memorize, rather than on settings where memorization is necessary for successful performance.

Separately, memorization has been studied through models such as the perceptron with random labels \cite{gardner1988optimal}, or in associative memory networks like the Hopfield model \cite{hopfield1982neural,amit1985storing}, which explicitly store patterns but do not generalize beyond them. While these classical setups have been generalized to the context of modern architectures  \cite{krotov2016dense,chung2018classification,kim2023provable,mahdavi2023memorization,nichani2024understanding}, they further reinforced the notion of separation between generalization and memorization.

Other lines of work treat memorization and generalization as fundamentally antagonistic. In privacy-sensitive applications such as medical diagnostics or text generation, memorization of specific training examples is considered a failure mode—raising concerns about data leakage and the unintended exposure of sensitive information. This has motivated extensive research on auditing, avoiding, or reversing memorization in trained models, particularly in the context of differential privacy and machine unlearning~\cite{carlini2021extracting, nasr2018machine,ginart2019making}. A perspective analogous to memorization being undesirable arises in generative modeling, where memorizing the training set leads to overfitting and degraded sample quality. For instance, diffusion models are expected to learn the underlying data distribution rather than reproduce memorized training samples, and recent works show that indeed diffusion models start to generalize once they fail to memorize~\cite{yoon2023diffusion,biroli2024dynamical,pham2025memorization}. These works further reinforce the conventional view that memorization compromises generalization and should be suppressed or mitigated.

\paragraph{Memorization and Generalization Must Go Hand in Hand}
In contrast to lines of work that treat memorization as an undesirable—or at best tolerable—side effect, our work is motivated by the observation that, in many realistic tasks, generalization and memorization are not merely compatible, but jointly essential. This is nowhere more evident than in the performance of modern large language models (LLMs), which routinely demonstrate the ability to infer grammatical and semantic structure while also recalling specific factual content. These models simultaneously solve tasks requiring abstract reasoning (e.g., code synthesis, translation, summarization) and factual precision (e.g., recalling scientific knowledge, historical dates, or rare terminology). Rather than viewing memorization as a nuisance or limitation, we treat it as an integral component of the learning objective, on equal footing with generalization. 
Our aim is thus to formulate a minimal setting in which generalization and memorization are both explicit evaluation criteria and their interplay can be analyzed in closed form.

\paragraph{The Rules-and-Facts (RAF) Task Model}

Modern AI models must both generalize patterns and memorize facts -- any complete theory of learning must account for both. To advance toward such a theory, we introduce an analytically tractable framework, which we call the \emph{Rules-and-Facts (RAF) model}, that enables the study of generalization and memorization as simultaneous and necessary objectives.
In the RAF model, each training label is either generated by a structured teacher rule or assigned at random as an isolated fact that must be memorized. A parameter \( \varepsilon \in [0, 1] \) controls the proportion of factual (non-generalizable) examples, with the remaining \( 1 - \varepsilon \) fraction adhering to the structured rule. This formulation captures the fundamental tension—and potential synergy—between generalization and memorization within a unified theoretical framework.

In order to enable analytical tractability we will consider $n$ samples in $d$ dimensions drawn from a Gaussian distribution: $\bm x^\mu\in \R^d \overset{\mathrm{i.i.d.}}{\sim} \cN(\bm 0,\mathbbm{I}_d)$. The labels of each sample are then obtained as follows  
\begin{itemize}
  \item[(i)] {\bf The rules}: with probability $1 - \eps$ a sample $\bm x^\mu$ receives the teacher label
  \begin{equation}
    y^\mu = {\rm sign}\left( \frac{\bm w_\star^\top \bm x^\mu}{\sqrt d}\right),
    \label{eq:teacher-label}
  \end{equation}
  where $\bm w_\star \in \mathbb R^d$ is the vector of teacher-weights that is the same for all the samples. The teacher weights are generated as  $\bm w_\star \overset{\mathrm{i.i.d.}}{\sim} \cN(\bm 0,\mathbbm{I}_d)$.
  \item[(ii)] {\bf The facts}: with probability $\eps$ the label is an unstructured exception $y^\mu \overset{\mathrm{i.i.d.}}{\sim} {\rm Rad}\{\tfrac12\}$ (i.e., a random ±1 label with equal probability).
  
\end{itemize}
\noindent
We consider that $n$ samples of input-label pairs $\{(\bm x^\mu,y^\mu)\}_{\mu =1}^n$ are observed, but not the teacher vector $\bm w_\star$, nor when a label comes from the rule or a fact. 

We consider the high-dimensional limit where $d,n\to \infty$ while $\alpha=\!n/d = \Theta(1)$, $\eps = \Theta(1)$.
The learner observes $n$ input-label pairs $\{ (\bm x^\mu, y^\mu)\}_{\mu=1}^n$ and its objective is to recover the underlying rule encoded by $\bm w_\star$ while also memorizing the unstructured labels. The two main parameters of the RAF data generative model are hence the fraction of facts to be memorized $\eps$ and the sample complexity $\alpha$. 

The RAF model bridges two classical paradigms in the statistical physics of learning: the teacher--student model, which describes the inference of structured rules (recovered when \( \varepsilon = 0 \))~\cite{gardner1989three,seung1992statistical}, and Gardner-style capacity analysis, which quantifies the ability to memorize arbitrary data (recovered when \( \varepsilon = 1 \))~\cite{gardner1988optimal}. 
In this sense, the RAF model combines these classical settings while changing the question asked to the learner: performance is evaluated through both generalization of the rule and recall of the facts.
Related mixture constructions, in which an $\varepsilon$
fraction of the data belongs to a distinct population, have also been
used to study outliers in high-dimensional learning~\cite{vilucchio2024asymptotic}.

We also stress upfront that our aim with this work is not to model the full complexity of real-world language or reasoning tasks, but to isolate and analyze the minimal ingredients required for a study of simultaneous generalization and memorization.

\paragraph{The analyzed neural networks}
A central question, then, is under what conditions a learner can succeed at both tasks -- learning the underlying rule and memorizing the unstructured facts -- and how this ability depends on structural properties such as model architecture or regularization. As suggested by existing results, successful memorization of arbitrary facts requires a model with sufficiently many parameters. The simplest overparameterized models that remain analytically tractable in the high-dimensional regime are the kernel regression and the related random features regression \cite{rahimi2007random}. By analyzing their performance on the  RAF model, we can quantitatively characterize the interplay between generalization (achieved by learning the rule) and memorization of the facts. We now describe the neural network models used in our theoretical analysis and define the metrics employed to evaluate both generalization and memorization performance.

 For a given loss function $\ell$ and regularization strength $\lambda > 0$, the learner model is trained on the dataset $\mathcal{D}= \{ (\boldsymbol{x}^\mu, y^\mu) \}_{\mu = 1}^n$ by minimizing the so-called empirical risk over weights $\boldsymbol{w}$,
\begin{equation} \label{eq:empirical risk}
    \mathcal{L}(\boldsymbol{w}; \mathcal{D}) = \sum_{\mu=1}^n \ell \bigl( y^\mu, f_{\boldsymbol{w}}(\boldsymbol{x}^\mu) \bigr) + \frac{\lambda}{2}  r(\boldsymbol{w} ).
\end{equation}
The function $f_{\boldsymbol{w}}$ is related to the architecture of the learner model, and $r$ is the regularization. Our theoretical analysis considers three of the most basic learners: a linear classifier, considered as a baseline, and two more expressive models, random features and kernel regression. In particular:

\begin{itemize}

   \item \textbf{Linear model aka single layer perceptron} where 
  \begin{equation}
  \label{eq:perceptron}
      f_{\boldsymbol{w}}(\boldsymbol{x}) = \frac{1}{\sqrt d} \boldsymbol{w} \cdot \boldsymbol{x}\, , \quad \quad \quad r(\boldsymbol{w} )= ||\boldsymbol{w} ||_2^2
  \end{equation}
  with $\boldsymbol{w} \in \mathbb{R}^d$.

  \item \textbf{Random feature regression} where
  \begin{equation}
  \label{eq:random-features}
      f_{\boldsymbol{w}}(\boldsymbol{x}) = \frac{1}{\sqrt p} \boldsymbol{w} \cdot \sigma \biggl( \frac{1}{\sqrt d} F^\top \boldsymbol{x} \biggr) \, , \quad \quad \quad r(\boldsymbol{w})= ||\boldsymbol{w} ||_2^2
  \end{equation}
  with $\boldsymbol{w} \in \mathbb{R}^p$, $F \in \mathbb{R}^{d \times p}$ a fixed matrix with elements $F_{ia} \overset{\mathrm{i.i.d.}}{\sim} \cN(0,1)$, and $\sigma: \mathbb{R} \to \mathbb{R}$ a non-linear activation function applied component-wise. In this work, we consider a range of possible activation functions, for example $\textrm{erf}(\cdot)$, $\textrm{sign}(\cdot)$, or $\textrm{ReLU}(\cdot)$.

\item \textbf{Kernel regression.}
For a symmetric positive semi-definite kernel
$K : \mathbb{R}^d \times \mathbb{R}^d \to \mathbb{R}$, we have
\begin{equation}
\label{eq:kernel_regression}
f_{\boldsymbol{w}}(\boldsymbol{x}) = \sum_{\nu = 1}^n w_\nu 
K(\boldsymbol{x},\boldsymbol{x}^\nu),
\qquad
r(\boldsymbol{w}) = \sum_{\mu,\nu=1}^n 
w_\mu w_\nu K(\boldsymbol{x}^\mu,\boldsymbol{x}^\nu).
\end{equation}
where $\boldsymbol{x}^\mu,\boldsymbol{x}^\nu\in \mathbb{R}^d$ are two training samples. In this work, we restrict ourselves to a class of dot-product kernels obeying a few technical assumptions described in Appendix~\ref{app:kernel_limit}.
These assumptions, however, cover many of the standard kernels used in practice.
The kernel used will be crucial for assessing the estimator's capacity to simultaneously memorize and generalize.

\end{itemize}

A classical result of \cite{rahimi2007random} establishes a relation between the kernel regression \eqref{eq:kernel_regression} and random features in the large width, $p \to \infty$, limit of \eqref{eq:random-features}. In particular, they establish that in the large width limit, random features regression with non-linearity $\sigma$ leads to kernel regression with the following kernel
  \begin{equation}
  \label{eq:kernel-reg}
      K(\boldsymbol{x}^\mu,\boldsymbol{x}^\nu) = \mathbb{E}_{\boldsymbol{f}}\!\left[ \sigma\!\left(\frac{\boldsymbol{f}^\top \boldsymbol{x}^\mu}{\sqrt d} \right)\sigma\!\left(\frac{\boldsymbol{f}^\top \boldsymbol{x}^\nu }{\sqrt d}\right) \right]
  \end{equation}
  where $\boldsymbol{x}^\mu,\boldsymbol{x}^\nu\in \mathbb{R}^d$ are two training samples and $\boldsymbol{f} \in \mathbb{R}^d$ is sampled in the same way as the columns of $F$. The converse does not hold in general: not every kernel admits a representation via the random-features construction used here. This is why we focus on the class of dot product kernels that do. We will use the relation between kernel regression and random features in their large-width limit to treat them together. From the random-features viewpoint, kernel regression corresponds to the infinite-width (hence maximally overparameterized) limit.

Regarding the loss, we will consider the two most commonly used for classification with kernel methods: 
\begin{align}
\label{eq:losses}
  \ell_\mathrm{square}(y,z) &= \tfrac12 (y-z)^2, & \ell_\mathrm{hinge}(y,z) &= \max (0,\, 1 - y z). 
\end{align}
In particular, kernel regression with square loss is often referred to as {\bf kernel ridge regression} (KRR), and kernel classification with hinge loss is called {\bf support vector machine} (SVM) \cite{cortes1995support}. \

Since we combine a convex loss with the ridge regularizers specified in Eqs.~\eqref{eq:perceptron}--\eqref{eq:kernel_regression} and take $\lambda > 0$, the resulting empirical risk \eqref{eq:empirical risk} admits a unique minimizer, which we refer to as the empirical risk minimizer, and denote by $\hat{\boldsymbol{w}}$:
\begin{equation} \label{eq: ERM estimator}
    \hat{\boldsymbol{w}}(\mathcal{D}) =
    \operatorname*{arg\,min}_{\boldsymbol{w} } \,\mathcal{L}(\boldsymbol{w}; \mathcal{D})
    = \operatorname*{arg\,min}_{\boldsymbol{w}} \left[ \, \sum_{\mu = 1}^n \ell \bigl(y^\mu, f_{\boldsymbol{w}}(\boldsymbol{x}^\mu) \bigr) + \frac{\lambda}{2} r (\boldsymbol{w} ) \right].
\end{equation}

\paragraph{Evaluation quantities}

The performance of all the models is then evaluated on two downstream tasks.
\begin{itemize} 
\item \textbf{Factual recall task}: When the learned model sees back a training sample for which the label was random, it should return back that label. By desiring good performance on this task, we model the ability to recall facts. 
\item \textbf{Teacher-rule task}: The model should be able to apply the teacher rule on previously unseen input samples. By desiring this, we model the ability to apply the learned rule in new contexts.  
\end{itemize}
We seek a model that performs well on both the factual-recall and teacher-rule tasks, i.e., that memorizes and generalizes simultaneously.

To quantify the performance on the factual recall and teacher-rule tasks, we introduce the \textit{memorization and generalization errors}.
\\
We consider the training dataset as the disjoint union of two sets, namely $\mathcal{D} = \{ (\boldsymbol{x}^\mu, y^\mu) \}_{\mu \in \mathcal{I}_\mathrm{mem}} \cup \{ (\boldsymbol{x}^\mu, y^\mu) \}_{\mu \in \mathcal{I}_\mathrm{gen}}$, where $\mathcal{I}_\mathrm{mem}$ and $\mathcal{I}_\mathrm{gen}$ denote the indices of the randomly-labeled and the learnable samples, respectively. The memorization error is thereafter defined as the average fraction of fact-labels in the dataset that are misclassified by $\hat{\boldsymbol{w}}$:
\begin{equation} \label{eq: memorization error}
    \mathcal{E}_\mathrm{mem}^{(n)} = \frac{1}{  |\mathcal{I}_\mathrm{mem}|  } \sum_{\mu \in \mathcal{I}_\mathrm{mem}} \mathbbm{1}_{y^\mu \neq \operatorname{sign} ( f_{\hat{\boldsymbol{w}}}(\boldsymbol{x}^\mu) )},
\end{equation}
where $\mathbbm{1}$ denotes the indicator function.
\\
Let $\boldsymbol{x}^\mathrm{new} \sim \mathcal{N}(\boldsymbol{0}, \mathbbm{I}_d)$ be a new input and $y^\mathrm{new}$ be the label attributed to it by the teacher rule as in Eq.~\eqref{eq:teacher-label}. The generalization error of the point estimator $\hat{\boldsymbol{w}}$ is then defined as
\begin{equation} \label{eq: generalization error}
    \mathcal{E}_\mathrm{gen}^{(n)} =\mathbb{E}_{\boldsymbol{x}^\mathrm{new}, y^\mathrm{new}} \left[ \mathbbm{1}_{ y^\mathrm{new} \neq \operatorname{sign} ( f_{\hat{\boldsymbol{w}}}(\boldsymbol{x}^\mathrm{new}) )} \right].
\end{equation}

\paragraph{Are the facts not simply label noise?}
At the level of the training data observed by the learner, the answer is yes. After marginalizing over the latent rule/fact assignment, the labels can equivalently be written as
\begin{equation}
y^\mu=\eta^\mu {\rm sign} \!\left(\frac{(\bw^\star)^\top\bx^\mu}{\sqrt d}\right),
\qquad
{P}(\eta^\mu=-1)=\frac{\varepsilon}{2},
\end{equation}
which is the standard teacher--student model with symmetric output noise \cite{engel2001statistical}. If the generalization error $\mathcal{E}_\mathrm{gen}$ were the sole evaluation criterion, the RAF model would therefore reduce to this familiar setting, and the question would become whether the ability to interpolate noisy labels is compatible with good generalization, as in the literature on benign overfitting discussed above.

The crucial difference lies in the evaluation criterion. Although the rule/fact assignment is not revealed to the learner, it is retained by the evaluator to compute the memorization error. Success therefore requires the same learned predictor to achieve both low generalization error and low memorization error. This makes the RAF task conceptually distinct from previously studied settings that do not simultaneously require both factual recall and teacher-rule generalization.

This seemingly small change in the evaluation criterion substantially changes the conclusions. All nonlinear kernels considered here can interpolate the training labels perfectly in the ridgeless limit. For a given kernel, however, the regularization that minimizes the generalization error need not be zero: positive regularization can improve rule generalization while degrading factual recall. Consequently, kernels that are equivalent when judged only by their optimally regularized generalization error can behave very differently on the simultaneous task. In particular, for kernel ridge regression, the class of kernels that attains the optimal generalization error after tuning regularization is much larger than the class that attains the same error while also perfectly memorizing the facts. Characterizing the special kernel geometries for which both objectives are realized by the same predictor is one of the central results of this work.

\paragraph{The high-dimensional limit}

In general, the performance of a neural network depends on the realization of the dataset $\mathcal{D} = \{ (\boldsymbol{x}^\mu, y^\mu) \}_{\mu=1}^n$ and the realization of the random features~$F$. However, in the present setting, the quantities of interest are self-averaging in the high-dimensional limit where
\begin{equation} 
n, d, p \to \infty\, , \quad \alpha = \frac{n}{d} = \Theta(1)\, ,  \quad \kappa = \frac{p}{d} = \Theta(1)\, . 
\label{eq:highd_limit}
\end{equation}
The parameter $\alpha$ is commonly referred to as the sample complexity of the problem, while $\kappa$ plays the role of the overparameterization ratio of the random feature architecture in Eq.~\eqref{eq:random-features}. All the other hyperparameters, such as $\eps$ or $\lambda$, are kept of order one.
The kernel regression limit is reached when $\kappa \to \infty$.

The  RAF model is inspired by two classical paradigms in the statistical physics of learning: the teacher-student model~\cite{gardner1989three,seung1992statistical} and Gardner-style capacity analysis~\cite{gardner1988optimal}. Both these models were analyzed in the high-dimensional limit for their generalization and memorization capability using the replica method.  
Random-feature regression/classification and kernel regression in high dimensions were characterized by \cite{mei2022generalization,gerace2020generalisation,loureiro2021learning,mei2022generalization_kernel} for high-dimensional teacher-student data models, but not for the specific structure of labels as in the RAF model, and without focus on the interplay between generalization and memorization.   
Methodologically we borrow from the above works to characterize the high-dimensional limit of the memorization and generalization errors as a function of the various parameters: the fraction of facts $\eps$, the sample complexity $\alpha$, the regularization strength $\lambda$, the form of the loss $\ell$, the overparameterization ratio $\kappa$, as well as the activation function $\sigma$ in the random features and the specific form of the kernel in the kernel regression. 

\paragraph{Key take-away}
A central mechanism emerging from our analysis is that, in overparameterized models, different components of the representation can be allocated to distinct objectives. In particular, the kernel geometry effectively separates a component aligned with the teacher rule from components that can be used to interpolate unstructured labels. As we will show, this decomposition — captured by the parameters $\mu_1$ and $\mu_\star$ (defined in Eq.~\eqref{eq:activations_kernel}) -- controls how capacity is distributed between generalization and memorization, and thereby enables their simultaneous realization.

This perspective provides a simple interpretation of the role of overparameterization: excess capacity is not merely used to fit arbitrary data, but can be selectively allocated so as to preserve alignment with the underlying rule while storing non-compressible information. In this sense, the coexistence of generalization and memorization is not paradoxical, but a direct consequence of how capacity is organized in overparametrized models.
The experiments of Sec.~\ref{subsec:learned_representations} suggest that feature learning can produce an analogous organization of the representation, although understanding how training constructs it lies beyond the present theory.

\section{Characterization of the performance in high-dimension} \label{sec: main equations}

In this section, we present a summary of the key equations that describe the high-dimensional performance of the learners introduced in Section \ref{sec: intro}, when trained on data generated according to the RAF model. Specifically, we characterize the memorization and generalization errors defined in Eq.~\eqref{eq: memorization error} and Eq.~\eqref{eq: generalization error} in terms of a set of low-dimensional equations.

The technical development is organized in the Appendices so as to separate the ingredients inherited from previous work from the new analysis. The random-features replica computation, which closely follows Ref.~\cite{gerace2020generalisation}, is presented in Appendix~\ref{app:replica_analysis}; its reduction in the highly overparameterized limit to the kernel equations used below is derived in Appendix~\ref{app: eqs_kernel_limit}. The resulting equations are specialized to the RAF output channel and analyzed in detail for kernel ridge regression and support vector machines in Appendices~\ref{app: KRR RAF} and~\ref{app: SVM RAF}, respectively. The Bayes-optimal benchmark is derived in Appendix~\ref{app: BO RAF}. We begin by stating this benchmark and then turn to the characterization of the ERM solution defined in Eq.~\eqref{eq: ERM estimator}.

\subsection{Bayes-optimal generalization error}
\label{sec:BO}

In the Bayes-optimal (BO) setting, we assume that the learner has access to the
true data-generating model, namely a Gaussian prior on the teacher weights, $\bw_\star \sim \cN(\bm 0,\mathbbm{I}_d)$, together with the mixed ``rules-and-facts'' output channel
\begin{equation}
P_{\rm out^\star}(y \mid z)\;=\;\varepsilon\biggl[\,\frac12\delta(y+1) + \frac12\delta(y-1) \biggr]+ (1-\varepsilon)\,\delta\bigl(y - \mathrm{sign}(z)\bigr),
\label{eq:raf-channel}
\end{equation}
where $z = \bw_\star \cdot \bx / \sqrt{d}$ and $\bx \sim \cN(\bm 0,\mathbb{I}_d)$. What the learner does not have access to is the realization of teacher weights $\bw_\star$ nor the information whether the label of a given training sample was drawn from the teacher rule or at random.

Given a new input $\bx^{\rm new}\sim\cN(\bm 0,\mathbb{I}_d)$, the posterior predictive distribution of the corresponding teacher label is
\begin{equation}
  P\bigl(y^{\rm new}\mid \bx^{\rm new},\cD\bigr)
  \;=\;
  \int \dd \bw \,
    \delta\!\left(y^{\rm new} - \mathrm{sign}\!\left(\frac{\bw \cdot \bx^{\rm new}}{\sqrt{d}}\right)\right)
    P(\bw \mid\cD),
\end{equation}
where $P(\bw \mid\cD)$ is the posterior of the RAF model. 

The Bayes-optimal classifier, which minimizes the error defined in Eq.~\eqref{eq: generalization error}, predicts the label that maximizes the posterior predictive probability, namely
\begin{equation}
  \hat y_{\rm BO}(\bx^{\rm new};\cD)
  \;=\;
  \argmax_{y\in\{-1,+1\}}
  P\bigl(y\mid \bx^{\rm new},\cD\bigr).
\label{eq:BO-predictor}
\end{equation}
The Bayes-optimal generalization error is then
\begin{equation}
  \cE_{\rm gen}^{\rm BO,(n)}
  \;=\;
  \EE\bigl[
    \mathbbm{1}\bigl\{y^{\rm new}
      \neq \hat y_{\rm BO}(\bx^{\rm new};\cD)\bigr\}
  \bigr],
  \label{eq:BO-gen-err-finiten}
\end{equation}
where the expectation is taken with respect to the joint distribution of the training set $\cD$ and the fresh test pair $(\bx^{\rm new},y^{\rm new})$.

In the limit $n,d\to\infty$ with $\alpha = n/d$ and $\varepsilon$ fixed, the generalization error in~\eqref{eq:BO-gen-err-finiten} admits a closed-form expression and is given by
\begin{equation}
  \cE_{\rm gen}^{\rm BO}
  \;=\; \lim_{n\to\infty} \cE_{\rm gen}^{\rm BO,(n)}
  \;=\; \frac{1}{\pi}\,\arccos\bigl(\sqrt{q_b}\bigr),
  \label{eq:BO-perceptron}
\end{equation}
where $q_b \in [0,1]$ is the solution of the system of two coupled scalar equations
\begin{align}
  q_b 
  &= \frac{\hat q_b}{1 + \hat q_b},
  \label{eq:bo-qb}
  \\
  \hat{q}_b
  &= \frac{4 \alpha (1-\varepsilon)^2}{(2\pi)^{3/2}(1-q_b)}
     \int_{\mathbb{R}} 
     \exp\left(
       -\frac{1+q_b}{2(1-q_b)}\,\xi^2
     \right)\,
     \frac{\dd \xi}{
       1 + (1-\varepsilon)\,
           \mathrm{erf}\!\Bigl(
             \frac{\sqrt{q_b}\,\xi}{\sqrt{2(1-q_b)}}
           \Bigr)
     }.
  \label{eq:bo-hatqb}
\end{align}
The order parameter $q_b$ is referred to as the Bayes-optimal overlap with the ground truth weights, and $\hat{q}_b\in [0,\infty)$ is its conjugate parameter. Eqs.~\eqref{eq:bo-qb}–\eqref{eq:bo-hatqb} are obtained by plugging the output channel \eqref{eq:raf-channel} into the general equations derived in \cite{aubin2020generalization}, see Appendix~\ref{App:BO-test-error} for more details. 
The Bayes-optimal error in~\eqref{eq:BO-perceptron} provides the information-theoretically best performance in learning the teacher-rule in the RAF data model, and thus gives a lower bound for the generalization error of any other estimator, including the ones we are going to discuss in the following.

For each $(\alpha,\varepsilon)$, the system~\eqref{eq:bo-qb}–\eqref{eq:bo-hatqb} admits a unique solution $(q_b,\hat q_b)$, and this solution determines the Bayes-optimal generalization error via~\eqref{eq:BO-perceptron}.  
In the limiting case $\varepsilon=0$, the equations above reduce to the ones of the Bayes-optimal generalization of the teacher-student perceptron with spherical weights studied in~\cite{seung1992statistical,JB,aubin2020generalization}, while $\varepsilon=1$ yields the random guessing estimator, with $q_b=0$ and $\cE_{\rm gen}^{\rm BO}=1/2$.

In the large sample complexity limit $\alpha \to \infty$, the Bayes-optimal generalization error decays according to the power law
\begin{equation} \label{eq:BO_rate_main}
    \mathcal{E}_\mathrm{gen}^\mathrm{BO} \underset{\alpha \to \infty}{\simeq} \frac{C_\mathrm{BO}(\varepsilon)}{\alpha},
\end{equation}
where
\begin{equation}
    C_{\rm BO}(\varepsilon) = \frac{1}{\pi J(\varepsilon)},
    \qquad
    J(\varepsilon) = \frac{4 (1 - \varepsilon)^2}{(2 \pi)^{3/2}} \int_{\mathbb{R}} \dd t \, \frac{e^{- t^2}}{ 1 + (1 - \varepsilon)\, \mathrm{erf}\!\left( \frac{t}{\sqrt{2}} \right)},
\end{equation}
for all $0 \leq \varepsilon < 1$. The derivation of Eq.~\eqref{eq:BO_rate_main} is provided in Appendix~\ref{app:decay_BO}.

\subsection{Performance of the perceptron, random features, and kernel regression}
\label{sec:ERM_equations}
In this subsection, we present the asymptotic expressions for the memorization and generalization errors, defined in Eqs.~\eqref{eq: memorization error}--\eqref{eq: generalization error}, for the three estimators considered in this work. In the high-dimensional limit, these errors are determined by a set of low-dimensional state equations. The general state equations for random-features regression at finite $\kappa$ are derived in Appendix~\ref{app:replica_analysis}. Taking their highly overparameterized limit $\kappa \to \infty$ yields the general state equations for dot-product kernel regression, as derived in Appendix~\ref{app: eqs_kernel_limit}, with the linear model recovered as a special case. Here, we report the kernel state equations already specialized to the RAF output channel. This specialization is detailed in Appendix \ref{app:loss-dep kernel SE RAF}, while the resulting equations are further analyzed for kernel ridge regression and support vector machines in Appendices~\ref{app: KRR RAF} and \ref{app: SVM RAF}, respectively.

For the RAF output channel and any convex loss $\ell$ satisfying $\ell(-y,-z) = \ell(y,z)$, including the square and hinge losses considered in this work, the generalization and memorization errors of the linear perceptron~\eqref{eq:perceptron}, random features~\eqref{eq:random-features}, and kernel method~\eqref{eq:kernel_regression}, evaluated at the unique global minimizer of Eq.~\eqref{eq: ERM estimator}, admit the following expressions in the high-dimensional limit~\eqref{eq:highd_limit}. The generalization error~\eqref{eq: generalization error} is given by:
\begin{equation}
  \cE_{\rm gen}
  \;=\; \lim_{n\to\infty} \cE_{\rm gen}^{\rm (n)}\;=\;\frac{1}{\pi}\,\arccos\!\left(\frac{m}{\sqrt{q}}\right).
  \label{eq:RF-gen-err}
\end{equation}
The memorization error~\eqref{eq: memorization error} is given by:
\begin{equation}
  \cE_{\rm mem}^{\rm }
  \;=\; \lim_{n\to\infty} \cE_{\rm mem}^{\rm (n)}
  \;=\; \frac{1}{2}\,
  \EE_{\xi\sim\cN(0,1)}
  \Bigl[
    \theta\!\left(\mathcal{P}_{V \ell(-1,\cdot)}(\sqrt{q}\xi)\right) + \theta\!\left(-\mathcal{P}_{V \ell(1,\cdot)}(\sqrt{q}\xi)\right) 
  \Bigr],
  \label{eq:mmem-error}
\end{equation}
where $\theta(\cdot)$ is the Heaviside step function. The function $\mathcal{P}_{V \ell(y,\cdot)}(\omega)$, with $y\in\{-1,+1\}$ and $\omega=\sqrt{q}\xi$ with $\xi\sim\cN(0,1)$, denotes the loss-dependent proximal operator:
\begin{equation}
\mathcal{P}_{V \ell(y,\cdot)}(\omega)
= \argmin_{z\in\R}\Bigl\{  \ell(y, z) 
   + \frac{(z-\omega)^2}{2V}\Bigr\}.
\label{eq:proximal_generic}
\end{equation}

In the case of kernel regression, the three scalar order parameters $m, q, V$ entering the errors in~\eqref{eq:RF-gen-err} and~\eqref{eq:mmem-error} are given self-consistently by the following set of six equations (see Appendices \ref{app:loss-dep kernel SE RAF} and \ref{app: kernel SE RAF}):
\begin{equation}
  m = \mu_1 \frac{\hat{m} }{\lambda + \hat{V} },
  \qquad
  q = \mu_1^2 \frac{\hat{m}^2  + \hat{q} }{(\lambda +\hat{V} )^2},
  \qquad
  V =  \frac{\mu_1^2}{\lambda + \hat{V} } + \frac{\mu_\star^2}{\lambda},
  \label{eq:rf-MQSigma-def}
\end{equation}
for $m, q, V$, and
\begin{equation}
\begin{aligned}
\label{eq:RF-saddle-kernel}
    \hat{m} &= \mu_1 \alpha (1 - \varepsilon) \sqrt{\frac{q}{2 \pi (q - m^2)}} \int_\mathbb{R} \frac{\dd \xi}{\sqrt{2\pi}} e^{- \frac{q \xi^2}{2 (q - m^2)}} \left[ f_\mathrm{out}(1, \sqrt{q} \xi, V) - f_\mathrm{out}(-1, \sqrt{q} \xi, V) \right],
    \\
    \hat{q} &= \frac{\mu_1^2 \alpha}{2} \int_\mathbb{R} \frac{\dd \xi}{\sqrt{2\pi}} e^{-\xi^2/2} \left[ f_\mathrm{out}(-1, \sqrt{q} \xi, V)^2 + f_\mathrm{out}(1, \sqrt{q} \xi, V)^2 \right]
    \\
    &\quad + \frac{\mu_1^2 \alpha (1 - \varepsilon)}{2} \int_\mathbb{R} \frac{\dd \xi}{\sqrt{2\pi}} e^{-\xi^2/2} \operatorname{erf}\!\left( \frac{m \xi}{\sqrt{2 (q - m^2)}} \right) \left[ f_\mathrm{out}(1, \sqrt{q} \xi, V)^2 - f_\mathrm{out}(-1, \sqrt{q} \xi, V)^2 \right],
    \\
    \hat{V} &= - \frac{\mu_1^2 \alpha}{2} \int_\mathbb{R} \frac{\dd \xi}{\sqrt{2\pi}} e^{-\xi^2/2} \left[ \partial_\omega f_\mathrm{out}(-1, \sqrt{q} \xi, V) + \partial_\omega f_\mathrm{out}(1, \sqrt{q} \xi, V) \right]
    \\ 
    &\quad - \frac{\mu_1^2 \alpha (1 - \varepsilon)}{2} \int_\mathbb{R} \frac{\dd \xi}{\sqrt{2\pi}} e^{-\xi^2/2} \operatorname{erf}\!\left( \frac{m \xi}{\sqrt{2 (q - m^2)}} \right) \left[ \partial_\omega f_\mathrm{out}(1, \sqrt{q} \xi, V) - \partial_\omega f_\mathrm{out}(-1, \sqrt{q} \xi, V) \right].
\end{aligned}
\end{equation}
for the conjugate variables $\hat m ,\hat q ,\hat V$. The function $f_{\rm out}(y,\omega,V)$ is commonly referred to as the output channel denoising function, and is a loss-dependent quantity defined as:
\begin{equation}
\label{eq:denoising_function}
    f_{\rm out}(y,\omega,V) =\frac{ \mathcal{P}_{V \ell(y,\cdot)}(\omega)-\omega}{V}
\end{equation}
with $\mathcal{P}_{V \ell(y,\cdot)}(\omega)$ defined in~\eqref{eq:proximal_generic}, while $\partial_\omega f_{\rm out}(y,\omega,V)$ is its partial derivative.

The kernel enters Eqs.~\eqref{eq:rf-MQSigma-def}-\eqref{eq:RF-saddle-kernel} through the two parameters $\mu_1, \mu_\star \in \mathbb{R}$. For inputs $\bx^\mu,\bx^\nu$ drawn uniformly on the sphere of radius $\sqrt{d}$, the kernel associated with the Gaussian random features learner in Eq.~\eqref{eq:kernel-reg} is a dot-product kernel, namely
\begin{equation}
K(\rho)
\;=\;
\E_{(g,g')\sim\cN(\boldsymbol{0},\Sigma(\rho))}\!\bigl[\sigma(g)\sigma(g')\bigr],
\label{eq:kernel-rho-repr}
\end{equation}
where
\begin{equation}
    \Sigma(\rho)=\begin{pmatrix}1&\rho\\ \rho&1\end{pmatrix}
\end{equation}
with $\rho$ denoting the normalized inner product of the inputs $\bx^\mu,\bx^\nu$:
\begin{equation}
    \rho = \frac{(\bx^\mu)^\top\bx^\nu}{d}.
\end{equation}
Assuming that $\sigma \in L^2(\mathcal{N}(0,1))$, substituting its Hermite expansion into Eq.~\eqref{eq:kernel-rho-repr} yields a power-series expansion of $K(\rho)$ in $\rho$, with non-negative coefficients $\{ a_\ell \}_{\ell \geq 0}$. The parameters $\mu_1$ and $\mu_\star$ can then be expressed equivalently in terms of the Hermite coefficients of $\sigma$ or the coefficients of the associated dot-product kernel. Specifically, the following identities hold (see Appendix~\ref{app:kernel_limit}):
\begin{equation}
\label{eq:activations_kernel}
\mu_1^2 = K'(0),\qquad
\mu_\star^2 = K(1)-K(0)-K'(0).
\end{equation}
For high-dimensional Gaussian independent samples, as we consider in this work, the above argument still holds and one has $\rho=O(d^{-1/2})$, while $\rho \to 1$ for coincident arguments. Consequently, the generalization and memorization errors are governed by the local behavior of the kernel around $\rho = 0$ and $\rho = 1$.

This observation highlights the central role of the two parameters $\mu_1$ and $\mu_\star$ in quantifying the trade-off between rule learning and fact memorization in the RAF model. In particular, $\mu_1$ and $\mu_\star$ admit an interpretation in terms of the Hermite expansion of the activation $\sigma$ (equivalently, the polynomial expansion of the kernel): $\mu_1$ is the first Hermite coefficient, capturing the linear component of $\sigma(\cdot)$ and thus governing rule learning and generalization, whereas $\mu_\star$ aggregates the higher-order contributions, encoding nonlinear components that allow for memorization\footnote{Note that the notation $\mu_\star$ is taken from previous works \cite{mei2022generalization} and the $\star$ notation does not reflect any relation to the way we denote the teacher weights $\bw_\star$.}.

Also, note that for $\mu_1 = 1$ and $\mu_\star = 0$, Eqs.~\eqref{eq:rf-MQSigma-def}-\eqref{eq:RF-saddle-kernel} coincide with the state equations describing the performance of the linear regression or, equivalently, the single-layer perceptron learner in~\eqref{eq:perceptron} (see also \cite{aubin2020generalization}). This is consistent with the fact that this choice of parameters corresponds to the linear kernel $K(\rho)=\rho$.

The order parameters $m$ and $q$ admit a clear interpretation in terms of the weights $\hat{\boldsymbol{w}}$ solving the ERM problem in random features regression. In this setting, they are given by
\begin{equation}
\begin{split}
    m &= \frac{\mu_1}{d} \left( \frac{F \hat{\boldsymbol{w}}}{\sqrt{p}} \right)^\top \boldsymbol{w}_\star,
    \\[4pt]
    q &= \frac{\mu_1^2}{d} \,\left\|\frac{F \hat{\boldsymbol{w}}}{\sqrt{p}}\right\|_2^2 + \frac{\mu_\star^2}{p}\,\|\hat{\boldsymbol{w}}\|_2^2
\end{split}
\end{equation}
and concentrate in the high-dimensional limit~\eqref{eq:highd_limit}. When $\kappa \to \infty$, these quantities characterize the asymptotic performance of kernel regression, as well as that of the linear model in the special case $\mu_1 = 1$, $\mu_\star = 0$.

It is worth emphasizing that the equations illustrated above depend on the kernel only through the two parameters $\mu_1$ and $\mu_\star$, and not on the full sequence of expansion coefficients $\{a_\ell\}_{\ell\ge0}$. This reduction allows us to parametrize a broad class of kernels using only these two effective parameters. In particular, for any prescribed pair $\mu_1, \mu_\star$, one can construct an admissible dot-product kernel whose expansion coefficients reproduce these values. Importantly, this construction is not unique: many different kernels share the same pair $\mu_1, \mu_\star$ and are therefore equivalent as far as our analysis is concerned. As a consequence, for the purposes of the asymptotic analysis, it is sufficient to consider an arbitrary representative within this equivalence class. For concreteness, we choose a simple truncated kernel of the form
\begin{equation}
\label{eq:quadratic-kernel}
K(\rho) \;=\; \mu_1^2 \rho \;+\; \mu_\star^2 \rho^2,
\end{equation}
for which $a_1=\mu_1^2$, $a_2= \mu_\star^2$, and all higher-order coefficients are set to zero. Other choices (e.g., involving higher-order terms) would lead to the same asymptotic predictions.

For illustrative purposes, we report the following kernels together with their corresponding parameters $\mu_1, \mu_\star$:
\begin{equation}
\label{eq:kernels-example}
\begin{array}{l@{\qquad}l@{\qquad}l}
K_{\rm erf}=\frac{2}{\pi}\arcsin\!\left(\frac{2\rho}{3}\right)\, , & \displaystyle \mu_1=\frac{2}{\sqrt{3\pi}}\, , 
& \displaystyle \mu_\star=\sqrt{\frac{2}{\pi}\arcsin\!\Bigl(\frac{2}{3}\Bigr)-\frac{4}{3\pi}},\\[10pt]
K_{\mathrm{sign}}=\frac{2}{\pi}\arcsin(\rho)\, , & \displaystyle \mu_1=\sqrt{\frac{2}{\pi}}, 
& \displaystyle \mu_\star=\sqrt{1-\frac{2}{\pi}},\\[10pt]
K_{\mathrm{ReLU}}=\frac{1}{2\pi}\Bigl(
        \sqrt{1-\rho^2}
        + (\pi - \arccos\rho)\,\rho
      \Bigr) \, , & \displaystyle \mu_1=\frac{1}{2}, 
& \displaystyle \mu_\star=\sqrt{\frac{1}{2}\Bigl(\frac{1}{2}-\frac{1}{\pi}\Bigr)}.
\end{array}
\end{equation}
A table of other examples of kernels for which our theory applies is presented in Appendix~\ref{appD:kernels}.

\paragraph{Square loss}
    For the square loss $\ell_{\rm square}(y,z)=\tfrac12(y-z)^2$, the expression for the denoising function in Eq.~\eqref{eq:denoising_function} and its derivative are given by (see Appendix \ref{app: KRR RAF}):
\begin{equation}
\label{eq:denoising-linear}
  f_{\rm out}^\mathrm{square}(y,\omega,V) =\frac{ y-\omega}{1+V} , \qquad   \partial_\omega  f_{\rm out}^\mathrm{square}(y,\omega,V)  = - \frac{1}{1+V}.
\end{equation}
This leads to a drastic simplification of the three equations for $\hat{m},\hat{q},\hat{V}$: 
\begin{equation}  
\label{eq:square-kernel-hat}
    \hat{m} = \frac{\mu_1 \alpha (1-\varepsilon)\sqrt{\frac{2}{\pi}} }{1+V},
    \qquad
    \hat{q} = \mu_1^2 \alpha \frac{1+q -2 \sqrt{\frac{2}{\pi}}  (1-\varepsilon)m  }{(1 +V)^2}, \qquad 
    \hat{V} = \frac{\mu_1 ^2 \alpha }{1+V}.
\end{equation}
Furthermore, the expression of the memorization error in Eq.~\eqref{eq:mmem-error} can also be simplified as
\begin{equation}
 \cE_{\rm mem}^\mathrm{square}
  \;=\; \frac{1}{2}\mathrm{erfc}\!\left(
      \frac{V}{\sqrt{2q}} \right),
\label{eq:mem-err-perceptron}
\end{equation}
where $\mathrm{erfc}(x)=1-\mathrm{erf}(x)$. 

For the hinge loss, the memorization error takes the same functional form. The corresponding denoising function and order parameter equations are provided in Appendix~\ref{app: SVM RAF}.

\section{Results and their implications}
\label{sec:results}

Results in this section can be reproduced using the codes provided in \url{https://github.com/SPOC-group/RAF}.

\subsection{The classical trade-off between generalization and memorization} 

\begin{figure}[!ht]
    \centering
    \includegraphics[width=\linewidth]{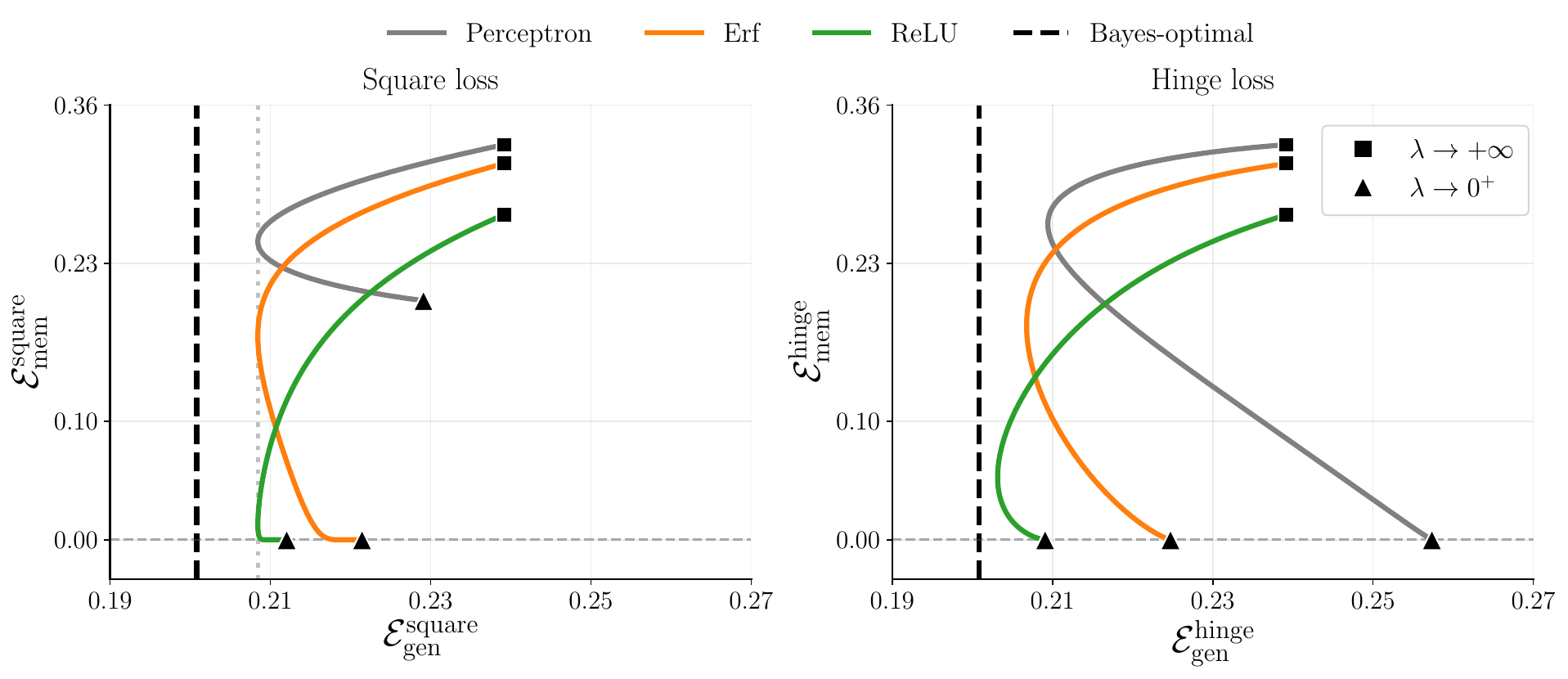}
    \caption{
    \textbf{Generalization--memorization trade-off induced by regularization for the  RAF model at fraction of facts $\varepsilon=0.1$ and sample complexity $\alpha=2.0/(1-\varepsilon)$.} Left: KRR, square loss. Right: SVM, hinge loss.
    Each plot shows the parametric curve $\lambda \mapsto (\cE_{\rm gen}(\lambda),\cE_{\rm mem}(\lambda))$ for different models: a linear perceptron and kernel regression with kernels corresponding to erf and ReLU in Eq.~\eqref{eq:kernels-example}, together with the Bayes-optimal generalization baseline $\cE_{\rm gen}^{\rm BO}=0.2008$ (bold black dashed vertical line).
    For the square loss, the optimal generalization error is the same $\cE_{\rm gen}^{\rm opt,square}=0.2084$ for all three depicted models (thin gray dotted vertical line). 
    Endpoints correspond to $\lambda\to 0^+$ (triangle) and $\lambda\to+\infty$ (square). For the hinge loss, the minimum test errors are $\mathcal{E}_{\rm gen}^{\rm opt,hinge}=0.2094$ (perceptron), $\mathcal{E}_{\rm gen}^{\rm opt,hinge}=0.2068$ (erf), $\mathcal{E}_{\rm gen}^{\rm opt, hinge}=0.2031$ (ReLU). 
    }
    \label{fig:two_panels_eps01_a02_square_hinge_BO_endpoints}
\end{figure}

We start by discussing the performance of the linear model, aka single-layer perceptron, in terms of its ability to achieve low generalization and memorization error. In Fig.~\ref{fig:two_panels_eps01_a02_square_hinge_BO_endpoints} we plot the parametric curve $\lambda \mapsto \bigl(\cE_{\rm gen}(\lambda), \cE_{\rm mem}(\lambda)\bigr)$, for fixed representative values of $(\alpha,\varepsilon)$ obtained with the square (left) and hinge (right) loss. The curve corresponding to the linear perceptron is depicted by a full gray line. We see how the regularization $\lambda$ orchestrates the trade-off between memorization and generalization. As $\lambda \to 0^+$ (marked with a black triangle), the memorization error is the lowest and gets monotonically worse as the regularization $\lambda$ increases up to its worst value at $\lambda\to \infty$ (marked with a black square). The limiting values in $\lambda$ are discussed in Appendices \ref{app: reg limits and opt lambda KRR} and \ref{app: reg limits and opt lambda SVM}. 
On the other hand, the generalization error achieves its minimum at an intermediate value of the regularization $\lambda_{\rm opt}$ and increases again beyond this point.

For the setting of Fig.~\ref{fig:two_panels_eps01_a02_square_hinge_BO_endpoints}, we see that the hinge loss provides a lower memorization error at small regularization. 
The sample complexity $\alpha$ below which, at $\lambda \to 0^+$, the memorization error is zero depends on the considered loss. For the square loss, this is below the interpolation threshold $\alpha_{\rm interp.}^{\rm square} = 1$ for all values of $0<\varepsilon \leq 1$ (see derivation in Appendix \ref{app: reg limits and opt lambda KRR}). 
For the hinge loss, the threshold below which the memorization error at $\lambda \to 0^+$ is zero is found by solving the self-consistent equation
\begin{equation}
\label{eq:interpolation-hinge-main}
    1= \alpha^{\rm hinge}_{\rm interp.}(\varepsilon) \left[ \frac{1}{2} -\frac{1-\varepsilon}{\pi} \arctan\left(\frac{\alpha^{\rm hinge}_{\rm interp.}(\varepsilon) (1-\varepsilon)}{\pi} \right)\right]
\end{equation}
as derived in Appendix \ref{app: reg limits and opt lambda SVM} for $0< \eps \leq 1$. In particular, we have $\alpha^{\rm hinge}_{\rm interp.}(\varepsilon=1)=2$ \cite{gardner1988optimal} and as $\varepsilon \to 0^+$ it diverges as $\alpha^{\rm hinge}_{\rm interp.}(\varepsilon) \simeq [ 2\pi^2 / (3\varepsilon)]^{1/3}$.
For the sample complexity in Fig.~\ref{fig:two_panels_eps01_a02_square_hinge_BO_endpoints}, we do not get zero memorization error for the square loss, but we do for the hinge loss.

Importantly, for the single-layer perceptron model, there is no value of regularization $\lambda$ where both memorization and generalization are simultaneously reasonably low. This is rather intuitive; assume that we had access to the teacher's weight vector $\boldsymbol{w}_\star$, then the memorization error would be bad, given that the teacher's labels are uncorrelated with the random labels representing facts. The single-layer perceptron presents an inevitable trade-off between generalization and memorization, exactly in line with the traditional learning theory paradigm that generalization can start where memorization ends. 
 
\subsection{The power of overparameterization}
\label{subsec:power_overparam}

More expressive models, i.e., in general those having more trainable parameters, will have a larger capacity to memorize samples. The key question is whether such memorization can be achieved without sacrificing generalization. To start investigating this question, we also plot in 
Fig.~\ref{fig:two_panels_eps01_a02_square_hinge_BO_endpoints} the parametric curve $\lambda \mapsto \bigl(\cE_{\rm gen}(\lambda), \cE_{\rm mem}(\lambda)\bigr)$ for two representative kernel regressions, one corresponding to the infinitely overparameterized random features with erf activation (in orange), the other to ReLU activation (in green). The corresponding kernels are given in Eq.~\eqref{eq:kernels-example}.  

The results show that overparameterization dramatically improves the set of simultaneously low errors $(\cE_{\rm gen},\cE_{\rm mem})$ achievable by varying the regularization $\lambda$. The corresponding curves in Fig.~\ref{fig:two_panels_eps01_a02_square_hinge_BO_endpoints} demonstrate that approaching $\lambda\to 0^+$ does not necessarily entail catastrophic loss of generalization: for a wide range of $\lambda$, kernel regression can simultaneously attain small $\cE_{\rm mem}$ and a generalization error that remains low. In other words, overparameterization creates a regime of useful (and in the  RAF model, necessary) benign overfitting: the model uses excess capacity to memorize the unstructured fraction of the data while preserving a representation aligned with the teacher rule. This observation is conceptually distinct from much of the benign-overfitting literature \cite{bartlett2020benign,mei2022generalization}, where memorization is typically treated as a by-product to be tolerated. In the RAF model, memorization is part of the task objective: the learner is expected to store the factual subset.

In Fig.~\ref{fig:two_panels_eps01_a02_square_hinge_BO_endpoints} we see that, in particular, the ReLU kernel enables regimes in which the model fits the random labels nearly perfectly while maintaining good generalization. The erf kernel does not achieve good memorization and generalization simultaneously. This motivates a follow-up question of how the generalization-memorization trade-off depends on the kernel, i.e., the two parameters $\mu_1$, $\mu_\star$, and the data, via the $\varepsilon, \alpha$ parameters. We investigate this question thoroughly in the next section. 

\begin{figure}[!ht]
  \centering
  \includegraphics[width=1.0\linewidth]{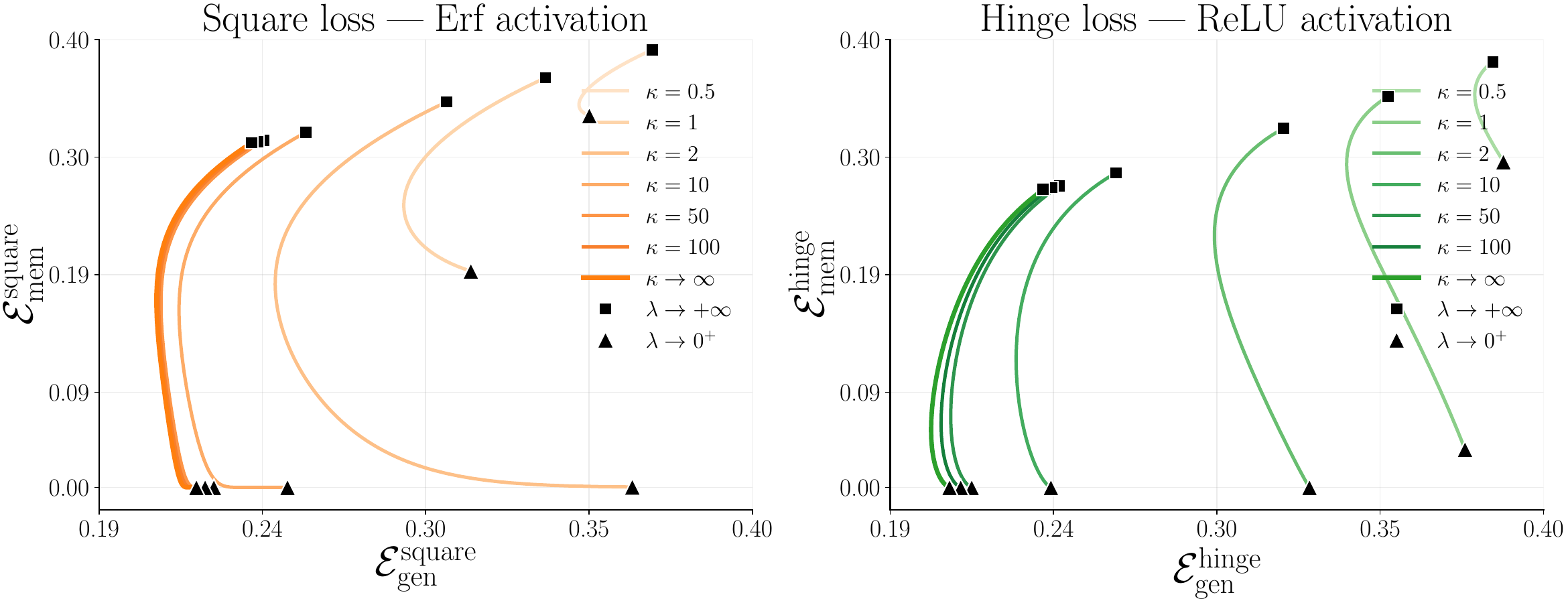}
  \caption{
  \textbf{Finite-width random features vs.\ kernel limit.}
  Parametric memorization-generalization trade-off curves $\lambda\mapsto\bigl(\cE_{\rm gen}(\lambda),\cE_{\rm mem}(\lambda)\bigr)$ for increasing model widths $\kappa$, showing convergence to the $\kappa \to \infty$ kernel prediction.
  Data model parameters: $\varepsilon = 0.1,\,  \alpha(1-\varepsilon)=2$. The endpoints are numerically obtained in the random features case. Left: Erf activation, square loss. Right: ReLU activation, hinge loss.}
  \label{fig:square_panel_stability2}
\end{figure}

Fig.~\ref{fig:square_panel_stability2} further illustrates the role of overparameterization by depicting the parametric memorization-generalization trade-off $\lambda\mapsto(\cE_{\rm gen}(\lambda),\cE_{\rm mem}(\lambda))$ for the random feature regression with finite width $\kappa=p/d$. The left panel corresponds to the Erf activation and square loss, the right one to ReLU and hinge loss. We see that as the number of learnable parameters $p=\kappa d$ increases, the capacity of the model to memorize and generalize simultaneously improves, and the curves rapidly converge to the $\kappa \to \infty$ kernel prediction. 
Perfect memorization is possible only for $\kappa$ large enough, while it is always possible in the kernel limit for any $\mu_\star > 0$ (see Appendices \ref{app: reg limits and opt lambda KRR} and \ref{app: reg limits and opt lambda SVM}).
This further justifies our choice to focus on the $\kappa \to \infty$ kernel limit and investigate the dependence on the kernel geometry fixed by $\mu_1,\mu_\star$.

\subsection{
Kernel geometry controls the rule--fact allocation}
\label{subsec:choice_nonlinearity}

We now turn to the question: \emph{which kernels are best suited for the  RAF model, and why?}
In Section \ref{sec:ERM_equations} we established that the kernel enters our high-dimensional theory only through the two scalar coefficients $\mu_1, \mu_\star$ given in Eq.~\eqref{eq:activations_kernel}, and that these are linked to the kernel abilities of rule learning and factual recalling, respectively. 

In addition, we establish analytically that, for all convex sign-symmetric losses (see Sec.~\ref{sec: main equations}) and for a class of data distributions extending beyond the RAF model, the order parameters entering the memorization and generalization errors are invariant under the reparameterization $(\mu_\star, \mu_1, \lambda) \to (r \mu_\star, r \mu_1, r^2 \lambda)$, with $r > 0$ (see Appendix~\ref{app: kernel symmetries}). Consequently, the parametric curve of memorization versus generalization obtained by varying $\lambda$, as in Fig.~\ref{fig:two_panels_eps01_a02_square_hinge_BO_endpoints}, depends on the kernel only through the ratio $\mu_1 / \mu_\star$. We therefore introduce the angle in the $(\mu_\star,\mu_1)$ plane
\begin{equation} 
\gamma = \arctan\!\left( \frac{\mu_1}{\mu_\star} \right)\, . \label{eq:angle}
\end{equation}
We identify $\gamma = \pi / 2$ with $\mu_\star = 0$; in particular, linear regression is recovered for $\mu_1 = 1$ and $\mu_\star = 0$. The angle $\gamma$ thus provides a single-parameter description of kernel geometry in the RAF model, controlling the balance between alignment with the teacher rule and the effective capacity to fit unstructured facts.

Having shown in Fig.~\ref{fig:two_panels_eps01_a02_square_hinge_BO_endpoints} the overall impact of regularization, we now focus on two particularly relevant choices:
\begin{itemize}
\item[--] \textbf{Interpolating regularization} $\lambda\to 0^+$: this is the natural choice if memorization is a hard constraint (facts must be stored). Any kernel with $\mu_\star>0$ achieves zero memorization error for any of the considered losses when $\lambda\to 0^+$, at any $(\varepsilon,\alpha)$. This result is derived in \ref{app: KRR RAF} for the square loss and in \ref{app: SVM RAF} for the hinge loss. We will be interested in the generalization error $\cE_{\rm gen}(\lambda\to 0^+)$ for this interpolating regularization. 
\item[--] \textbf{Regularization cross-validated on the generalization error} among all positive regularizations $\lambda=\lambda_{\rm opt} > 0$: this is the natural choice if rule learning is primary and one hopes memorization will come as a byproduct. As is common in the field, we consider only positive values of $\lambda$ to keep the optimization problem \eqref{eq: ERM estimator} strictly convex. We are then interested in the value of the generalization error at the optimal regularization, $\cE_{\rm gen}(\lambda_{\rm opt})$, and the value of the memorization error at that same regularization, $\cE_{\rm mem}(\lambda_{\rm opt})$.
\end{itemize}

\begin{figure}
  \centering
  \includegraphics[width=0.98\linewidth]{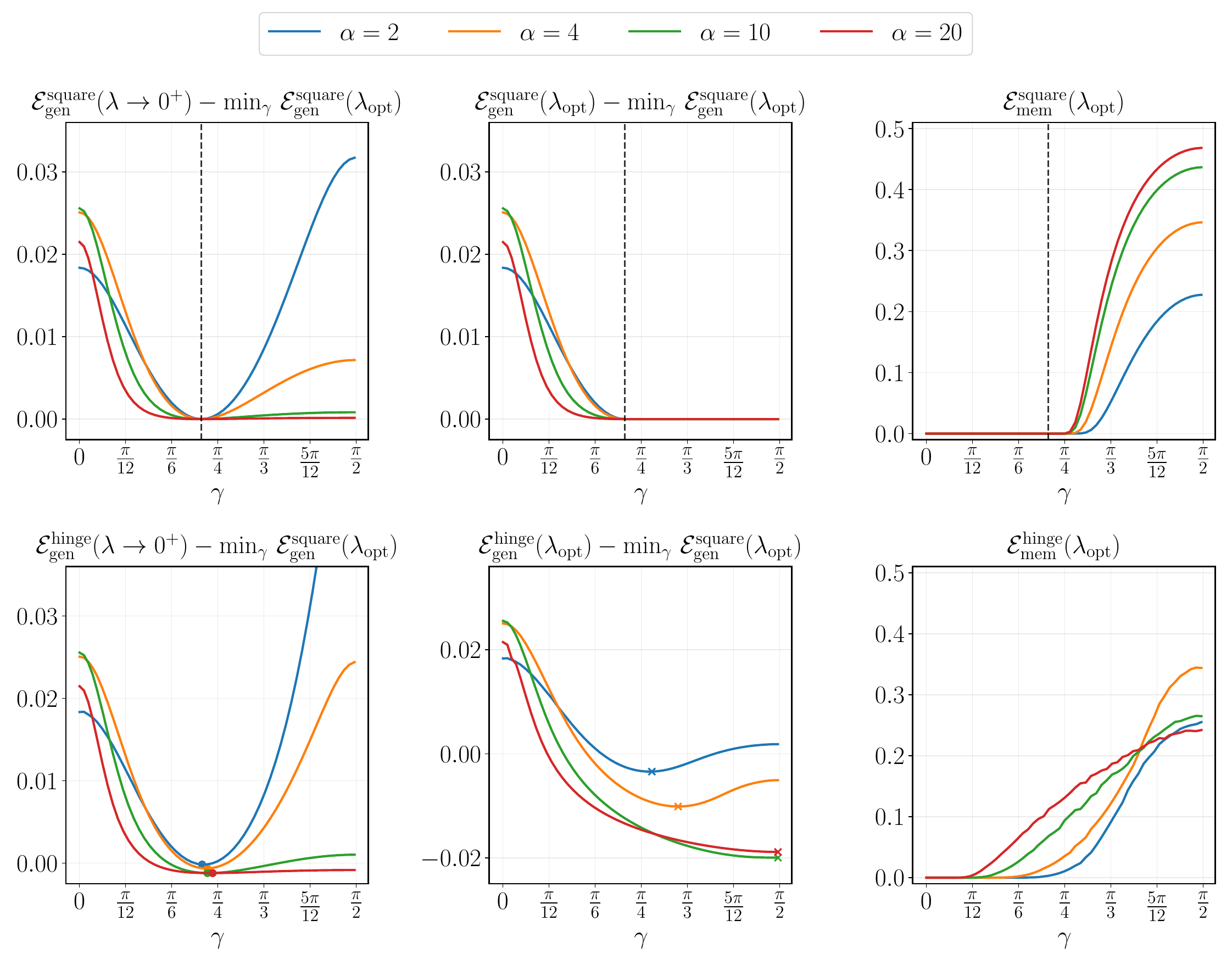}

  \vspace{0.5cm}
  
  \begin{tabular}{ccccc}
    \toprule
    $\alpha$ 
    & $\mathcal{E}_\mathrm{gen}^\mathrm{BO}$
    & $\min _\gamma \mathcal{E}_\mathrm{gen}^\mathrm{square}(\lambda_\mathrm{opt})$
    & $\min_\gamma \mathcal{E}_\mathrm{gen}^\mathrm{hinge}(\lambda_\mathrm{opt})$
    & $\min_\gamma \mathcal{E}_\mathrm{gen}^\mathrm{hinge}(\lambda \to 0^+)$ \\
    \midrule
    2  & 0.2430 & 0.2479 & 0.2445 & 0.2477 \\
    4  & 0.1702 & 0.1864 & 0.1763 & 0.1858 \\
    10 & 0.0853 & 0.1208 & 0.1009 & 0.1196 \\
    20 & 0.0449 & 0.0858 & 0.0669 & 0.0846 \\
    \bottomrule
  \end{tabular}

  \vspace{0.5cm}
  
  \caption{
  \textbf{Generalization and memorization dependence on the angle $\gamma = \arctan\!\left( \mu_1 / \mu_\star \right)$.} KRR (square loss) in the top panels, SVM (hinge loss) in the bottom panels. 
  Left panels: $\cE_{\rm gen}(\lambda \to 0^+) - \min_\gamma \cE_{\rm gen}^{\rm square}(\lambda_\mathrm{opt})$; center panels: $\cE_{\rm gen}(\lambda_{\rm opt}) - \min_\gamma \cE_{\rm gen}^{\rm square}(\lambda_\mathrm{opt})$; right panels: $\cE_{\rm mem}(\lambda_{\rm opt})$. The fraction of facts in the training set is fixed to  $\varepsilon = 0.2$. The sample complexities are  $\alpha \in \{2, 4, 10, 20\}$ in each panel.
  The dashed black vertical line in the upper panels is the optimal angle $\gamma^{\rm opt}_{\rm mem}(\varepsilon)$ in Eq.~\eqref{eq:angle_opt_mem}, for which, for the square loss, both minimum generalization and perfect memorization are simultaneously reached. 
  For the hinge loss (bottom panels), the angle where $\mathcal{E}_\mathrm{gen}^\mathrm{hinge}(\lambda_\mathrm{opt})$ is minimal is marked by a cross, while the angle where $\mathcal{E}_\mathrm{gen}^\mathrm{hinge}(\lambda\to0^+)$ is minimum is marked by a dot. 
  The table summarizes the Bayes-optimal error, the minimum generalization for the square loss (that coincides for $\lambda \to 0^+$ and $\lambda_{\rm opt}$), and the minimum generalization for the hinge loss -- both for $\lambda \to 0^+$ and $\lambda_{\rm opt}$. 
  }
  \label{fig: Figure 3}
\end{figure}

In Fig.~\ref{fig: Figure 3} we present how the generalization errors $\cE_{\rm gen}(\lambda \to 0^+)$ (left panel), $\cE_{\rm gen}(\lambda_{\rm opt})$ (middle panel) and the memorization error $\cE_{\rm mem}(\lambda_{\rm opt})$ (right panel) depend on the angle $\gamma$ defined in Eq.~\eqref{eq:angle}. The different curves correspond to different sample complexities $\alpha$. In order to make the values of the generalization errors comparable across a range of $\alpha$ we subtracted from each curve the value that corresponds to the angle $\gamma$ for which the lowest $\cE_{\rm gen}(\lambda_{\rm opt})$ is achieved for the square loss. 
The corresponding subtracted values are summarized in the Table below the figure.

\paragraph{Memorization/interpolation regime} 
Let us first focus on the behaviour of the generalization error in the perfect memorization regime, at $\lambda \to 0^+$, corresponding to left-hand side panels in Fig.~\ref{fig: Figure 3}.

For the square loss, we are able to derive analytically, see Appendix \ref{app: KRR RAF}, that at $\lambda \to 0^+$ the generalization error is minimized for kernels corresponding to the angle 
\begin{equation}
        \gamma^{\rm opt}_{\rm mem}(\varepsilon) = \arctan\!\left[ \left(\frac{\pi}{2(1-\varepsilon)^2}-1\right)^{-1/2} \right].
        \label{eq:angle_opt_mem}
\end{equation}
This angle is depicted by the vertical black dashed line in the top panels of Fig.~\ref{fig: Figure 3}. 
The values of the generalization error at this optimal angle, $\min _\gamma \mathcal{E}_\mathrm{gen}^\mathrm{square}(\lambda\to 0^+)$, are summarized in the Table below Fig.~\ref{fig: Figure 3}.
Notice that while the optimal angle $\gamma^{\rm opt}_{\rm mem}$ depends on the data through the fraction of facts $\varepsilon$, it does not depend on the sample complexity $\alpha$.

For the hinge loss in the memorization regimes $\lambda\to 0^+$, we see from Fig.~\ref{fig: Figure 3}
 and the corresponding data, that the phenomenology is qualitatively the same as for the square loss, and quantitatively extremely close. The angle at which the test error is minimized for $\lambda \to 0^+$ is numerically very close (but not equal) to $\gamma_{\rm mem}^{\rm opt}$ from Eq.~\eqref{eq:angle_opt_mem}, and the lowest achieved error $\min_\gamma \mathcal{E}_\mathrm{gen}^\mathrm{hinge}(\lambda \to 0^+)$ is numerically extremely close to those achieved with the square loss $\min_\gamma \mathcal{E}_\mathrm{gen}^\mathrm{hinge}(\lambda \to 0^+) \approx \min_\gamma \mathcal{E}_\mathrm{gen}^\mathrm{square}(\lambda \to 0^+)$, compare the values in Table below Fig.~\ref{fig: Figure 3} (3rd and 5th column).

 We thus conclude that in the memorization regime, i.e., at $\lambda \to 0^+$, of the RAF model, both for the KRR and SVM, kernels corresponding, or being very close, to the optimal memorization angle \eqref{eq:angle_opt_mem} achieve the lowest generalization error.

\paragraph{Optimal generalization} 
The next natural question that arises is whether, by relaxing the perfect memorization constraint, one can further improve the generalization error. Fig.~\ref{fig: Figure 3} shows that here the answer depends on the considered loss. For the square loss (KRR), the generalization error at $\gamma^{\rm opt}_{\rm mem}(\varepsilon)$ is already the best achievable one, while for the hinge loss (SVM) an even better generalization error can be achieved. 

For the square loss, we are able to derive explicitly the optimal regularization that minimizes the generalization error (see Appendix~\ref{app: reg limits and opt lambda KRR}):
\begin{equation}
\lambda_{\rm opt}
=
\max\!\left\{
\mu_1^2\!\left(\frac{\pi}{2(1-\varepsilon)^2}-1\right) - \mu_\star^2,\; 0^+
\right\}.
\label{eq:lambda_opt_square}
\end{equation}
Note that Eq.~\eqref{eq:angle_opt_mem} is then directly related to this value. 
This result then leads to the following properties (see Appendix~\ref{app: reg limits and opt lambda KRR} for details): 
\begin{itemize}
\item The optimal regularization $\lambda_{\rm opt}\to 0^+$ for all kernels satisfying $\gamma \le \gamma^{\rm opt}_{\rm mem}(\varepsilon)$. Consequently, the corresponding memorization error is zero in that region. For kernels with $\gamma > \gamma^{\rm opt}_{\rm mem}(\varepsilon)$, both the optimal regularization and the corresponding memorization error are strictly positive. 
\item The generalization error at the optimal regularization $\lambda_{\rm opt}$ has the same lowest possible value, $\min_\gamma \cE_{\rm gen}^{\rm square}(\lambda_\mathrm{opt})$, for all kernels satisfying $\gamma \ge \gamma^{\rm opt}_{\rm mem}(\varepsilon)$, and is strictly worse than $\min_\gamma \cE_{\rm gen}^{\rm square}(\lambda_\mathrm{opt})$ for kernels that have $\gamma < \gamma^{\rm opt}_{\rm mem}(\varepsilon)$.
\end{itemize}
It thus follows that, for the kernel ridge regression, kernels that satisfy $\gamma = \gamma^{\rm opt}_{\rm mem}(\varepsilon)$ are able to simultaneously memorize perfectly while achieving optimal rule learning. For $\gamma > \gamma^{\rm opt}_{\rm mem}(\varepsilon)$, the kernel ridge regression is excellent for extracting the rule, but at the cost of limited memorization capability. For $\gamma < \gamma^{\rm opt}_{\rm mem}(\varepsilon)$, the kernel ridge regression perfectly memorizes even at optimal regularization, but at the expense of not extracting the rule reliably. 

For the hinge loss, we do not have analytical results when it comes to the optimal regularization, yet we can perform the cross-validation numerically. From the results, depicted in the bottom panels of Fig.~\ref{fig: Figure 3}, we see that the generalization error can be further lowered by adjusting the regularization value, to the detriment of the memorization error. 
The trade-off depicted in Fig.~\ref{fig: Figure 3} leads to the conclusion that for the hinge loss there is not a value of the angle $\gamma$, among those achieving zero memorization error, that would lead to the lowest achievable generalization error at the optimal regularization, $\min_\gamma \mathcal{E}_\mathrm{gen}^\mathrm{hinge}(\lambda_\mathrm{opt})$.

\begin{figure}[!ht]
    \centering
    \includegraphics[width=\linewidth]{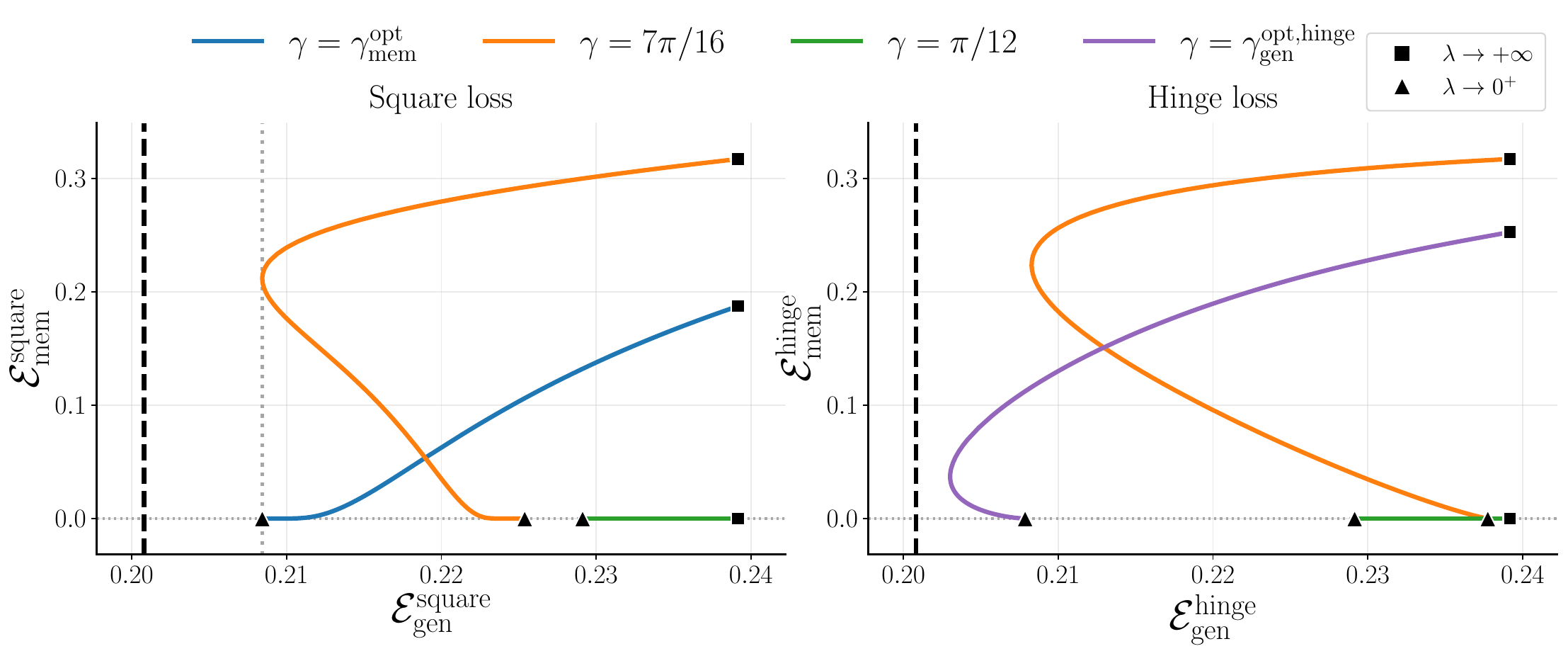}
    \caption{
    \textbf{Memorization-generalization trade-off curves $\lambda \mapsto (\mathcal{E}_{\mathrm{gen}}(\lambda),\mathcal{E}_{\mathrm{mem}}(\lambda))$ for kernel methods in the RAF model at $\varepsilon=0.1$ and $\alpha(1-\varepsilon)=2.0$.} The left panel reports square loss, while the right panel reports hinge loss. 
    Curves are shown for representative kernel geometries parameterized by the angles $\gamma$ \eqref{eq:angle}. 
For square loss (left), we display the curve for the optimal angle $\gamma_{\rm mem}^{\rm opt}= 0.8011$, in Eq.~\eqref{eq:angle_opt_mem}.    
    For hinge loss (right), we display the angle achieving optimal generalization $\gamma_{\rm gen}^{\rm opt,hinge}=0.9774$. For both losses, we also display one representative lower value of the angle, and another higher one. 
    Triangles mark the limit $\lambda\to0^{+}$, squares mark the limit $\lambda\to+\infty$, the vertical dashed line indicates the Bayes-optimal generalization baseline and the vertical dotted line indicates the optimal generalization error for the square loss.
    }
    \label{fig:two_panels_kernels}
\end{figure}

With the insights we just described, we return to the memorization-generalization trade-off curves depicted in Fig.~\ref{fig:two_panels_eps01_a02_square_hinge_BO_endpoints}. In Fig.~\ref{fig:two_panels_kernels} we depict the trade-off curves for the optimal angle $\gamma^{\rm opt}_{\rm mem}$ \eqref{eq:angle_opt_mem} for the square loss (left panel), whereas for the hinge loss (right panel) we depict the angle $\gamma_{\rm gen}^{\rm opt, hinge}$ that minimized the generalization error. For both losses, we then show one representative large value of the angle and a smaller one. 

\subsection{The large-\texorpdfstring{$\alpha$}{alpha} generalization rate}
\label{subsec:cost_memorization}
The analytical expressions for the square loss from Eqs.~\eqref{eq:rf-MQSigma-def} and \eqref{eq:square-kernel-hat} allow us to derive the decay rate of the generalization error when the fraction of facts $\varepsilon$ is fixed and the sample complexity $\alpha \to \infty$. As a result, for kernel ridge regression with $0\le\varepsilon<1$, the leading large-$\alpha$ decay of the generalization error is
\begin{equation}
\label{eq:square_large_alpha_rate_results}
\cE_{\rm gen}^\mathrm{square}
\underset{\alpha \to +\infty}{\simeq} \frac{\sqrt{\pi - 2 (1 - \varepsilon)^2}}{\sqrt{2} \pi (1 - \varepsilon)} \, \alpha^{- 1 / 2}.
\end{equation}
In particular, the leading $\alpha^{-1/2}$ coefficient is universal in the sense that it does not depend on $\lambda$ nor on the kernel geometry $\mu_1, \mu_\star$. This is shown in Appendix~\ref{app:kernel_large_alpha}.

This result has several consequences. On the positive side, the generalization rate $\alpha^{-1/2}$ is achieved even for $\lambda\to0^+$. Thus, for all the considered kernel methods with $\mu_\star>0$, generalization at rate $\alpha^{-1/2}$ is achievable together with perfect memorization. 
On the negative side, at fixed $\varepsilon>0$, with the square loss, no choice of kernel or regularization can beat the $\alpha^{-1/2}$ rate despite the fact that the Bayes-optimal rate is $\alpha^{-1}$ (presented in section \ref{sec:BO}).

In the fact-free case $\varepsilon=0$, where only the teacher rule \eqref{eq:teacher-label} produces labels, studied extensively in \cite{aubin2020generalization}, it is well known that a linear student model with the hinge loss is able to match the Bayes-optimal rate $\alpha^{-1}$. Interested in the question of whether such a fast rate can be matched for $\varepsilon>0$, we next investigate the support vector machine.

\begin{figure}[!ht]
    \centering
    \includegraphics[width=0.5\linewidth]{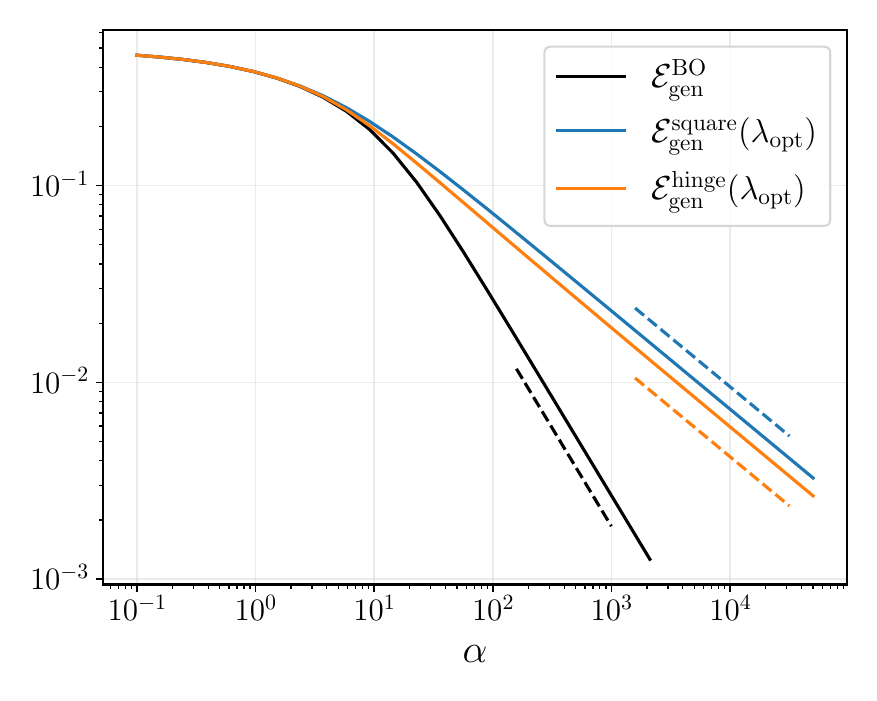}
\caption{The figure displays generalization errors $\mathcal{E}_\mathrm{gen}^\mathrm{BO}$ and $\mathcal{E}_\mathrm{gen}^\mathrm{square}(\lambda_\mathrm{opt})$, $\mathcal{E}_\mathrm{gen}^\mathrm{hinge}(\lambda_\mathrm{opt})$ for $\gamma = \arctan(\mu_1 / \mu_\star) = \pi / 4$ against $\alpha$ in log-log scale. The fraction of facts in the training set is fixed to a representative value, which is $\varepsilon = 0.5$. The dashed black segment has a slope equal to $-1$, while the other two segments have slopes equal to $-1/2$. This figure illustrates the suboptimality of the generalization rate of the kernel methods in the RAF model.}
\label{fig: Figure 6}
\end{figure}

For the hinge loss, the closed-form simplification leading to Eq.~\eqref{eq:lambda_opt_square} is no longer available. We thus resort to numerical cross-validation of the high-dimensional equations. Numerical evidence presented in Fig.~\ref{fig: Figure 6} and  Appendix~\ref{app:kernel_SVM_large_alpha} convinced us that the large-$\alpha$ generalization decay remains very close to $\alpha^{-1/2}$ independently of the kernel geometry parameters $\mu_1,\mu_\star$ and $\varepsilon$. 

Based on the above evidence, we thus conjecture that also for the hinge loss for any $\varepsilon>0$ and any regularization $\lambda$ and kernel given by $\mu_1,\mu_\star$, the generalization decay rate is $\alpha^{-1/2}$. If we were considering generalization irrespective of the desire to memorize the facts, then we could apply a construction of a loss for which gradient descent matches the steps of the Bayes-optimal approximate message passing algorithm, as detailed in \cite{advani2016equivalence,aubin2020generalization}, which would achieve the Bayes-optimal rate. Such a method would, however, not have the capacity to memorize the facts. This opens the following intriguing question: 
\begin{center}
Does there exist a neural network that is able to reach a generalization rate of $\alpha^{-1}$ on data drawn from the RAF model and at the same time memorize the facts? 
\end{center}
Our work indicates that linear and kernel methods are insufficient for that purpose. It is possible that a wide two-layer neural network with a trainable first layer (as opposed to fixed features, as we considered in the theoretical part of this work) will achieve this goal. However, an analysis of such a neural network learning on the RAF data remains a technically open problem, which we leave for future investigation. 

\section{Towards structured data and learned representations}
\label{sec:real_data_representations}   

The analysis above deliberately combines isotropic Gaussian inputs and a linear rule with analytically tractable linear, random-feature, and kernel estimators in order to isolate the mechanisms governing simultaneous memorization and generalization. We now examine how this picture changes under two steps towards modern learning settings: first, when the inputs and the rule are constructed from real data, and second, when the representation itself is learned by a neural network. Our aim here is not to obtain quantitative predictions for these substantially more complex settings, but first to ask whether the qualitative trade-off persists on structured data for fixed representations, and second to determine how it changes when the representation itself is learned.

\subsection{Qualitative comparison on structured data}
\label{main:Cifar}
\begin{figure}[!th]
    \centering
    \includegraphics[width=0.9\linewidth]{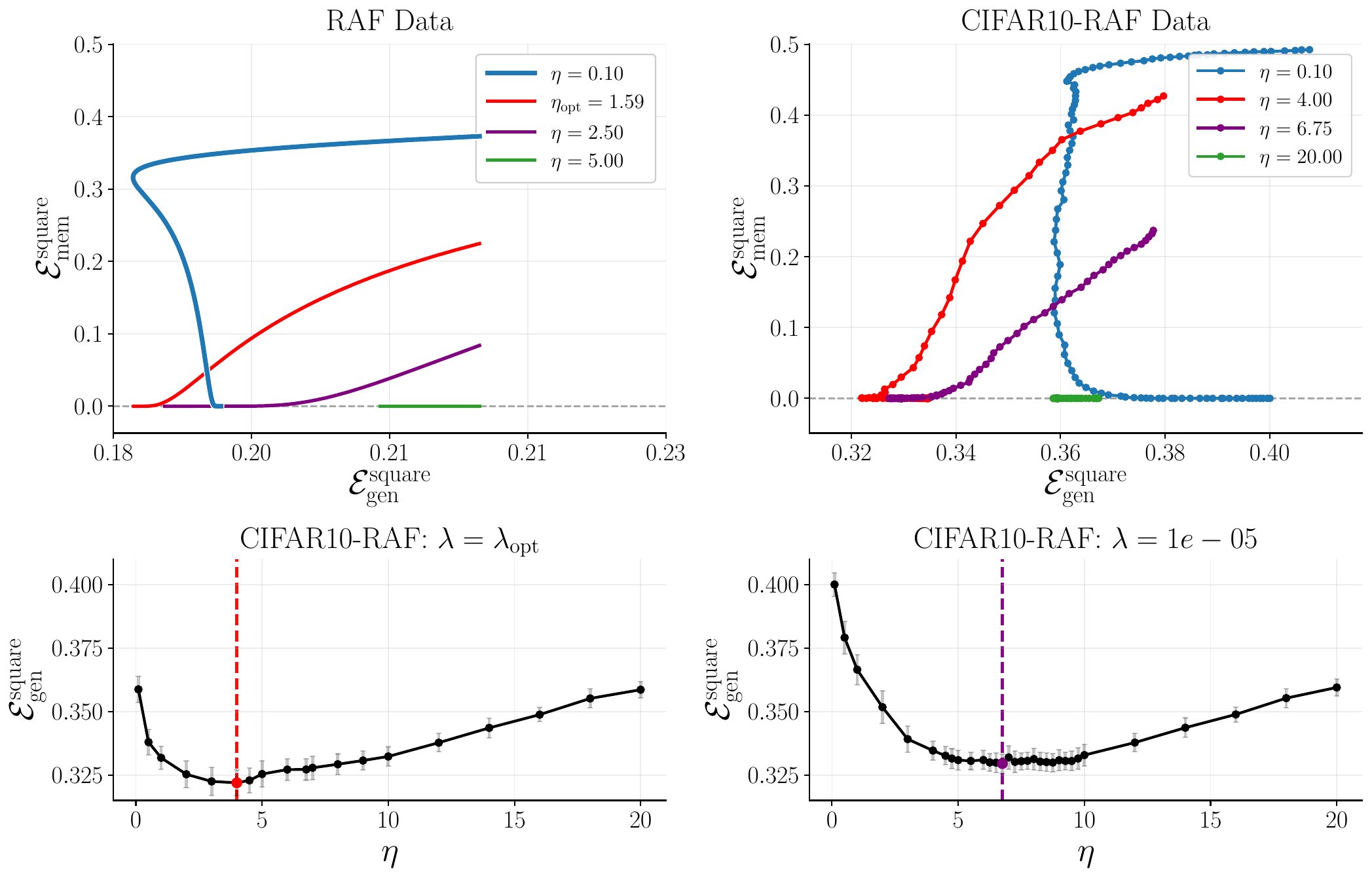}
    \caption{\textbf{Qualitative comparison between theory for the RAF model (upper left panel) and experiments on real data, namely the CIFAR10-RAF task (lower and right panels).}    
In all panels, we use RBF kernel ridge regression with $\alpha = 4.0$ and $\varepsilon = 0.2$. The upper panels show the parametric memorization--generalization curves $\lambda \mapsto (\mathcal E_{\mathrm{gen}}(\lambda), \mathcal E_{\mathrm{mem}}(\lambda))$.
The kernel bandwidth $\eta$ for CIFAR10-RAF is selected based on the lower panels: generalization error as a function of $\eta$ at optimal regularization (left), with a minimum around $\eta \approx 4$ (red dashed line), and at small regularization (right), with a minimum around $\eta \approx 6.75$ (purple dashed line).
The parametric curves are then shown for these two values of $\eta$, together with one larger and one smaller representative value. The RAF model exhibits qualitatively similar behavior as a function of $\eta$, with the corresponding values shifted. 
}
\label{fig:RAF-CIFAR}
\end{figure}

In Fig.~\ref{fig:RAF-CIFAR} left top panel, we illustrate the described memorization-generalization phenomenology for the widely used isotropic RBF kernel that we write as a function of the normalized correlation $\rho$ and as a function of the bandwidth-like parameter $\eta$: 
\begin{equation}
K(\rho)=e^{-\eta(1-\rho)}\,  \quad {\rm where} \quad \rho=\bx^\top \bx' / (\|\bx\|\,\|\bx'\|).
\end{equation}
As we derived, for Gaussian input data in high dimension, the asymptotic performance depends on the kernel only through
\begin{equation}
    \mu_1=\sqrt{\eta e^{-\eta}},  \quad \mu_\star=\sqrt{1-e^{-\eta}(1+\eta)}.
\end{equation}
In Fig.~\ref{fig:RAF-CIFAR} left top panel, each curve corresponds to a different value of the bandwidth parameter $\eta$ for the RAF model at given representative values of $(\alpha,\varepsilon)$.
The highlighted value $\eta_{\mathrm{opt}}$ is chosen so that the kernel corresponds to the optimal angle $\gamma_{\rm mem}^{\rm opt}$ \eqref{eq:angle_opt_mem}.

To test whether the qualitative phenomenology identified above persists for structured inputs, we reproduce the rules-and-facts construction using the widely studied CIFAR10 benchmark \cite{krizhevsky2009learning}.
We build a training set that we call CIFAR10-RAF to mimic the RAF construction: two of the CIFAR10 classes provide the structured part of the task, with labels fixed by the class (cat as $+1$, dog as $-1$), while a third class (automobile) provides the factual/random subset, whose labels are assigned independently $\pm1$ with equal probability.
The dataset is constructed so that the total sample size corresponds to $\alpha=4.0$ (this dataset has $d=3072$) and a fraction $\varepsilon=0.2$ of the training examples belong to the randomized subset.
The test set contains $n_{\rm test}=2000$ new samples only from the two structured classes, so the reported test error measures generalization to the rule component, while the memorization error is computed on the training randomly labeled subset.
Inputs are preprocessed by feature-wise $z$-scoring, and the effective bandwidth is taken relative to the typical train-set squared distance,
$\eta_{\mathrm{eff}}=\eta/\mathrm{median}(\|\bx_i-\bx_j\|^2)$. The regularization in the CIFAR10 plot is spanned in $\lambda \in [10^{-5},10^2]$.

For the purpose of testing how the theory is reflected on the CIFAR10 (instead of i.i.d. Gaussian) input, we run kernel ridge regression with a full RBF kernel ${K(\bx,\bx')=\exp\bigl(-\eta \|\bx-\bx'\|_2^2\bigr)}$ on the CIFAR10-RAF dataset using the standard square loss. We depict the memorization--generalization trade-off (parametrically in the regularization $\lambda$) in Fig.~\ref{fig:RAF-CIFAR} top right panel. The curves shown are averages over 20 repeated random draws of the CIFAR10 training/test subsets and randomized labels.

We remind that for structured data, the performance of the kernel method depends on the kernel through more than just the $\mu_1,\mu_\star$ parameters, see Appendix \ref{appE:cifar-matching}. Still in Fig.~\ref{fig:RAF-CIFAR}, we observe a qualitatively similar phenomenology for the RAF and the CIFAR10-RAF data when tuning the RBF bandwidth $\eta$. Some aspects are absent in the CIFAR10-RAF data -- such as the independence of the optimal generalization error on $\eta$.
However overall, we see that the generalization error for CIFAR10-RAF depends on the bandwidth $\eta$ similarly as the theory predicts for the hinge loss, i.e. the generalization error at $\lambda \to 0^+$ is minimized at one value of the bandwidth, here roughly $\eta_{\rm mem} \approx 6.75$, and the generalization error at optimal regularization is minimized at a lower bandwidth value, here roughly at $\eta_{\rm gen} \approx 4.0$. Representative lower and higher values of the bandwidth display qualitatively similar behaviour as predicted by the theory. In Appendix~\ref{appE:cifar-matching} we present the same comparison as in Fig.~\ref{fig:RAF-CIFAR} for the hinge loss for which we also observe a qualitative agreement between the CIFAR10-RAF and the theory. 

Obtaining quantitative agreement between the theory and the performance of kernel methods on CIFAR10-RAF would require both a more refined model of the data structure and a description of the kernel beyond the two-parameter reduction to $\mu_1$ and $\mu_\star$. Models incorporating correlated or mixture-structured data \cite{monasson1992properties,loureiro2021learning,loureiro2021learninggaussian,dandi2023universality}, related mixtures of teachers \cite{sarao2025bias}, and approaches based on empirical Gram matrices or higher-order kernel expansions \cite{karkada2026predicting,dietrich1999statistical} offer possible routes towards such refinements. These extensions could be adapted to evaluate factual recall alongside test performance, but would still provide a theory of fixed feature maps or kernels. Our purpose here is different: to establish that the qualitative trade-off persists on structured data before asking how the picture changes when the representation itself is learned, as in standard neural networks.

\subsection{Learned representations mitigate the trade-off}
\label{subsec:learned_representations}

The comparison above changes the data distribution while keeping the representation fixed through the choice of kernel. We now ask how the memorization--generalization trade-off changes when the representation itself is learned from CIFAR10-RAF. Better test performance is expected, but it does not by itself answer this question: a network could improve its generalization on the structured examples simply by discarding the randomly labelled facts. The relevant question is therefore whether feature learning can improve rule generalization while keeping the individual factual associations accessible.

We use the CIFAR10-RAF construction of Sec.~\ref{main:Cifar} with $\alpha=2$ and $\varepsilon=0.2$. The training set contains $n=6144$ examples, including $n_{\rm mem}=1228$ randomly labelled automobiles. In the main comparison we use a wide two-layer CNN and a parameter-matched four-layer CNN; additional random-feature and pre-activation WRN-16-4 controls are reported in Appendix~\ref{app:learned_representation_protocol}. For each trainable architecture, we first train the complete network once at zero explicit regularization with square loss. We
then remove the scalar output, freeze the resulting nonlinear representation
$\phi$, and refit only an affine readout. For each fixed feature matrix $\Phi$, this readout is the exact solution of the convex problem
\begin{equation}
    \widehat{w}_{\lambda}
    = \operatorname*{arg\,min}_{w}
    \left\{
        \frac{1}{2n}\lVert\Phi w-y\rVert_2^2
        + \frac{\lambda}{2n}\lVert w\rVert_2^2
    \right\}.
    \label{eq:fixed_representation_ridge}
\end{equation}
Every point on the resulting $\lambda$-dependent curves is therefore a global optimum after the representation has been frozen. Full numerical details and all controls are reported in Appendix~\ref{app:learned_representation_protocol}.

\begin{figure}[!t]
    \centering
    \includegraphics[width=0.8\textwidth]{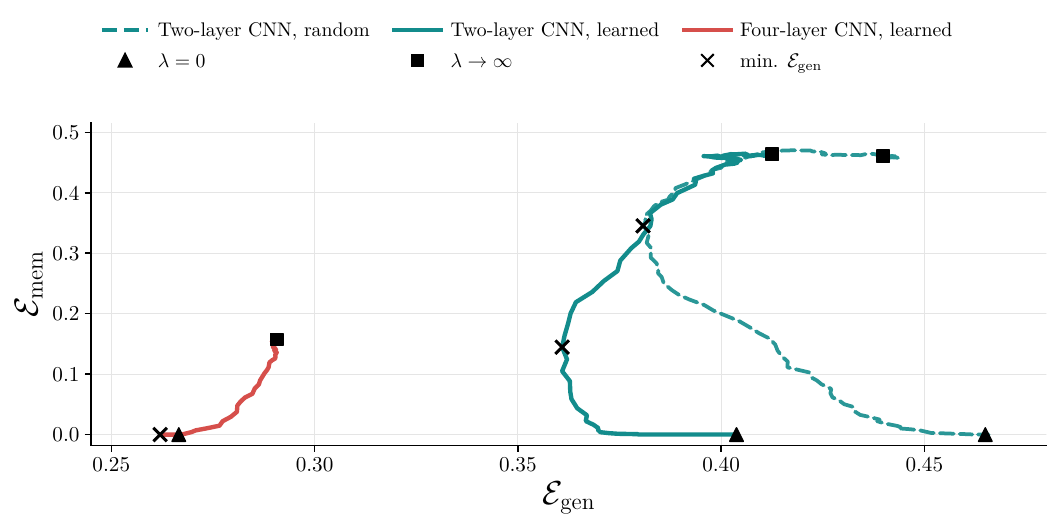}
    \caption{\textbf{Feature learning can preserve factual recall with optimal test performance.}
    Memorization error $\mathcal{E}_{\rm mem}(\lambda)$ is plotted against clean cat--dog test error $\mathcal{E}_{\rm gen}(\lambda)$ along three ridge-readout paths. The dashed curve is the random two-layer representation (equivalently, the frozen RF-CNN); the solid teal and red curves are the two- and four-layer representations learned without weight decay and then frozen. Each curve spans the readout regularization from $\lambda=0$ to the limit $\lambda\to\infty$. Black triangles and squares mark these endpoints; crosses mark the smallest mean clean test error on each path. Lines connect exact five-seed mean solutions in increasing $\lambda$, without smoothing.
    The dataset is common to all seeds. 
    }
    \label{fig:learned_representation_ridge}
\end{figure}

Figure~\ref{fig:learned_representation_ridge} shows the resulting memorization-generalization curves, with $\lambda$ varying parametrically from zero to the $\lambda\to\infty$ limit. As a fixed-feature reference, we include the randomly initialized representation of the wide two-layer CNN. It exhibits the familiar trade-off: the unregularized readout memorizes the randomized labels, but the readout with the best test performance forgets a substantial fraction of them. Learning the two-layer representation improves both test performance and factual recall at the test-optimal point, although the trade-off does not disappear completely in this case.

The representation learned by the four-layer CNN behaves differently. Its unregularized readout already combines substantially better test performance with perfect factual recall, and the readout attaining the minimum test error still classifies every randomized example correctly. Its improved rule generalization is therefore not obtained by making the individual facts inaccessible.

This behavior is reminiscent of the special kernel geometry identified by our KRR analysis at the optimal angle $\gamma_{\rm mem}^{\rm opt}$, for which the zero-regularization solution already attains optimal generalization and positive regularization is therefore unnecessary. The present experiment does not explain how neural-network training constructs such representations, nor does it establish that architectural depth alone is responsible for the observed progression. It shows more narrowly that feature learning can produce representations in which rule-relevant and fact-specific information remain simultaneously accessible to a linear readout. Appendix~\ref{app:learned_representation_protocol} reports the complete readout controls, including an additional WRN-16-4 architecture, and end-to-end weight-decay sweeps with the same qualitative tendency. Developing an analytical understanding of how training in deeper neural networks produces this favorable geometry, and which architectural ingredients are responsible, is therefore a central open question raised by these results.

\section{Conclusion}

We introduced the Rules-and-Facts (RAF) framework to initiate a theoretical study of how neural networks can simultaneously generalize structured rules and memorize unstructured facts. While both capabilities are central to modern AI systems, they have largely been examined in isolation within existing theoretical studies.

Our analysis, carried out in the high-dimensional limit, reveals a clear structural picture. Linear models exhibit an unavoidable trade-off between rule learning and fact memorization. In contrast, overparameterized models---even in their simplest analytically tractable form of random features and kernel regression---admit regimes where memorization and generalization coexist. Regularization and kernel geometry, through the parameter $\gamma$ defined in Eq.~(39), govern how capacity is allocated between the two objectives. For kernel ridge regression, geometries with $\gamma\geq\gamma_{\rm mem}^{\rm opt}$ can attain the lowest possible generalization error after tuning the regularization. For $\gamma>\gamma_{\rm mem}^{\rm opt}$, however, doing so requires positive regularization and therefore comes at the cost of factual recall. Precisely at $\gamma=\gamma_{\rm mem}^{\rm opt}$, the zero-regularization solution simultaneously attains optimal generalization and perfect factual recall. For the hinge loss, kernel geometry produces a qualitatively similar trade-off, but the two optima do not coincide exactly.

Beyond the specific results, the RAF framework provides a minimal setting in which tensions commonly observed in practice---benign overfitting, interpolation, and capacity allocation---can be analyzed quantitatively and in closed form. Our experiments ask whether learned neural-network representations display the favorable behavior predicted by the KRR theory at its special geometry. For the deeper architectures tested, the representation learned without explicit weight decay improves test performance while maintaining perfect factual recall, and the readout selected for the best test performance continues to classify every fact correctly. In this operational sense, feature learning appears to construct the kind of favorable geometry identified by the KRR theory.

Understanding how neural-network training constructs such representations, and which architectural properties are responsible, is therefore an important open question. A first direction is to extend the analysis to feature-learning architectures and determine whether they can achieve faster generalization rates, potentially approaching the Bayes-optimal scaling, while preserving memorization. Second, richer architectures such as multi-layer networks and attention-based models may allow different functional components to specialize, potentially separating rule-like computation from fact-like storage. In particular, understanding whether and when memorization localizes in specific architectural submodules---such as attention heads rather than feedforward layers, or vice versa---remains an intriguing question.

Finally, the interplay between memorization and generalization has long been studied in cognitive science and neuroscience, see e.g.~\cite{sun2023organizing}, often framed in terms of complementary learning systems. The RAF task isolates a minimal version of the same computational tension, but whether its analysis can inform models of biological learning remains an open question. At a broader level, RAF shares with Valiant's notion of ``educability'' \cite{valiant2024importance} an interest in systems capable of retaining and using different forms of acquired information, although educability addresses a much wider range of processes. Exploring whether the RAF framework can have implications in either context is an interesting avenue for future work.

An important direction for future work is to move beyond the dichotomy between beneficial and detrimental memorization. In realistic learning systems, memorization may play multiple roles: enabling the storage of rare but essential information, or giving rise to unwanted retention of spurious or sensitive data. Our work may inspire new frameworks for controlled settings in which such distinct forms of memorization can be disentangled and studied systematically. 

More broadly, many tasks intrinsically require an active interaction between memorization and generalization, rather than their mere coexistence. Exceptions may refine rules, stored facts may guide inference under ambiguity, and memory may serve as a scaffold for abstraction. Understanding these mechanisms — and the architectural or dynamical principles that regulate them — represents a promising avenue toward a richer theory of learning in both artificial \cite{jelassi2024mixture} and biological systems \cite{sun2023organizing}.

\section*{Acknowledgment}

We thank Emanuele Troiani, Vittorio Erba, Freya Behrens, Valentina Njaradi, and Florent Krzakala for fruitful discussions. 
We acknowledge that the initial idea to study the RAF model came from a conversation with ChatGPT that is transcribed in the Appendix \ref{app:chatgpt} for reference.  

We acknowledge funding from the Swiss National Science Foundation grants SNSF SMArtNet (grant number 212049), and the Simons Collaboration on the Physics of Learning and Neural Computation via the Simons Foundation grant (\#1257413 (LZ)).

\bibliography{biblio}

\newpage

\appendix

\input{Appendices/Appendix_Replica_review}

\newpage

\input{Appendices/Appendix_Kernel_review}

\newpage

\input{Appendices/Appendix_KRR-SVM-RAF_review}

\newpage

\input{Appendices/Appendix_BO-RAF_review}

\newpage

\input{Appendices/Appendix_Numerics_review}

\input{Appendices/Appendix_AI.tex}

\end{document}

%% file: math_commands.tex
\renewcommand{\bw}{{\boldsymbol{w}}}

\renewcommand{\bx}{{\boldsymbol{x}}}

\renewcommand{\vec}[1]{\boldsymbol{#1}}

  \providecommand{\R}{\mathbb{R}}

  \makeatletter
  \def\sign{\@ifnextchar*{\@sgnargscaled}{\@ifnextchar[{\sgnargscaleas}{\@ifnextchar{\bgroup}{\@sgnarg}{\sgn} }}}
  \def\@sgnarg#1{\sgn\rbr{#1}}
  \def\@sgnargscaled#1{\sgn\rbr*{#1}}
  \def\@sgnargscaleas[#1]#2{\sgn\rbr[#1]{#2}}
  \makeatother

  \providecommand{\dd}{\bm{d}}

\providecommand{\E}{\mathbb{E}}

  \providecommand{\cD}{\mathcal{D}}
  \providecommand{\cE}{\mathcal{E}}

  \providecommand{\cN}{\mathcal{N}}

  \providecommand{\cZ}{\mathcal{Z}}


\newcommand{\speedup}[1]{{\color{gray}(\ifdim #1 pt > 0.3pt #1\else $< #1$\fi{}$\times$)}}

\usepackage{amsfonts}
\usepackage{amsmath}
\usepackage{amsthm}
\usepackage{amssymb}
\usepackage{dsfont}

\usepackage{xcolor}
\usepackage{color}
\usepackage{graphicx}

\usepackage{verbatim}
\usepackage{xspace} 
\usepackage{enumerate}
\usepackage{enumitem}

\makeatletter
\newsavebox{\@brx}
\newcommand{\llangle}[1][]{\savebox{\@brx}{\(\m@th{#1\langle}\)}
  \mathopen{\copy\@brx\mkern2mu\kern-0.9\wd\@brx\usebox{\@brx}}}
\newcommand{\rrangle}[1][]{\savebox{\@brx}{\(\m@th{#1\rangle}\)}
  \mathclose{\copy\@brx\mkern2mu\kern-0.9\wd\@brx\usebox{\@brx}}}
\makeatother

  \providecommand{\R}{\mathbb{R}}

  \providecommand{\EE}[2]{{\mathbb E}_{#1}\left.#2\right. }

  \providecommand{\dd}{\mathbf{d}}

  \providecommand{\cD}{\mathcal{D}}
  \providecommand{\cE}{\mathcal{E}}

  \providecommand{\cN}{\mathcal{N}}

  \providecommand{\cZ}{\mathcal{Z}}

\renewcommand{\vec}{\mathbf}
  \usepackage{bm}

\providecommand{\mycomment}[3]{\todo[caption={},size=footnotesize,color=#1!20]{\textbf{#2: }#3}}
\providecommand{\inlinecomment}[3]{
  {\color{#1}#2: #3}}
\newcommand\commenter[2]
{
  \expandafter\newcommand\csname i#1\endcsname[1]{\inlinecomment{#2}{#1}{##1}}
  \expandafter\newcommand\csname #1\endcsname[1]{\mycomment{#2}{#1}{##1}}
}

  \definecolor{mydarkblue}{rgb}{0,0.08,0.45}
  \usepackage{hyperref}
  \hypersetup{ 
    pdftitle={},
    pdfauthor={},
    pdfsubject={},
    pdfkeywords={},
    pdfborder=0 0 0,
    pdfpagemode=UseNone,
    colorlinks=true,
    linkcolor=mydarkblue,
    citecolor=mydarkblue,
    filecolor=mydarkblue,
    urlcolor=mydarkblue,
    pdfview=FitH}

\usepackage{url}

%% file: Appendices/Appendix_Replica_review.tex
\section{Proof of Eq.~\texorpdfstring{\eqref{eq:mmem-error}}{(A)} from the random-features replica analysis}
\label{app:replica_analysis}
In this appendix we derive Eq.~\eqref{eq:mmem-error} for the memorization error. The derivation builds on the replica analysis of the empirical risk minimization (ERM) problem for random-features regression developed in \cite{gerace2020generalisation}. While the steps of the replica computation are analogous, we explicitly distinguish randomly labeled samples from teacher-labeled ones in order to identify the terms in the free energy density that contribute to the memorization error of the empirical risk minimizer. Moreover, our analysis covers activation functions that are either centered or non-centered with respect to the Gaussian measure, corresponding respectively to $\mathbb{E}_{g \sim \mathcal{N}(0,1)}[\sigma(g)] = 0$ and $\mathbb{E}_{g \sim \mathcal{N}(0,1)}[\sigma(g)] \neq 0$. The main line of the derivation is carried out under the assumption that $\mathbb{E}_{g \sim \mathcal{N}(0,1)}[\sigma(g)] \neq 0$. At the relevant stages, we separately report the corresponding results that would be obtained by imposing $\mathbb{E}_{g \sim \mathcal{N}(0,1)}[\sigma(g)] = 0$ from the outset, as assumed in \cite{gerace2020generalisation}.

Since we obtain the kernel state equations as the large-width limit of the random-features state equations, and since linear regression is recovered as the special case corresponding to a linear kernel, the resulting equation for the memorization error derived from the random-features computation formally applies to all these cases. The same holds for Eq.~\eqref{eq:RF-gen-err} for the generalization error, which was derived in \cite{gerace2020generalisation}.

\subsection{Setting and notation}
First, we recall the setting and introduce the notation used throughout the following sections.

Consider a training dataset $\mathcal{D} = \{ (\boldsymbol{x}^\mu, y^\mu) \}_{\mu = 1}^n$ consisting of $n$ independently labeled samples. The inputs $\boldsymbol{x}^\mu \in \mathbb{R}^d$ are i.i.d. draws from $\mathcal{N}(\boldsymbol{0}, \mathbb{I}_d)$, while the labels $y^\mu \in \{ \pm 1 \}$ are generated according to the output distribution
\begin{equation} \label{eq: appendix - RAF output channel}
    P_\mathrm{out^\star}\!\left( y^\mu \,\middle|\, \tfrac{1}{\sqrt{d}} \boldsymbol{w}_\star \cdot \boldsymbol{x}^\mu \right) = \varepsilon P_\mathrm{mem}(y^\mu) + (1 - \varepsilon) P_\mathrm{gen}\!\left( y^\mu \,\middle|\, \tfrac{1}{\sqrt{d}} \boldsymbol{w}_\star \cdot \boldsymbol{x}^\mu \right).
\end{equation}
In Eq.~\eqref{eq: appendix - RAF output channel} we distinguish the channel associated with the random labels
\begin{equation} \label{eq: appendix - mem output channel}
    P_\mathrm{mem}(y^\mu) = \frac{1}{2} \delta(y^\mu + 1) + \frac{1}{2} \delta(y^\mu - 1)
\end{equation}
from the channel linked to the teacher
\begin{equation}
    P_\mathrm{gen}\!\left( y^\mu \,\middle|\, \tfrac{1}{\sqrt{d}} \boldsymbol{w}_\star \cdot \boldsymbol{x}^\mu \right) = \delta\!\left(y^\mu - \operatorname{sign}\left(\tfrac{1}{\sqrt{d}} \boldsymbol{w}_\star \cdot \boldsymbol{x}^\mu \right)\right).
\end{equation}
In addition, the teacher weights are assumed to follow a Gaussian prior $P_{\boldsymbol{w}_\star} = \mathcal{N}(\boldsymbol{0}, \mathbb{I}_d)$.

Let $\hat{\boldsymbol{w}}$ be the minimizer of the strictly convex empirical risk
\begin{equation} \label{eq: appendix - empirical risk}
    \mathcal{L}(\boldsymbol{w}; \mathcal{D}) = \sum_{\mu=1}^n \ell \bigl( y^\mu, f_{\boldsymbol{w}}(\boldsymbol{x}^\mu) \bigr) + \frac{\lambda}{2}  r(\boldsymbol{w} ) = \sum\limits_{\mu \in \mathcal{I}_\mathrm{mem}} \ell (y^\mu, f_{\boldsymbol{w}}(\boldsymbol{x}^\mu) ) + \sum\limits_{\mu \in \mathcal{I}_\mathrm{gen}} \ell (y^\mu, f_{\boldsymbol{w}}(\boldsymbol{x}^\mu) ) + \frac{\lambda}{2}  r(\boldsymbol{w} ),
\end{equation}
where $\ell$ is a convex loss, $f_{\boldsymbol{w}}$ is the pre-activation function of the model, $r$ is an appropriately defined ridge regularization on the trained weights, and $\lambda$ is a positive constant. In \eqref{eq: appendix - empirical risk}, we emphasize that the data-dependent term of the empirical risk can be decomposed into two sums by introducing the index sets $\mathcal{I}_\mathrm{mem}$ and $\mathcal{I}_\mathrm{gen}$, corresponding to the randomly-labeled samples and the teacher-appointed ones, respectively.

We evaluate the minimizer $\hat{\boldsymbol{w}}$ through its capacity to memorize the random labels in the training set and its ability to generalize on the learnable ones. These properties are measured by the memorization error
\begin{equation} \label{eq: appendix - memorization error at finite n}
    \mathcal{E}_\mathrm{mem}(\hat{\boldsymbol{w}}; \mathcal{D}, \varepsilon_n) = \frac{1}{ \varepsilon_n n } \sum_{\mu \in \mathcal{I}_\mathrm{mem}} \mathbbm{1}_{y^\mu \neq \operatorname{sign}( f_{\hat{\boldsymbol{w}}}(\boldsymbol{x}^\mu) )} = \frac{1}{\varepsilon_n n} \sum_{\mu \in \mathcal{I}_\mathrm{mem}} \theta\left( - y^\mu f_{\hat{\boldsymbol{w}}}(\boldsymbol{x}^\mu) \right)
\end{equation}
and the generalization error
\begin{equation} \label{eq: appendix - generalization error at finite n}
    \mathcal{E}_\mathrm{gen}(\hat{\boldsymbol{w}}) = \mathbb{E}_{ \boldsymbol{x}^\mathrm{new}, y^\mathrm{new}} \left[ \mathbbm{1}_{ y^\mathrm{new} \neq \operatorname{sign} ( f_{\hat{\boldsymbol{w}}}(\boldsymbol{x}^\mathrm{new}) )} \right] = \mathbb{E}_{ \boldsymbol{x}^\mathrm{new}, y^\mathrm{new}} \left[ \theta\left( - y^\mathrm{new} f_{\hat{\boldsymbol{w}}}(\boldsymbol{x}^\mathrm{new}) \right) \right].
\end{equation}
In \eqref{eq: appendix - memorization error at finite n} we introduced the fraction of randomly-labeled samples in the training dataset ${\varepsilon_n = |\mathcal{I}_\mathrm{mem}| / n}$, which is a scaled binomial random variable.

In what follows, we focus on the random features setting, where $f_{\boldsymbol{w}}(\boldsymbol{x}^\mu) = \frac{1}{\sqrt{p}} \boldsymbol{w} \cdot \sigma\!\left(\frac{1}{\sqrt{d}} F^\top \boldsymbol{x}^\mu\right)$ and $r(\boldsymbol{w})= ||\boldsymbol{w} ||_2^2$. Here, $\boldsymbol{w} \in \mathbb{R}^p$, $F \in \mathbb{R}^{d \times p}$ is a fixed random matrix with i.i.d. entries $F_{ia} \sim \mathcal{N}(0,1)$, and $\sigma: \mathbb{R} \to \mathbb{R}$ a non-linear activation function applied component-wise.

Finally, we study the performance of the minimizer in the high-dimensional limit, where
\begin{equation} 
n, d, p \to \infty\, , \quad \alpha = \frac{n}{d} = \Theta(1)\, ,  \quad \kappa = \frac{p}{d} = \Theta(1)\, . 
\end{equation}

\subsection{Gibbs formulation of the ERM problem}
\label{sec: appendix - Gibbs formulation of the minimization problem}
The ERM in the high-dimensional limit is equivalent to solving a set of low-dimensional self-consistent equations for a set of scalar order parameters that fully characterize the minimizer. Within statistical physics, this is shown via the replica method.

To begin with, we define the following Gibbs measure over the weights $\boldsymbol{w} \in \mathbb{R}^p$:
\begin{equation} \label{eq: appendix - gibbs measure}
    \mu_{\beta}(\dd \boldsymbol{w}) = \frac{1}{\mathcal{Z}_{\beta}(\mathcal{D})} e^{- \beta \mathcal{L}(\boldsymbol{w}; \mathcal{D})} \dd \boldsymbol{w},
\end{equation}
where $\mathcal{Z}_\beta(\mathcal{D})$ is the normalization or partition function of the measure:
\begin{equation}
    \mathcal{Z}_{\beta}(\mathcal{D}) = \int_{\mathbb{R}^p} \dd \boldsymbol{w} \, e^{ -\beta \mathcal{L}(\boldsymbol{w}; \mathcal{D}) }.
\end{equation}
Sending the inverse temperature $\beta$ to infinity, the Gibbs measure \eqref{eq: appendix - gibbs measure} concentrates on the global minimizer of the risk \eqref{eq: appendix - empirical risk}.

Then, the goal of the replica method is to compute the free energy density
\begin{equation} \label{eq: appendix - free energy density}
    f_{\beta} = - \frac{1}{\beta} \lim_{p \to + \infty} \frac{1}{p} \mathbb{E}_\mathcal{D} \Bigl[ \log \mathcal{Z}_{\beta}(\mathcal{D}) \Bigr]
\end{equation}
and evaluate it for $\beta \to +\infty$. Indeed, in this limit the free energy density \eqref{eq: appendix - free energy density} converges to the global minimum of the energy functional of the problem, that is just the empirical risk \eqref{eq: appendix - empirical risk}. Then, recasting the right hand side of Eq.~\eqref{eq: appendix - free energy density} as a saddle-point integral in the large $p$ limit, the order parameters describing the asymptotic minimizer arise naturally as the associated saddle-point solutions.

Rather than computing the free energy density defined in \eqref{eq: appendix - free energy density}, as done in \cite{gerace2020generalisation}, we will compute a closely related quantity, namely
\begin{equation} \label{eq: appendix - free energy density with field}
    f_{\beta, h} = - \frac{1}{\beta} \lim_{p \to + \infty} \frac{1}{p} \mathbb{E}_{\mathcal{D}, \varepsilon_n} \Bigl[ \log \mathcal{Z}_{\beta,h}(\mathcal{D}, \varepsilon_n) \Bigr],
\end{equation}
where the corresponding partition function is defined as
\begin{equation}
    \mathcal{Z}_{\beta,h}(\mathcal{D}, \varepsilon_n) = \int_{\mathbb{R}^p} \dd \boldsymbol{w} \, e^{ -\beta \mathcal{L}(\boldsymbol{w}; \mathcal{D}) - \beta h n \mathcal{E}_\mathrm{mem}(\boldsymbol{w}; \mathcal{D}, \varepsilon_n) }.
\end{equation}
We introduce this modified definition because it enables us to extract the asymptotic memorization error of the risk minimizer, which is given by
\begin{equation} \label{eq: appendix - mem error from free energy density}
    \mathcal{E}_\mathrm{mem} = \frac{\kappa}{\alpha} \lim_{\beta \to \infty} \frac{\partial f_{\beta, h}}{\partial h} \Bigg|_{h=0}.
\end{equation}
When $h=0$, the free energy density $f_{\beta, h}$ reduces to the original free energy density $f_\beta$.

\subsection{Replica computation of the free energy density \texorpdfstring{$f_{\beta, h}$}{f[beta, h]}}
\label{app:replica_comp}
Using the replica trick, the free energy density $f_{\beta, h}$ can be rewritten as
\begin{equation} \label{eq: appendix - free energy density with field replica trick}
    f_{\beta, h} = - \frac{1}{\beta} \lim_{p \to + \infty} \lim_{r \to 0^+} \frac{1}{p} \frac{\mathbb{E}_{\mathcal{D}, \varepsilon_n} [ \mathcal{Z}_{\beta, h}(\mathcal{D}, \varepsilon_n)^r ] - 1}{r}.
\end{equation}
Therefore, the whole problem reduces to compute the average replicated partition function.

\subsubsection*{The replicated partition function}
The partition function can be massaged as follows:
\begin{equation}
    \begin{split}
        &\mathcal{Z}_{\beta, h}(\mathcal{D}, \varepsilon_n)
        = \int_{\mathbb{R}^p} \dd \boldsymbol{w} \, e^{ -\beta \left[ \, \sum\limits_{\mu = 1}^n \ell (y^\mu, f_{\boldsymbol{w}}(\boldsymbol{x}^\mu) ) + \frac{\lambda}{2} || \boldsymbol{w} ||_2^2 \, \right] - \beta \frac{h}{\varepsilon_n} \sum\limits_{\mu \in \mathcal{I}_\mathrm{mem}} \theta(- y^\mu f_{\boldsymbol{w}}(\boldsymbol{x}^\mu)) }
        \\[6pt]
        &= \int_{\mathbb{R}^p} \dd \boldsymbol{w} \, e^{ -\beta \sum_{\mu \in \mathcal{I}_\mathrm{mem}} \ell (y^\mu, f_{\boldsymbol{w}}(\boldsymbol{x}^\mu) ) - \beta \sum_{\mu \in \mathcal{I}_\mathrm{gen}} \ell (y^\mu, f_{\boldsymbol{w}}(\boldsymbol{x}^\mu) ) - \frac{\beta \lambda}{2} || \boldsymbol{w} ||_2^2 - \beta \frac{h}{\varepsilon_n} \sum_{\mu \in \mathcal{I}_\mathrm{mem}} \theta(- y^\mu f_{\boldsymbol{w}}(\boldsymbol{x}^\mu)) }
        \\
        &= \int_{\mathbb{R}^p} \dd \boldsymbol{w} \, e^{ -\beta \sum_{\mu \in \mathcal{I}_\mathrm{mem}} \left[ \ell (y^\mu, f_{\boldsymbol{w}}(\boldsymbol{x}^\mu) ) + \frac{h}{\varepsilon_n} \theta(- y^\mu f_{\boldsymbol{w}}(\boldsymbol{x}^\mu)) \right] } \, e^{ -\beta \sum_{\mu \in \mathcal{I}_\mathrm{gen}} \ell (y^\mu, f_{\boldsymbol{w}}(\boldsymbol{x}^\mu) ) } \, e^{- \frac{\beta \lambda}{2} || \boldsymbol{w} ||_2^2}
        \\
        &= \int_{\mathbb{R}^p} \dd \boldsymbol{w} \, \Biggl( \, \prod_{\mu \in \mathcal{I}_\mathrm{mem}} \underbrace{e^{ -\beta \bigl[ \ell (y^\mu, f_{\boldsymbol{w}}(\boldsymbol{x}^\mu) ) + \frac{h}{\varepsilon_n} \theta(- y^\mu f_{\boldsymbol{w}}(\boldsymbol{x}^\mu)) \bigr] }}_{P_{\ell \theta}(y^\mu \,|\, f_{\boldsymbol{w}}(\boldsymbol{x}^\mu))} \, \Biggr) \Biggl( \, \prod_{\mu \in \mathcal{I}_\mathrm{gen}} \underbrace{e^{ -\beta \ell (y^\mu, f_{\boldsymbol{w}}(\boldsymbol{x}^\mu) ) }}_{P_\ell(y^\mu \,|\, f_{\boldsymbol{w}}(\boldsymbol{x}^\mu))} \, \Biggr) \underbrace{e^{- \frac{\beta \lambda}{2} || \boldsymbol{w} ||_2^2}}_{P_{\boldsymbol{w}}(\boldsymbol{w})}
        \\
        &= \int_{\mathbb{R}^p} \dd \boldsymbol{w} \, P_{\boldsymbol{w}}(\boldsymbol{w}) \Biggl( \, \prod_{\mu \in \mathcal{I}_\mathrm{mem}} P_{\ell \theta}(y^\mu \,|\, f_{\boldsymbol{w}}(\boldsymbol{x}^\mu)) \, \Biggr) \Biggl( \, \prod_{\mu \in \mathcal{I}_\mathrm{gen}} P_\ell(y^\mu \,|\, f_{\boldsymbol{w}}(\boldsymbol{x}^\mu)) \, \Biggr).
    \end{split}
\end{equation}
Thus, the replicated partition function is equal to
\begin{equation}
    \mathcal{Z}_{\beta, h}(\mathcal{D}, \varepsilon_n)^r = \int_{(\mathbb{R}^p)^r} \Biggl( \, \prod_{a=1}^r \dd \boldsymbol{w}^a \, P_{\boldsymbol{w}}(\boldsymbol{w}^a) \Biggr) \Biggl( \, \prod_{\substack{\mu \in \mathcal{I}_\mathrm{mem} \\ a=1,\dots,r}} P_{\ell \theta}(y^\mu \,|\, f_{\boldsymbol{w}^a}(\boldsymbol{x}^\mu) \, \Biggr) \Biggl( \, \prod_{\substack{\mu \in \mathcal{I}_\mathrm{gen} \\ a=1,\dots,r}} P_\ell(y^\mu \,|\, f_{\boldsymbol{w}^a}(\boldsymbol{x}^\mu)) \, \Biggr).
\end{equation}

\subsubsection*{Averaging over the disorder}
The average replicated partition function can be written as
\begin{equation}
    \mathbb{E}_{\mathcal{D}, \varepsilon_n} [ \mathcal{Z}_{\beta, h}(\mathcal{D}, \varepsilon_n)^r ] = \mathbb{E}_{\varepsilon_n} \, \mathbb{E}_{\boldsymbol{w}_\star} \, \mathbb{E}_{\mathcal{D} | \varepsilon_n, \boldsymbol{w}_\star} [ \mathcal{Z}_{\beta, h}(\mathcal{D}, \varepsilon_n)^r ],
\end{equation}
where the conditional average over the dataset reads
\begin{equation}
\begin{aligned}
&\mathbb{E}_{\mathcal{D} \mid \varepsilon_n, \boldsymbol{w}_\star}
\!\left[ \mathcal{Z}_{\beta,h}(\mathcal{D},\varepsilon_n)^r \right]
= \int_{(\mathbb{R}^p)^r}
   \Biggl( \, \prod_{a=1}^r \dd \boldsymbol{w}^a\, P_{\boldsymbol{w}}(\boldsymbol{w}^a) \, \Biggr)
   \\
&\quad \times
   \int_{(\mathbb{R}^d)^{|\mathcal{I}_\mathrm{mem}|}}
   \Biggl( \, \prod_{\mu\in\mathcal{I}_{\mathrm{mem}}}
      \dd \boldsymbol{x}^\mu\, P_{\boldsymbol{x}}(\boldsymbol{x}^\mu) \, \Biggr)
   \int_{\mathbb{R}^{|\mathcal{I}_\mathrm{mem}|}}
   \Biggl( \, \prod_{\mu\in\mathcal{I}_{\mathrm{mem}}}
      \dd y^\mu\, P_\mathrm{mem}(y^\mu) \, \Biggr)
   \\
&\quad \times
   \int_{(\mathbb{R}^d)^{|\mathcal{I}_\mathrm{gen}|}}
   \Biggl( \, \prod_{\mu\in\mathcal{I}_{\mathrm{gen}}}
      \dd \boldsymbol{x}^\mu\, P_{\boldsymbol{x}}(\boldsymbol{x}^\mu) \, \Biggr)
   \int_{\mathbb{R}^{|\mathcal{I}_\mathrm{gen}|}}
   \Biggl( \, \prod_{\mu\in\mathcal{I}_{\mathrm{gen}}}
      \dd y^\mu\,
      P_\mathrm{gen}\!\left(
         y^\mu \,\middle|\,
         \tfrac{1}{\sqrt d}\,\boldsymbol{w}_\star\!\cdot\!\boldsymbol{x}^\mu
      \right) \, \Biggr)
   \\
&\quad \times
   \Biggl( \, \prod_{\substack{\mu\in\mathcal{I}_{\mathrm{mem}}\\ a=1,\dots,r}}
      P_{\ell\theta}\!\left(y^\mu \mid f_{\boldsymbol{w}^a}(\boldsymbol{x}^\mu)
      \right) \, \Biggr)
   \Biggl( \, \prod_{\substack{\mu\in\mathcal{I}_{\mathrm{gen}}\\ a=1,\dots,r}}
      P_{\ell}\!\left(y^\mu \mid f_{\boldsymbol{w}^a}(\boldsymbol{x}^\mu)
      \right) \, \Biggr).
\end{aligned}
\end{equation}

To perform the average over the inputs, we introduce auxiliary variables representing the preactivations of the teacher and the replicated students. For each $\mu = 1,\dots,n$ and $a = 1,\dots,r$, we define
\begin{equation}
    \begin{split}
        & \nu_\mu = \frac{1}{\sqrt{d}} \boldsymbol{w}_\star \cdot \boldsymbol{x}^\mu,
        \\
        & \lambda_\mu^a = f_{\boldsymbol{w}^a}(\boldsymbol{x}^\mu) = \frac{1}{\sqrt{p}} \boldsymbol{w}^a \cdot \sigma\!\left( \frac{1}{\sqrt{d}} F^\top \boldsymbol{x}^\mu \right).
    \end{split}
\end{equation}
We enforce these definitions in the replicated partition function using Dirac delta functions:
\begin{equation}
    \begin{split}
        &\mathbb{E}_{\mathcal{D} | \varepsilon_n, \boldsymbol{w}_\star} [ \mathcal{Z}_{\beta, h}(\mathcal{D}, \varepsilon_n)^r ]
        = \int_{(\mathbb{R}^p)^r} \biggl( \, \prod_{a=1}^r \dd \boldsymbol{w}^a \, P_w(\boldsymbol{w}^a) \biggr)
        \\
        & \times 
        \Biggl\{ \prod_{\mu \in \mathcal{I}_\mathrm{mem}}
        \int_\mathbb{R} \dd y^\mu \, P_\mathrm{mem}(y^\mu) \int_{\mathbb{R}^r} \Biggl( \, \prod_{a=1}^r \dd \lambda_\mu^a \, P_{\ell \theta}(y^\mu \,|\, \lambda_\mu^a) \, \Biggr) \, \mathbb{E}_{\boldsymbol{x}^\mu} \Biggl[ \, \prod_{a = 1}^r \delta \bigl( \lambda_\mu^a - f_{\boldsymbol{w}^a}(\boldsymbol{x}^\mu) \bigr) \, \Biggr] \Biggr\}
        \\
        & \times \Biggl\{ \prod_{\mu \in \mathcal{I}_\mathrm{gen}} \int_\mathbb{R} \dd y^\mu \int_\mathbb{R} \dd \nu_\mu \, P_\mathrm{gen}(y^\mu \,|\, \nu_\mu) \int_{\mathbb{R}^r} \Biggl( \, \prod_{a=1}^r \dd \lambda_\mu^a \, P_\ell(y^\mu \,|\, \lambda_\mu^a) \Biggr)
        \\
        & \qquad \times \mathbb{E}_{\boldsymbol{x}^\mu} \Biggl[ \, \delta\!\left( \nu_\mu - \tfrac{1}{\sqrt{d}} \boldsymbol{w}_\star \cdot \boldsymbol{x}^\mu \right) \prod_{a = 1}^r \delta\bigl( \lambda_\mu^a - f_{\boldsymbol{w}^a}(\boldsymbol{x}^\mu) \bigr) \, \Biggr] \Biggr\}.
    \end{split}
\end{equation}
Let $\boldsymbol{\lambda}_\mu = (\lambda_\mu^1, \dots, \lambda_\mu^r)^\top$. Then, applying a replicated Gaussian equivalence argument analogous to that used in \cite{gerace2020generalisation}, and based on the Gaussian equivalence property (GEP) introduced in \cite{goldt2020modeling}, we claim that 
\begin{equation}
    \begin{split}
        &\mathbb{E}_{\boldsymbol{x}^\mu} \Biggl[ \, \prod_{a = 1}^r \delta\bigl( \lambda_\mu^a - f_{\boldsymbol{w}^a}(\boldsymbol{x}^\mu) \bigr) \, \Biggr] = \mathcal{N}(\boldsymbol{\lambda}_\mu; \boldsymbol{B}, Q), \\
        \\
        & \mathbb{E}_{\boldsymbol{x}^\mu} \Biggl[ \, \delta\!\left( \nu_\mu - \tfrac{1}{\sqrt{d}} \boldsymbol{w}_\star \cdot \boldsymbol{x}^\mu \right) \prod_{a = 1}^r \delta\bigl( \lambda_\mu^a - f_{\boldsymbol{w}^a}(\boldsymbol{x}^\mu) \bigr) \, \Biggr] = \mathcal{N}\!\left( \begin{pmatrix} \nu_\mu \\ \boldsymbol{\lambda}_\mu \end{pmatrix}; \begin{pmatrix} 0 \\ \boldsymbol{B} \end{pmatrix}, \Sigma \right).
    \end{split}
\end{equation}
The mean vector $\boldsymbol{B} \in \mathbb{R}^r$ has components
\begin{equation}
    B^a = \frac{\mu_0}{\sqrt{p}} \boldsymbol{w}^a \cdot \mathbf{1}_p \qquad a = 1,\dots,r,
\end{equation}
while the covariance matrices $Q \in \mathbb{R}^{r \times r}$ and $\Sigma \in \mathbb{R}^{(r+1) \times (r+1)}$ are expressed in terms of the so called overlap order parameters as
\begin{equation}
    \begin{split}
        &\Sigma^{00} = \frac{1}{d} || \boldsymbol{w}_\star ||_2^2,
        \\[6pt]
        &\Sigma^{0a} = \frac{\mu_1}{d} \boldsymbol{w}_\star \cdot \boldsymbol{s}^a \qquad a=1,\dots,r,
        \\
        &\Sigma^{ab} = Q^{ab} = \frac{\mu_\star^2}{p} \boldsymbol{w}^a \cdot \boldsymbol{w}^b + \frac{\mu_1^2}{d} \boldsymbol{s}^a \cdot \boldsymbol{s}^b \qquad 1 \leq a \leq b \leq r,
    \end{split}
\end{equation}
where $\boldsymbol{s}^a = \frac{1}{\sqrt{p}} F \boldsymbol{w}^a$, $a=1,\dots,r$. Under this equivalence, the activation function $\sigma$ enters only through the three coefficients $\mu_0$, $\mu_1$ and $\mu_\star$, defined by
\begin{equation} \label{eq: appendix - definition mus}
    \mu_0 = \mathbb{E}_{g \sim \mathcal{N}(0,1)} \bigl[ \sigma(g) \bigl], \qquad
    \mu_1 = \mathbb{E}_{g \sim \mathcal{N}(0,1)} \bigl[ g \, \sigma(g) \bigl], \qquad
    \mu_\star^2 = \mathbb{E}_{g \sim \mathcal{N}(0,1)} \bigl[ \sigma(g)^2 \bigr] - \mu_0^2 - \mu_1^2.
\end{equation}

The GEP holds in the proportional high-dimensional limit $p, d \to \infty$ with fixed $p / d$, provided that the matrix $F$ satisfies appropriate balance conditions. We refer the reader to \cite{goldt2020modeling} for a precise statement of these conditions and to
Appendix B of \cite{gerace2020generalisation} for further details on the application of the GEP to the replica analysis of random features.

Therefore, after averaging over the inputs, we are left with
\begin{equation} \label{eq: appendix - average replicated partition function before disentanglement}
    \begin{split}
        &\mathbb{E}_{\mathcal{D}, \varepsilon_n} [ \mathcal{Z}_{\beta, h}(\mathcal{D}, \varepsilon_n)^r ]
        = \mathbb{E}_{\varepsilon_n} \int_{\mathbb{R}^d} \dd\boldsymbol{w}_\star \, P_{\boldsymbol{w}_\star}(\boldsymbol{w}_\star) \int_{(\mathbb{R}^p)^r} \biggl( \, \prod_{a=1}^r \dd \boldsymbol{w}^a \, P_w(\boldsymbol{w}^a) \biggr)
        \\
        & \times 
        \Biggl\{ \prod_{\mu \in \mathcal{I}_\mathrm{mem}}
        \int_\mathbb{R} \dd y^\mu \, P_\mathrm{mem}(y^\mu) \int_{\mathbb{R}^r} \Biggl( \, \prod_{a=1}^r \dd \lambda_\mu^a \, P_{\ell \theta}(y^\mu \,|\, \lambda_\mu^a) \, \Biggr) \, \mathcal{N}(\boldsymbol{\lambda}_\mu; \boldsymbol{B}, Q) \Biggr\}
        \\
        & \times \Biggl\{ \prod_{\mu \in \mathcal{I}_\mathrm{gen}} \int_\mathbb{R} \dd y^\mu \int_\mathbb{R} \dd \nu_\mu \, P_\mathrm{gen}(y^\mu \,|\, \nu_\mu) \int_{\mathbb{R}^r} \Biggl( \, \prod_{a=1}^r \dd \lambda_\mu^a \, P_\ell(y^\mu \,|\, \lambda_\mu^a) \Biggr) \, \mathcal{N}\!\left( \begin{pmatrix} \nu_\mu \\ \boldsymbol{\lambda}_\mu \end{pmatrix}; \begin{pmatrix} 0 \\ \boldsymbol{B} \end{pmatrix}, \Sigma \right) \Biggr\}.
    \end{split}
\end{equation}

\subsubsection*{Rewriting as a saddle point problem}
The next step is to disentangle the integrals over the labels and the previously introduced auxiliary variables from those over the teacher weights and the replicated student weights. Indeed, these integrals are coupled through the two Gaussian distributions appearing in Eq.~\eqref{eq: appendix - average replicated partition function before disentanglement}. To achieve this decoupling, we insert the following Dirac delta functions:
\begin{equation}
    \begin{split}
        1 &\varpropto \int_{\mathbb{R}^r} \left( \, \prod_{a=1}^{r} \dd b^a \right) \left( \, \prod_{a=1}^{r} \delta\!\left(b^a - \tfrac{1}{\sqrt{p}} \bw^a \cdot \mathbf{1}_p \right) \right) \int_\mathbb{R} \dd \rho \, \delta \bigl( d \rho - || \boldsymbol{w}_\star ||_2^2 \bigr)
        \\
        & \quad \times \int_{\mathbb{R}^r} \left( \, \prod_{a=1}^r \dd m_s^a \, \right) \left( \, \prod_{a=1}^r \delta \bigl( d m_s^a - \boldsymbol{w}_\star \cdot \boldsymbol{s}^a \bigr) \, \right)
        \\
        & \quad \times \int_{\mathbb{R}^{r(r+1)/2}} \left( \, \prod_{1 \leq a \leq b \leq r} \dd q_s^{ab} \, \right) \left( \, \prod_{1 \leq a \leq b \leq r} \delta \bigl( d q_s^{ab} - \boldsymbol{s}^a \cdot \boldsymbol{s}^b \bigr) \, \right)
        \\
        & \quad \times \int_{\mathbb{R}^{r(r+1)/2}} \left( \, \prod_{1 \leq a \leq b \leq r} \dd q_w^{ab} \, \right) \left( \, \prod_{1 \leq a \leq b \leq r} \delta \bigl( p q_w^{ab} - \boldsymbol{w}^a \cdot \boldsymbol{w}^b \bigr) \, \right)
        \\[6pt]
        &= \int_{\mathbb{R}^r} \left( \, \prod_{a=1}^{r} \dd b^a \right) \int_{\mathbb{R}^r} \left( \, \prod_{a=1}^{r} \frac{\dd \hat{b}^a}{2 \pi} \right) e^{i \sum\limits_{a = 1}^r \hat{b}^a \left( b^a - \frac{1}{\sqrt{p}} \boldsymbol{w}^a \cdot \mathbf{1}_p \right)} \int_\mathbb{R} \dd \rho \, \int_\mathbb{R} \frac{\dd \hat{\rho}}{2 \pi} e^{i \hat{\rho} ( d \rho - || \boldsymbol{w}_\star ||_2^2 )}
        \\
        & \quad \times \int_{\mathbb{R}^r} \left( \, \prod_{a=1}^r \dd m_s^a \, \right) \int_{\mathbb{R}^r} \left( \, \prod_{a=1}^r \frac{\dd \hat{m}_s^a}{2 \pi} \, \right) e^{i \sum\limits_{a=1}^r \hat{m}_s^a ( d m_s^a - \boldsymbol{w}_\star \cdot \boldsymbol{s}^a )}
        \\
        & \quad \times \int_{\mathbb{R}^{r(r+1)/2}} \left( \, \prod_{1 \leq a \leq b \leq r} \dd q_s^{ab} \, \right) \int_{\mathbb{R}^{r(r+1)/2}} \left( \, \prod_{1 \leq a \leq b \leq r} \frac{\dd \hat{q}_s^{ab}}{2 \pi} \, \right) e^{i \sum\limits_{1 \leq a \leq b \leq r} \hat{q}_s^{ab} ( d q_s^{ab} - \boldsymbol{s}^a \cdot \boldsymbol{s}^b )}
        \\
        & \quad \times \int_{\mathbb{R}^{r(r+1)/2}} \left( \, \prod_{1 \leq a \leq b \leq r} \dd q_w^{ab} \, \right) \int_{\mathbb{R}^{r(r+1)/2}} \left( \, \prod_{1 \leq a \leq b \leq r} \frac{\dd \hat{q}_w^{ab}}{2 \pi} \, \right) e^{i \sum\limits_{1 \leq a \leq b \leq r} \hat{q}_w^{ab} ( p q_w^{ab} - \boldsymbol{w}^a \cdot \boldsymbol{w}^b )}.
    \end{split}
\end{equation}
After introducing this in Eq.~\eqref{eq: appendix - average replicated partition function before disentanglement}, the averaged replicated partition function can be cast in a form amenable to saddle-point evaluation as $p \to +\infty$:
\begin{align} \label{eq: appendix - average replicated partition function after disentanglement}
    &\mathbb{E}_{\mathcal{D}, \varepsilon_n} [ \mathcal{Z}_{\beta, h}(\mathcal{D}, \varepsilon_n)^r ] = \mathbb{E}_{\varepsilon_n} \int_{\mathbb{R}^r} \left( \, \prod_{a=1}^{r} \dd b^a \right) \int_{(i \mathbb{R})^r} \left( \, \prod_{a=1}^{r} \frac{\dd \hat{b}^a}{2 \pi} \right) \int_\mathbb{R} \dd \rho \, \int_{i \mathbb{R}} \frac{\dd \hat{\rho}}{2 \pi} \nonumber
    \\
    & \times \int_{\mathbb{R}^r} \left( \, \prod_{a=1}^r \dd m_s^a \, \right) \int_{(i \mathbb{R})^r} \left( \, \prod_{a=1}^r \frac{\dd \hat{m}_s^a}{2 \pi} \, \right) \int_{\mathbb{R}^{r(r+1)/2}} \left( \, \prod_{1 \leq a \leq b \leq r} \dd q_s^{ab} \, \right) \int_{(i \mathbb{R})^{r(r+1)/2}} \left( \, \prod_{1 \leq a \leq b \leq r} \frac{\dd \hat{q}_s^{ab}}{2 \pi} \, \right) \nonumber
    \\
    & \times \int_{\mathbb{R}^{r(r+1)/2}} \left( \, \prod_{1 \leq a \leq b \leq r} \dd q_w^{ab} \, \right) \int_{(i \mathbb{R})^{r(r+1)/2}} \left( \, \prod_{1 \leq a \leq b \leq r} \frac{\dd \hat{q}_w^{ab}}{2 \pi} \, \right) \, e^{p \Phi^{(r)}(\mathcal{O}, \hat{\mathcal{O}}; \varepsilon_n)}.
\end{align}
Here, for compactness, we have collected the vector $\boldsymbol{b} = (b^1,\dots,b^r)^\top$, together with all the order parameters, into $O$. Similarly, the corresponding conjugate variables are collected into $\hat{\mathcal{O}}$. The potential $\Phi^{(r)}(\mathcal{O}, \hat{\mathcal{O}}; \varepsilon_n)$ is defined as
\begin{equation} \label{eq: appendix - potential}
    \begin{split}
        &\Phi^{(r)}(\mathcal{O}, \hat{\mathcal{O}}; \varepsilon_n) = - \frac{1}{p} \sum_{a=1}^r b^a \hat{b}^a - \frac{1}{\kappa} \rho \hat{\rho} - \frac{1}{\kappa} \sum_{a=1}^r m_s^a \hat{m}_s^a - \frac{1}{\kappa} \sum_{1 \leq a \leq b \leq r} q_s^{ab} \hat{q}_s^{ab} - \sum_{1 \leq a \leq b \leq r} q_w^{ab} \hat{q}_w^{ab}
        \\
        &
        \qquad\qquad\qquad\,\,\, + \Psi_w^{(r)}(\hat{\mathcal{O}}) + \frac{1}{\kappa} \varepsilon_n \alpha \Psi_\mathrm{mem}^{(r)}(\boldsymbol{B}, Q) + \frac{1}{\kappa} (1 - \varepsilon_n) \alpha \Psi_\mathrm{gen}^{(r)}(\boldsymbol{B}, \Sigma),
    \end{split}
\end{equation}
where
\begin{equation}
    \begin{split}
        \Psi_w^{(r)}(\hat{\mathcal{O}}) &= \frac{1}{p} \log \int_{\mathbb{R}^d} \dd \boldsymbol{w}_\star \, P_{\boldsymbol{w}_\star}(\boldsymbol{w}_\star) \, e^{\hat{\rho} || \boldsymbol{w}_\star ||_2^2} \int_{(\mathbb{R}^p)^r} \left( \, \prod_{a=1}^r \dd \boldsymbol{w}^a \, P_{\boldsymbol{w}}(\boldsymbol{w}^a) \right)
        \\
        & \quad \times e^{\frac{1}{\sqrt{p}} \sum\limits_{a=1}^r \hat{b}^a \boldsymbol{w}^a \cdot \mathbf{1}_p + \sum\limits_{a=1}^r \hat{m}_s^a \boldsymbol{w}_\star \cdot \boldsymbol{s}^a + \sum\limits_{1 \leq a \leq b \leq r} \hat{q}_s^{ab} \boldsymbol{s}^a \cdot \boldsymbol{s}^b + \sum\limits_{1 \leq a \leq b \leq r} \hat{q}_w^{ab} \boldsymbol{w}^a \cdot \boldsymbol{w}^b},
    \end{split}
\end{equation}
\begin{equation}
    \begin{split}
        \Psi_\mathrm{mem}^{(r)}(\boldsymbol{B}, Q) &=
        \log \int_\mathbb{R} \dd y \, P_\mathrm{mem}(y) \int_{\mathbb{R}^r} \left( \, \prod_{a=1}^r \dd \lambda^a \, P_{\ell \theta}(y \,|\, \lambda^a) \right) \, \mathcal{N}(\boldsymbol{\lambda}; \boldsymbol{B}, Q),
    \end{split}
\end{equation}
\begin{equation}
    \begin{split}
        \Psi_\mathrm{gen}^{(r)}(\boldsymbol{B}, \Sigma) &=
        \log \int_\mathbb{R} \dd y \int_\mathbb{R} \dd \nu \, P_\mathrm{gen}(y \,|\, \nu) \int_{\mathbb{R}^r} \left( \, \prod_{a=1}^r \dd \lambda^a \, P_{\ell}(y | \lambda^a) \right) \, \mathcal{N}\!\left( \begin{pmatrix} \nu \\ \boldsymbol{\lambda} \end{pmatrix}; \begin{pmatrix} 0 \\ \boldsymbol{B} \end{pmatrix}, \Sigma \right).
    \end{split}
\end{equation}
After the disentanglement, the mean vector $\boldsymbol{B}$ has components
\begin{equation}
    B^a = \mu_0 b^a \qquad a=1,\dots,r,
\end{equation}
while the entries of the covariance matrices $Q$ and $\Sigma$ are given by
\begin{equation}
    \begin{split}
        &\Sigma^{00} = \rho,
        \\
        &\Sigma^{0a} = \mu_1 m_s^a \qquad a=1,\dots,r,
        \\
        &\Sigma^{ab} = Q^{ab} = \mu_\star^2 q_w^{ab} + \mu_1^2 q_s^{ab} \qquad 1 \leq a \leq b \leq r.
    \end{split}
\end{equation}

\subsubsection*{The replica-symmetric ansatz}
Applying the saddle-point method, the computation of the average replicated partition function reduces to an extremization problem over $\mathcal{O}$ and $\hat{\mathcal{O}}$, apart from the remaining average over $\varepsilon_n$. At this stage, the free energy density \eqref{eq: appendix - free energy density with field replica trick} can be written as
\begin{equation}
    f_{\beta, h} = - \frac{1}{\beta} \lim_{r \to 0^+} \lim_{p \to + \infty} \frac{1}{p} \frac{\mathbb{E}_{\varepsilon_n}\!\Biggl[ \exp\biggl\{ p \operatorname*{extr}\limits_{\mathcal{O}, \hat{\mathcal{O}}} \Phi^{(r)}(\mathcal{O}, \hat{\mathcal{O}}; \varepsilon_n) \biggr\} \Biggr] - 1}{r}.
\end{equation}

To proceed with the limit $r \to 0^+$, we restrict the extremization to the replica-symmetric (RS) ansatz:
\begin{equation} \label{eq: appendix - replica symmetric ansatz}
    \begin{aligned}
        b^a            &= b,               & \quad   \hat{b}^a            &= \hat{b},          &\text{for } a = 1,\dots,r, \\
        m^a_s          &= m_s,            & \quad \hat{m}^a_s          &= \hat{m}_s,          &\text{for } a = 1,\dots,r, \\
        q^{aa}_{s/w}   &= Q_{s/w},        & \quad \hat{q}^{aa}_{s/w}   &= -\tfrac{1}{2} \hat{Q}_{s/w}, &\text{for } a = 1,\dots,r, \\
        q^{ab}_{s/w}   &= q_{s/w},        & \quad \hat{q}^{ab}_{s/w}   &= \hat{q}_{s/w}, &\text{for } 1 \leq a < b \leq r.
    \end{aligned}
\end{equation}
Under this ansatz, we obtain
\begin{equation} \label{eq: appendix - free energy density replica symmetric}
    f_{\beta, h} = - \frac{1}{\beta} \lim_{r \to 0^+} \lim_{p \to + \infty} \frac{1}{r} \, \mathbb{E}_{\varepsilon_n}\!\left[ \, \operatorname*{extr}_{\mathcal{O}, \hat{\mathcal{O}}} \Phi^{(r)}(\mathcal{O}, \hat{\mathcal{O}}; \varepsilon_n) \, \right] = - \frac{1}{\beta} \lim_{r \to 0^+} \lim_{p \to + \infty} \frac{1}{r} \operatorname*{extr}_{\mathcal{O}, \hat{\mathcal{O}}} \Phi^{(r)}(\mathcal{O}, \hat{\mathcal{O}}; \varepsilon),
\end{equation}
where we have used that $\varepsilon_n$ converges to $\varepsilon$ in the high-dimensional limit. The first equality in \eqref{eq: appendix - free energy density replica symmetric} holds provided that, under the RS ansatz, $\lim_{r \to 0^+} \Phi^{(r)}(\mathcal{O}, \hat{\mathcal{O}}; \varepsilon_n) = 0$.

Substituting the RS ansatz \eqref{eq: appendix - replica symmetric ansatz} into Eq.~\eqref{eq: appendix - potential} and evaluating the resulting expression in the limit $r \to 0^+$, the consistency condition fixes $\rho = \mathbb{E}_{w_\star \sim \mathcal{N}(0,1)} \bigl[ w_\star^2 \bigr] = 1$ and $\hat{\rho} = 0$.

We can then take the limits $p \to +\infty$ and $r \to 0^+$ term by term in $\tfrac{1}{r} \Phi^{(r)}(\mathcal{O}, \hat{\mathcal{O}}; \varepsilon)$ under the RS ansatz. Below we report only the final result, since the derivation follows the same steps as those detailed in Appendix C of \cite{gerace2020generalisation}.

\subsubsection*{The replica-symmetric free energy density $f_{\beta, h}$}
Under the RS ansatz, the free energy density $f_{\beta, h}$ takes the form
\begin{equation} \label{eq: appendix - free energy density replica symmetric f_beta_h (full expression)}
    \begin{split}
        f_{\beta, h} &= - \frac{1}{\beta} \operatorname*{extr}_{\substack{b, m_s, q_s, V_s, q_w, V_w \\ \hat{b}, \hat{m}_s, \hat{q}_s, \hat{V}_s, \hat{q}_w, \hat{V}_w}} \biggl\{ -\frac{1}{\kappa} m_s \hat{m}_s + \frac{1}{2 \kappa} V_s \hat{V}_s + \frac{1}{2 \kappa} (q_s \hat{V}_s - \hat{q}_s V_s) + \frac{1}{2} V_w \hat{V}_w + \frac{1}{2} (q_w \hat{V}_w - \hat{q}_w V_w)
        \\
        & \quad + \psi_w(\hat{b}, \hat{m}_s, \hat{q}_s, \hat{V}_s, \hat{q}_w, \hat{V}_w) + \frac{1}{\kappa} \varepsilon \alpha \psi_\mathrm{mem}(B, q, V) + \frac{1}{\kappa} (1 - \varepsilon) \alpha \psi_\mathrm{gen}(B, m, q, V) \biggr\}.
    \end{split}
\end{equation}
In Eq.~\eqref{eq: appendix - free energy density replica symmetric f_beta_h (full expression)}, we have introduced $V_{s/w} = Q_{s/w} - q_{s/w}$ and $\hat{V}_{s/w} = \hat{Q}_{s/w} + \hat{q}_{s/w}$. We have also defined $B = \mu_0 b$, $m = \mu_1 m_s$, $q = \mu_1^2 q_s + \mu_\star^2 q_w$ and $V = \mu_1^2 V_s + \mu_\star^2 V_w$. The entropic term $\psi_w$ and the energetic terms $\psi_\mathrm{mem}$ and $\psi_\mathrm{gen}$ are obtained by taking the limits $p \to +\infty$ and $r \to 0^+$ of $\frac{1}{r} \Psi_w^{(r)}$, $\frac{1}{r} \Psi_\mathrm{mem}^{(r)}$ and $\frac{1}{r} \Psi_\mathrm{gen}^{(r)}$, respectively, under the RS ansatz:
\begin{equation} \label{eq: appendix - entropic term in the replica symmetric ansatz}
\begin{split}
    \psi_w(\hat{b}, \hat{m}_s, \hat{q}_s, \hat{V}_s, \hat{q}_w, \hat{V}_w) &= \lim_{p \to +\infty} \frac{1}{p} \, \mathbb{E}_{\boldsymbol{\xi} \sim \mathcal{N}(\boldsymbol{0}, \mathbb{I}_d), \boldsymbol{\eta} \sim \mathcal{N}(\boldsymbol{0}, \mathbb{I}_p)} \Biggl[ \int_{\mathbb{R}^d} \dd \boldsymbol{w}_\star \, P_{\boldsymbol{w}_\star}(\boldsymbol{w}_\star) \log \int_{\mathbb{R}^p} \dd \boldsymbol{w} \, P_{\boldsymbol{w}}(\boldsymbol{w})
    \\
    &\quad \times e^{- \frac{\hat{V}_w}{2} ||\boldsymbol{w}||_2^2 - \frac{\hat{V}_s}{2 p} ||F \boldsymbol{w}||_2^2 + \left( \sqrt{\hat{q}_w} \boldsymbol{\eta} + \frac{1}{\sqrt{p}} \hat{b} \mathbf{1}_p \right) \cdot \boldsymbol{w} + \left( \hat{m}_s \boldsymbol{w}_\star + \sqrt{\hat{q}_s} \boldsymbol{\xi} \right) \cdot \frac{1}{\sqrt{p}} F \boldsymbol{w} } \Biggr],
\end{split}
\end{equation}
\begin{equation} \label{eq: appendix - mem energetic term in the replica symmetric ansatz}
    \psi_\mathrm{mem}(B, q, V) = \mathbb{E}_{\xi \sim \mathcal{N}(0,1)} \left[ \, \int_\mathbb{R} \dd y \, P_\mathrm{mem}(y) \log \int_\mathbb{R} \dd \lambda \, P_{\ell \theta}(y \,|\, \lambda) \, \mathcal{N}\!\left( \lambda; B + \sqrt{q} \xi, V \right) \, \right],
\end{equation}
\begin{equation} \label{eq: appendix - gen energetic term in the replica symmetric ansatz}
\begin{split}
    \psi_\mathrm{gen}(B, m, q, V) &= \mathbb{E}_{\xi \sim \mathcal{N}(0,1)} \Biggl[ \, \int_\mathbb{R} \dd y \int_\mathbb{R} \dd \nu \, P_\mathrm{gen}(y \,|\, \nu) \, \mathcal{N}\!\left( \nu; \frac{m}{\sqrt{q}} \xi, 1 - \frac{m^2}{q} \right)
    \\
    &\qquad \qquad \qquad \qquad \times \log \int_\mathbb{R} \dd \lambda \, P_{\ell}(y \,|\, \lambda) \, \mathcal{N}\!\left( \lambda; B + \sqrt{q} \xi, V \right) \, \Biggr].
\end{split}
\end{equation}
Notice that the term proportional to $b \hat{b}$ in Eq.~\eqref{eq: appendix - potential} contributes only at subleading order to the free energy density and is therefore omitted from the limiting expression.

Setting $h = 0$ yields the RS version of Eq.~\eqref{eq: appendix - free energy density}. In this case, one has $P_{\ell \theta}(y \,|\, z) = P_\ell(y \,|\, z)$, and the terms $\psi_\mathrm{mem}$ and $\psi_\mathrm{gen}$ can therefore be combined into a single energetic contribution, which we denote by $\psi_y$ and define as follows:
\begin{equation}
    \psi_y(B, m, q, V) = \mathbb{E}_{\xi \sim \mathcal{N}(0,1)} \left[ \, \int_\mathbb{R} \dd y \, \mathcal{Z}_\mathrm{out^\star}\!\left(y, \frac{m}{\sqrt{q}} \xi, 1 - \frac{m^2}{q} \right) \log \mathcal{Z}_\mathrm{out}(y, B + \sqrt{q} \xi, V) \, \right],
\end{equation}
where
\begin{align}
\label{eq: appendix - output channel partition function}
\mathcal{Z}_{\mathrm{out}^\star}(y, \omega, \tau) &= \mathbb{E}_{z \sim \mathcal{N}(\omega, \tau)} \left[ P_{\mathrm{out}^\star}(y \,|\, z) \right],
\\
\label{eq: appendix - output ERM partition function}
\mathcal{Z}_{\mathrm{out}}(y, \omega, \tau)
&= \mathbb{E}_{z \sim \mathcal{N}(\omega, \tau)} \left[ P_{\ell}(y \,|\, z) \right].
\end{align}
Taking this into account, for $h = 0$ the RS free energy density \eqref{eq: appendix - free energy density replica symmetric f_beta_h (full expression)} becomes
\begin{equation} \label{eq: appendix - free energy density replica symmetric f_beta_0}
    \begin{split}
        f_{\beta, 0} &= - \frac{1}{\beta} \operatorname*{extr}_{\substack{b, m_s, q_s, V_s, q_w, V_w \\ \hat{b}, \hat{m}_s, \hat{q}_s, \hat{V}_s, \hat{q}_w, \hat{V}_w}} \biggl\{ - \frac{1}{\kappa} m_s \hat{m}_s + \frac{1}{2 \kappa} V_s \hat{V}_s + \frac{1}{2 \kappa} (q_s \hat{V}_s - \hat{q}_s V_s) + \frac{1}{2} V_w \hat{V}_w + \frac{1}{2} (q_w \hat{V}_w - \hat{q}_w V_w)
        \\
        & \quad + \psi_w(\hat{b}, \hat{m}_s, \hat{q}_s, \hat{V}_s, \hat{q}_w, \hat{V}_w) + \frac{1}{\kappa} \alpha \psi_y(B, m, q, V) \biggr\}.
    \end{split}
\end{equation}
Eq.~\eqref{eq: appendix - free energy density replica symmetric f_beta_0} is actually valid for any choice of the output distribution $P_\mathrm{out^\star}$. By contrast, Eq.~\eqref{eq: appendix - free energy density replica symmetric f_beta_h (full expression)} applies only when the output distribution has the specific form introduced in Eq.~\eqref{eq: appendix - RAF output channel}.

\subsubsection*{Recovering the RS free energy density of \cite{gerace2020generalisation}}
When $\mu_0 = \mathbb{E}_{g \sim \mathcal{N}(0,1)}\!\left[ \sigma(g) \right] = 0$, the computation follows the same steps as above, with the simplification that the replicated Gaussian equivalence yields a Gaussian distribution with zero mean. Consequently, no additional auxiliary variables are needed beyond those associated with the overlap order parameters. Using the notation of the previous subsection, the resulting RS free energy density $f_{\beta, h}$ reads
\begin{equation} \label{eq: appendix - free energy density replica symmetric f_beta_h mu_0=0}
    \begin{split}
        f_{\beta, h} &= - \frac{1}{\beta} \operatorname*{extr}_{\substack{m_s, q_s, V_s, q_w, V_w \\ \hat{m}_s, \hat{q}_s, \hat{V}_s, \hat{q}_w, \hat{V}_w}} \biggl\{ -\frac{1}{\kappa} m_s \hat{m}_s + \frac{1}{2 \kappa} V_s \hat{V}_s + \frac{1}{2 \kappa} (q_s \hat{V}_s - \hat{q}_s V_s) + \frac{1}{2} V_w \hat{V}_w + \frac{1}{2} (q_w \hat{V}_w - \hat{q}_w V_w)
        \\
        & \quad + \psi_w(\hat{m}_s, \hat{q}_s, \hat{V}_s, \hat{q}_w, \hat{V}_w) + \frac{1}{\kappa} \varepsilon \alpha \psi_\mathrm{mem}(q, V) + \frac{1}{\kappa} (1 - \varepsilon) \alpha \psi_\mathrm{gen}(m, q, V) \biggr\},
    \end{split}
\end{equation}
where
\begin{equation}
\begin{split}
    \psi_w(\hat{m}_s, \hat{q}_s, \hat{V}_s, \hat{q}_w, \hat{V}_w) &= \lim_{p \to +\infty} \frac{1}{p} \, \mathbb{E}_{\boldsymbol{\xi} \sim \mathcal{N}(\boldsymbol{0}, \mathbb{I}_d), \boldsymbol{\eta} \sim \mathcal{N}(\boldsymbol{0}, \mathbb{I}_p)} \Biggl[ \int_{\mathbb{R}^d} \dd \boldsymbol{w}_\star \, P_{\boldsymbol{w}_\star}(\boldsymbol{w}_\star) \log \int_{\mathbb{R}^p} \dd \boldsymbol{w} \, P_{\boldsymbol{w}}(\boldsymbol{w})
    \\
    &\quad \times e^{- \frac{\hat{V}_w}{2} ||\boldsymbol{w}||_2^2 - \frac{\hat{V}_s}{2 p} ||F \boldsymbol{w}||_2^2 + \sqrt{\hat{q}_w} \boldsymbol{\eta} \cdot \boldsymbol{w} + \left( \hat{m}_s \boldsymbol{w}_\star + \sqrt{\hat{q}_s} \boldsymbol{\xi} \right) \cdot \frac{1}{\sqrt{p}} F \boldsymbol{w} } \Biggr],
\end{split}
\end{equation}
\begin{equation}
    \psi_\mathrm{mem}(q, V) = \mathbb{E}_{\xi \sim \mathcal{N}(0,1)} \left[ \, \int_\mathbb{R} \dd y \, P_\mathrm{mem}(y) \log \int_\mathbb{R} \dd \lambda \, P_{\ell \theta}(y \,|\, \lambda) \, \mathcal{N}\!\left( \lambda; \sqrt{q} \xi, V \right) \, \right],
\end{equation}
\begin{equation}
\begin{split}
    \psi_\mathrm{gen}(m, q, V) &= \mathbb{E}_{\xi \sim \mathcal{N}(0,1)} \Biggl[ \, \int_\mathbb{R} \dd y \int_\mathbb{R} \dd \nu \, P_\mathrm{gen}(y \,|\, \nu) \, \mathcal{N}\!\left( \nu; \frac{m}{\sqrt{q}} \xi, 1 - \frac{m^2}{q} \right)
    \\
    &\qquad \qquad \qquad \qquad \times \log \int_\mathbb{R} \dd \lambda \, P_{\ell}(y \,|\, \lambda) \, \mathcal{N}\!\left( \lambda; \sqrt{q} \xi, V \right) \, \Biggr].
\end{split}
\end{equation}
Setting $h = 0$ in the expression above gives
\begin{equation} \label{eq: appendix - free energy density replica symmetric (Gerace et al.)}
    \begin{split}
        f_{\beta, 0} &= - \frac{1}{\beta} \operatorname*{extr}_{\substack{m_s, q_s, V_s, q_w, V_w \\ \hat{m}_s, \hat{q}_s, \hat{V}_s, \hat{q}_w, \hat{V}_w}} \biggl\{ - \frac{1}{\kappa} m_s \hat{m}_s + \frac{1}{2 \kappa} V_s \hat{V}_s + \frac{1}{2 \kappa} (q_s \hat{V}_s - \hat{q}_s V_s) + \frac{1}{2} V_w \hat{V}_w + \frac{1}{2} (q_w \hat{V}_w - \hat{q}_w V_w)
        \\
        & \quad + \psi_w(\hat{m}_s, \hat{q}_s, \hat{V}_s, \hat{q}_w, \hat{V}_w) + \frac{1}{\kappa} \alpha \psi_y(m, q, V) \biggr\}.
    \end{split}
\end{equation}
where
\begin{equation}
    \psi_y(m, q, V) = \mathbb{E}_{\xi \sim \mathcal{N}(0,1)} \left[ \, \int_\mathbb{R} \dd y \, \mathcal{Z}_\mathrm{out^\star}\!\left(y, \frac{m}{\sqrt{q}} \xi, 1 - \frac{m^2}{q} \right) \log \mathcal{Z}_\mathrm{out}(y, \sqrt{q} \xi, V) \, \right],
\end{equation}
Eq.~\eqref{eq: appendix - free energy density replica symmetric (Gerace et al.)} coincides exactly with the RS free energy density derived in \cite{gerace2020generalisation}.

\subsubsection*{Evaluating the entropic term for ridge regularization and Gaussian teacher prior}
To complete the replica computation, we now evaluate the entropic term $\psi_w$ in Eq.~\eqref{eq: appendix - entropic term in the replica symmetric ansatz} for the ridge prior
\begin{equation}
    P_{\boldsymbol{w}}(\boldsymbol{w}) = e^{-\frac{\beta \lambda}{2} ||\boldsymbol{w}||_2^2}
\end{equation}
and the Gaussian teacher prior $P_{\boldsymbol{w}_\star} = \mathcal{N}(\boldsymbol{0}, \mathbb{I}_d)$, which are the ones considered in this work.

To write the integral in matrix form, we introduce
\begin{equation}
    a = \beta \lambda + \hat{V}_w,
    \qquad
    t = \frac{\hat{V}_s}{a},
    \qquad
    \boldsymbol{u} = \sqrt{\hat{q}_s} \boldsymbol{\xi} + \hat{m}_s \boldsymbol{w}_\star,
\end{equation}
together with the matrices
\begin{equation}
    K_p = \mathbb{I}_p + t \frac{F^\top F}{p},
    \qquad
    G_d = \mathbb{I}_d + t \frac{F F^\top}{p}.
\end{equation}
The exponent in the integral over $\boldsymbol{w}$ can then be expressed as
\begin{equation}
\begin{split}
    &- \frac{\beta \lambda + \hat{V}_w}{2} ||\boldsymbol{w}||_2^2 - \frac{\hat{V}_s}{2 p} ||F \boldsymbol{w}||_2^2 + \left( \sqrt{\hat{q}_w} \boldsymbol{\eta} + \frac{1}{\sqrt{p}} \hat{b} \mathbf{1}_p \right) \cdot \boldsymbol{w} + \left( \hat{m}_s \boldsymbol{w}_\star + \sqrt{\hat{q}_s} \boldsymbol{\xi} \right) \cdot \frac{1}{\sqrt{p}} F \boldsymbol{w} =
    \\
    &= - \frac{1}{2} \boldsymbol{w}^\top A_p \boldsymbol{w} + \boldsymbol{h}^\top \boldsymbol{w},
\end{split}
\end{equation}
where
\begin{equation}
    A_p = a K_p,
    \qquad
    \boldsymbol{h} = \sqrt{\hat{q}_w} \boldsymbol{\eta} + \frac{F^\top \boldsymbol{u}}{\sqrt{p}} + \frac{1}{\sqrt{p}} \hat{b} \mathbf{1}_p.
\end{equation}
Therefore, at finite $p$ and $d$, the Gaussian integral over $\boldsymbol{w}$ yields
\begin{equation}
    \psi_{w,p} = \frac{1}{2} \log(2 \pi) - \frac{1}{2p} \log \det A_p + \frac{1}{2p} \mathbb{E}\!\left[ \boldsymbol{h}^\top A_p^{-1} \boldsymbol{h} \right],
\end{equation}
where the remaining expectation is with respect to $\boldsymbol{\eta}$, $\boldsymbol{\xi}$ and $\boldsymbol{w}_\star$. Since $\boldsymbol{\eta}$, $\boldsymbol{\xi}$ and $\boldsymbol{w}_\star$ are mutually independent and centered, all mixed terms in $\boldsymbol{h}^\top A_p^{-1} \boldsymbol{h}$ vanish upon averaging. Moreover,
\begin{equation}
    \mathbb{E}\!\left[ \boldsymbol{u} \boldsymbol{u}^\top \right] = (\hat{q}_s + \hat{m}_s^2) \mathbb{I}_d.
\end{equation}
Using $A_p^{-1} = a^{-1} K_p^{-1}$, together with the push-through identity
\begin{equation}
    K_p^{-1} F^\top = F^\top G_d^{-1},
\end{equation}
we thus obtain
\begin{equation}
    \psi_{w, p} = \frac{1}{2} \log(2 \pi) - \frac{1}{2} \log a - \frac{1}{2p} \operatorname{Tr} \log G_d + \frac{\hat{q}_w}{2 a p} \operatorname{Tr} K_p^{-1} + \frac{\hat{m}_s^2 + \hat{q}_s}{2 a p} \operatorname{Tr}\!\left[ \frac{FF^\top}{p} G_d^{-1} \right] + \frac{\hat{b}^2}{2 a p^2} \mathbf{1}_p^\top K_p^{-1} \mathbf{1}_p.
\end{equation}
Here we have also used Sylvester’s determinant identity, which gives $\det K_p = \det G_d$. The two trace contractions are reduced to $d$-dimensional resolvent traces by the identities
\begin{equation}
    \operatorname{Tr} K_p^{-1} = p - d + \operatorname{Tr} G_d^{-1}
\end{equation}
and
\begin{equation}
    \frac{1}{a} \operatorname{Tr}\!\left[ \frac{FF^\top}{p} G_d^{-1} \right] = \frac{1}{\hat{V}_s} (d - \operatorname{Tr} G_d^{-1}).
\end{equation}

The last term in $\psi_{w, p}$, which originates from the conjugate bias $\hat{b}$, must be treated separately. Assuming that $||K_p^{-1}||_\mathrm{op} = O(1)$ and $\hat{b} = O(1)$,
\begin{equation}
    \left| \frac{\hat{b}^2}{2 a p^2} \mathbf{1}_p^\top K_p^{-1} \mathbf{1}_p \right| \leq \frac{\hat{b}^2}{2 a p^2} || K_p^{-1}||_\mathrm{op} \, ||\mathbf{1}_p||_2^2 = \frac{\hat{b}^2}{2 a p} || K_p^{-1}||_\mathrm{op} = O(p^{-1}).
\end{equation}
Consequently, although this contribution is present at finite $p$, it vanishes in the high-dimensional limit. The leading-order entropic term is therefore independent of $\hat{b}$.

To evaluate the remaining traces, assume that $FF^\top / p$ has a well defined spectral distribution $\mu$ as $p, d \to +\infty$ with $\kappa = p /d$ finite, and define its Stieltjes transform by
\begin{equation}
    g(z) = \int_\mathbb{R} \frac{\dd \mu(x)}{x - z}.
\end{equation}
Then, the resolvent trace satisfies
\begin{equation}
    \lim_{p \to +\infty} \frac{1}{p} \operatorname{Tr} G_d^{-1} = \frac{1}{\kappa} \frac{a}{\hat{V}_s} g\left( - \frac{a}{\hat{V}_s} \right)
\end{equation}
and the logarithmic trace may be rewritten as
\begin{equation}
    \lim_{p \to +\infty} \frac{1}{p} \operatorname{Tr} \log G_d = \frac{1}{\kappa} \int_{-\infty}^{-a/\hat{V}_s} \dd z \, \left[ - \frac{1}{z} - g(z) \right].
\end{equation}
Substituting the two identities above into the expression for $\psi_w = \lim_{p \to +\infty} \psi_{w, p}$, and discarding the constant and subleading terms, we obtain
\begin{equation} \label{eq: appendix - entropic term for ridge student prior and gaussian teacher prior}
\begin{split}
    \psi_w(\hat{b}, \hat{m}_s, \hat{q}_s, \hat{V}_s, \hat{q}_w, \hat{V}_w) &= - \frac{1}{2} \log\!\left(\beta \lambda + \hat{V}_w \right) - \frac{1}{2 \kappa} \int_{-\infty}^{-\frac{\beta \lambda + \hat{V}_w}{\hat{V}_s}} \dd z \, \left[ -\frac{1}{z} - g(z) \right]
    \\
    &\qquad + \frac{1}{\kappa} \frac{\hat{m}_s^2 + \hat{q}_s}{2 \hat{V}_s} \left[ 1 - \frac{\beta \lambda + \hat{V}_w}{\hat{V}_s} g\!\left( - \frac{\beta \lambda + \hat{V}_w}{\hat{V}_s} \right) \right]
    \\
    &\qquad + \frac{\hat{q}_w}{2 (\beta \lambda + \hat{V}_w)} \left[ 1 - \frac{1}{\kappa} + \frac{1}{\kappa} \frac{\beta \lambda + \hat{V}_w}{\hat{V}_s} g\!\left( - \frac{\beta \lambda + \hat{V}_w}{\hat{V}_s} \right) \right].
\end{split}
\end{equation}

In the setting considered here, $F$ is a Gaussian projection matrix. Consequently, its limiting spectral distribution is the Marchenko–Pastur law with aspect ratio $d / p = \kappa^{-1}$. The corresponding Stieltjes transform, for $z < 0$, is
\begin{equation} \label{eq: appendix - Stieltjes Gaussian projection}
    g(z) = \frac{\kappa}{2 z} \left[ 1 - z - \frac{1}{\kappa} - \sqrt{\left( z - 1 - \frac{1}{\kappa} \right)^2 - \frac{4}{\kappa}} \, \right].
\end{equation}

\subsection{Random features regression state equations}
\label{app:eqs_random_features}
In the setting considered in this work, Eq.~\eqref{eq: appendix - free energy density replica symmetric f_beta_h (full expression)} takes the form
\begin{equation} \label{eq: appendix - f_beta_h mu0 not zero for Gaussian priors}
    \begin{split}
        &f_{\beta, h} = - \frac{1}{\beta} \operatorname*{extr}_{\substack{B, m_s, q_s, V_s, q_w, V_w \\ \hat{m}_s, \hat{q}_s, \hat{V}_s, \hat{q}_w, \hat{V}_w}} \Biggl\{ -\frac{1}{\kappa} m_s \hat{m}_s + \frac{1}{2 \kappa} V_s \hat{V}_s + \frac{1}{2 \kappa} (q_s \hat{V}_s - \hat{q}_s V_s) + \frac{1}{2} V_w \hat{V}_w + \frac{1}{2} (q_w \hat{V}_w - \hat{q}_w V_w)
        \\
        & + - \frac{1}{2} \log\!\left(\beta \lambda + \hat{V}_w \right) - \frac{1}{2 \kappa} \int_{-\infty}^{-\frac{\beta \lambda + \hat{V}_w}{\hat{V}_s}} \dd z \, \left[ -\frac{1}{z} - g(z) \right] + \frac{1}{\kappa} \frac{\hat{m}_s^2 + \hat{q}_s}{2 \hat{V}_s} \left[ 1 - \frac{\beta \lambda + \hat{V}_w}{\hat{V}_s} g\!\left( - \frac{\beta \lambda + \hat{V}_w}{\hat{V}_s} \right) \right]
        \\
        & + \frac{\hat{q}_w}{2 (\beta \lambda + \hat{V}_w)} \left[ 1 - \frac{1}{\kappa} + \frac{1}{\kappa} \frac{\beta \lambda + \hat{V}_w}{\hat{V}_s} g\!\left( - \frac{\beta \lambda + \hat{V}_w}{\hat{V}_s} \right) \right] + \frac{\varepsilon \alpha}{\kappa} \psi_\mathrm{mem}(B, q, V) + \frac{(1 - \varepsilon) \alpha}{\kappa} \psi_\mathrm{gen}(B, m, q, V) \Biggr\},
    \end{split}
\end{equation}
where $g$ is given by Eq.~\eqref{eq: appendix - Stieltjes Gaussian projection}, while $\psi_\mathrm{mem}$ and $\psi_\mathrm{gen}$ are defined in Eqs.~\eqref{eq: appendix - mem energetic term in the replica symmetric ansatz}--\eqref{eq: appendix - gen energetic term in the replica symmetric ansatz}. Since $b$ enters the free energy only through $B$, the extremization over $b$ can be equivalently replaced by an extremization over $B$. Moreover, the extremization over $\hat{b}$ is absent because the entropic term in Eq.~\eqref{eq: appendix - entropic term for ridge student prior and gaussian teacher prior} is independent of $\hat{b}$.

Correspondingly, Eq.~\eqref{eq: appendix - free energy density replica symmetric f_beta_0} becomes
\begin{equation}
    \begin{split}
        &f_{\beta, 0} = - \frac{1}{\beta} \operatorname*{extr}_{\substack{B, m_s, q_s, V_s, q_w, V_w \\ \hat{m}_s, \hat{q}_s, \hat{V}_s, \hat{q}_w, \hat{V}_w}} \Biggl\{ -\frac{1}{\kappa} m_s \hat{m}_s + \frac{1}{2 \kappa} V_s \hat{V}_s + \frac{1}{2 \kappa} (q_s \hat{V}_s - \hat{q}_s V_s) + \frac{1}{2} V_w \hat{V}_w + \frac{1}{2} (q_w \hat{V}_w - \hat{q}_w V_w)
        \\
        & + - \frac{1}{2} \log\!\left(\beta \lambda + \hat{V}_w \right) - \frac{1}{2 \kappa} \int_{-\infty}^{-\frac{\beta \lambda + \hat{V}_w}{\hat{V}_s}} \dd z \, \left[ -\frac{1}{z} - g(z) \right] + \frac{1}{\kappa} \frac{\hat{m}_s^2 + \hat{q}_s}{2 \hat{V}_s} \left[ 1 - \frac{\beta \lambda + \hat{V}_w}{\hat{V}_s} g\!\left( - \frac{\beta \lambda + \hat{V}_w}{\hat{V}_s} \right) \right]
        \\
        & + \frac{\hat{q}_w}{2 (\beta \lambda + \hat{V}_w)} \left[ 1 - \frac{1}{\kappa} + \frac{1}{\kappa} \frac{\beta \lambda + \hat{V}_w}{\hat{V}_s} g\!\left( - \frac{\beta \lambda + \hat{V}_w}{\hat{V}_s} \right) \right] + \frac{\alpha}{\kappa} \psi_y(B, m, q, V) \Biggr\}
    \end{split}
\end{equation}
Extremizing this free-energy density yields a system of self-consistent saddle-point equations valid at any finite $\beta$. In the zero-temperature limit $\beta \to +\infty$, their solution determines the order parameters characterizing the global minimizer of the empirical risk in the high-dimensional limit.

To take this limit, we introduce the following rescaling of the overlaps and their conjugates:
\begin{equation} \label{eq: appendix - rescaling for zero temperature}
\begin{aligned}
    m_s &\to m_s, &\qquad\qquad q_{s/w} &\to q_{s/w}, &\qquad\qquad V_{s/w} &\to \beta^{-1} V_{s/w},
    \\
    \hat{m}_s &\to \beta \hat{m}_s, &\qquad\qquad \hat{q}_{s/w} &\to \beta^2 \hat{q}_{s/w}, &\qquad\qquad \hat{V}_{s/w} &\to \beta \hat{V}_{s/w}.
\end{aligned}
\end{equation}
The order parameter $B$ is left unchanged. The explicit limiting procedure follows the steps detailed in Appendix D of \cite{gerace2020generalisation}; here, we report the resulting equations directly. In the limit $\beta \to +\infty$, extremization with respect to the conjugate variables $\hat{m}_s$, $\hat{q}_s$, $\hat{V}_s$, $\hat{q}_w$ and $\hat{V}_w$ leads to
\begin{equation} \label{eq: appendix - SE random features overlaps}
\left\{
\begin{aligned}
&m_s = \frac{\hat{m}_s}{\hat{V}_s} \frac{z + 1 + \kappa^{-1} - \Delta}{2 \kappa^{-1}} \\[4pt]
&q_s = \frac{\hat{m}_s^2 + \hat{q}_s}{\hat{V}_s^2} \frac{(2z + \kappa^{-1} + 1) \Delta - 2 z^2 - 3 (\kappa^{-1} + 1) z - (\kappa^{-1} - 1)^2}{2 \kappa^{-1} \Delta} \\&\qquad\, - \frac{\hat{q}_w}{(\lambda + \hat{V}_w) \hat{V}_s} \frac{z \Delta - z^2 - (\kappa^{-1} + 1) z}{2 \kappa^{-1} \Delta} \\[4pt]
&V_s = \frac{1}{\hat{V}_s} \frac{z + 1 + \kappa^{-1} - \Delta}{2 \kappa^{-1}} \\[4pt]
&q_w = \frac{\hat{q}_w}{(\lambda + \hat{V}_w)^2} \frac{(1 - \kappa^{-1}) \Delta + (\kappa^{-1} + 1) z + (\kappa^{-1} - 1)^2}{2 \Delta} - \frac{\hat{m}_s^2 + \hat{q}_s}{(\lambda + \hat{V}_w) \hat{V}_s} \frac{z \Delta - z^2 - (\kappa^{-1} + 1) z}{2 \Delta} \\[4pt]
&V_w = \frac{1}{\lambda + \hat{V}_w} \frac{1 - \kappa^{-1} - z + \Delta}{2}
\end{aligned}
\right.
\end{equation}
where $z = \frac{\lambda + \hat{V}_w}{\hat{V}_s}$ and $\Delta = \sqrt{(1 + \kappa^{-1} + z)^2 - 4 \kappa^{-1}}$. Extremization with respect to $B$, $m_s$, $q_s$, $V_s$, $q_w$ and $V_w$ instead yields
\begin{equation} \label{eq: appendix - SE random features conjugates mu0 not zero}
\left\{
\begin{aligned}
&\hat{m}_s = \mu_1 \alpha \, \mathbb{E}_{\xi \sim \mathcal{N}(0,1)} \left[ \int_\mathbb{R} \dd y \, \mathcal{Z}_{\mathrm{out}^\star}(y, \sqrt{\eta} \xi, 1 - \eta) \, f_{\mathrm{out}^\star}(y, \sqrt{\eta} \xi, 1 - \eta) \, f_{\mathrm{out}}(y, B + \sqrt{q} \xi, V) \right] \\[6pt]
&\hat{q}_s = \mu_1^2 \alpha \, \mathbb{E}_{\xi \sim \mathcal{N}(0,1)} \left[ \int_\mathbb{R} \dd y \, \mathcal{Z}_{\mathrm{out}^\star}(y, \sqrt{\eta} \xi, 1 - \eta) \, f_{\mathrm{out}}(y, B + \sqrt{q} \xi, V)^2 \right] \\[6pt]
&\hat{V}_s = - \mu_1^2 \alpha \, \mathbb{E}_{\xi \sim \mathcal{N}(0,1)} \left[ \int_\mathbb{R} \dd y \, \mathcal{Z}_{\mathrm{out}^\star}(y, \sqrt{\eta} \xi, 1 - \eta) \, \partial_\omega f_{\mathrm{out}}(y, B + \sqrt{q} \xi, V) \right] \\[6pt]
&\hat{q}_w = \frac{1}{\kappa} \mu_\star^2 \alpha \, \mathbb{E}_{\xi \sim \mathcal{N}(0,1)} \left[ \int_\mathbb{R} \dd y \, \mathcal{Z}_{\mathrm{out}^\star}(y, \sqrt{\eta} \xi, 1 - \eta) \, f_{\mathrm{out}}(y, B + \sqrt{q} \xi, V)^2 \right] \\[6pt]
&\hat{V}_w = - \frac{1}{\kappa} \mu_\star^2 \alpha \, \mathbb{E}_{\xi \sim \mathcal{N}(0,1)} \left[ \int_\mathbb{R} \dd y \, \mathcal{Z}_{\mathrm{out}^\star}(y, \sqrt{\eta} \xi, 1 - \eta) \, \partial_\omega f_{\mathrm{out}}(y, B + \sqrt{q} \xi, V) \right] \\[6pt]
&0 = \mathbb{E}_{\xi \sim \mathcal{N}(0,1)} \left[ \int_\mathbb{R} \dd y \, \mathcal{Z}_{\mathrm{out}^\star}(y, \sqrt{\eta} \xi, 1 - \eta) \, f_{\mathrm{out}}(y, B + \sqrt{q} \xi, V) \right]
\end{aligned}
\right.
\end{equation}
where $\eta = \frac{m^2}{q}$ and the functions $f_\mathrm{out^\star}$ and $f_\mathrm{out}$ are defined as
\begin{align}
\label{eq: appendix - Bayes-optimal denoising function}
f_{\mathrm{out}^\star}(y, \omega, \tau) &= \partial_\omega \log \mathcal{Z}_{\mathrm{out}^\star}(y, \omega, \tau),
\\[6pt]
\label{eq: appendix - ERM denoising function}
f_\mathrm{out}(y, \omega, \tau) &= \frac{\mathcal{P}_{\tau \ell(y, \cdot)}(\omega) - \omega}{\tau}, 
\end{align}
where $\mathcal{P}_{\tau \ell(y, \cdot)}$ is the proximal operator
\begin{equation}
    \mathcal{P}_{\tau \ell(y, \cdot)}(\omega) = \operatorname*{arg \, min}_{z \in \mathbb{R}} \left[ \ell(y, z) + \frac{(z - \omega)^2}{2 \tau} \right].
\end{equation}

The state equations above hold for any $\mu_0 \neq 0$. When the state equations derived in \cite{gerace2020generalisation} for $\mu_0 = 0$ are specialized to Gaussian feature matrices, the resulting system differs only slightly from the one obtained above. In particular, the equations determining the overlap order parameters are formally identical to Eqs.~\eqref{eq: appendix - SE random features overlaps}, whereas the conjugate equations read
\begin{equation} \label{eq: appendix - SE random features conjugates}
\left\{
\begin{aligned}
&\hat{m}_s = \mu_1 \alpha \, \mathbb{E}_{\xi \sim \mathcal{N}(0,1)} \left[ \int_\mathbb{R} \dd y \, \mathcal{Z}_{\mathrm{out}^\star}(y, \sqrt{\eta} \xi, 1 - \eta) \, f_{\mathrm{out}^\star}(y, \sqrt{\eta} \xi, 1 - \eta) \, f_{\mathrm{out}}(y, \sqrt{q} \xi, V) \right] \\[6pt]
&\hat{q}_s = \mu_1^2 \alpha \, \mathbb{E}_{\xi \sim \mathcal{N}(0,1)} \left[ \int_\mathbb{R} \dd y \, \mathcal{Z}_{\mathrm{out}^\star}(y, \sqrt{\eta} \xi, 1 - \eta) \, f_{\mathrm{out}}(y, \sqrt{q} \xi, V)^2 \right] \\[6pt]
&\hat{V}_s = - \mu_1^2 \alpha \, \mathbb{E}_{\xi \sim \mathcal{N}(0,1)} \left[ \int_\mathbb{R} \dd y \, \mathcal{Z}_{\mathrm{out}^\star}(y, \sqrt{\eta} \xi, 1 - \eta) \, \partial_\omega f_{\mathrm{out}}(y, \sqrt{q} \xi, V) \right] \\[6pt]
&\hat{q}_w = \frac{1}{\kappa} \mu_\star^2 \alpha \, \mathbb{E}_{\xi \sim \mathcal{N}(0,1)} \left[ \int_\mathbb{R} \dd y \, \mathcal{Z}_{\mathrm{out}^\star}(y, \sqrt{\eta} \xi, 1 - \eta) \, f_{\mathrm{out}}(y, \sqrt{q} \xi, V)^2 \right] \\[6pt]
&\hat{V}_w = - \frac{1}{\kappa} \mu_\star^2 \alpha \, \mathbb{E}_{\xi \sim \mathcal{N}(0,1)} \left[ \int_\mathbb{R} \dd y \, \mathcal{Z}_{\mathrm{out}^\star}(y, \sqrt{\eta} \xi, 1 - \eta) \, \partial_\omega f_{\mathrm{out}}(y, \sqrt{q} \xi, V) \right]
\end{aligned}
\right.
\end{equation}

\subsection{The memorization error formula}
As explained in Section \ref{sec: appendix - Gibbs formulation of the minimization problem}, we can access the asymptotic memorization error of the empirical risk minimizer through $f_{\beta, h}$. In the following, the quantities $B = \mu_0 b$, $q = \mu_1^2 q_s + \mu_\star^2 q_w$ and $V = \mu_1^2 V_s + \mu_\star^2 V_w$ are understood to be evaluated at the extremum of the right-hand side of Eq.~\eqref{eq: appendix - f_beta_h mu0 not zero for Gaussian priors}. Accordingly, they depend implicitly on $h$ and $\beta$. In particular, for $h = 0$ and in the limit $\beta \to +\infty$, they coincide with the solutions of the state equations \eqref{eq: appendix - SE random features overlaps}--\eqref{eq: appendix - SE random features conjugates mu0 not zero}. For notational simplicity, we leave their dependence on $h$ and $\beta$ implicit.

By stationarity of the order parameters at the extremum, their implicit dependence on $h$ does not contribute when differentiating the extremal value of $f_{\beta,h}$. Therefore, 
\begin{equation}
    \frac{\partial f_{\beta, h}}{\partial h} \Bigg|_{h=0} = \frac{\alpha}{\kappa} \, \mathbb{E}_{\xi \sim \mathcal{N}(0,1)}\!\left[ \, \int_\mathbb{R} \dd y \, P_\mathrm{mem}(y) \frac{\int_\mathbb{R} \dd z \, e^{-\beta \Bigl[ \ell(y, z) + \frac{(z - B - \sqrt{q} \xi)^2}{2 \beta V} \Bigr]} \theta(- y z)}{\int_\mathbb{R} \dd z \, e^{-\beta \Bigl[ \ell(y, z) + \frac{(z - B - \sqrt{q} \xi)^2}{2 \beta V} \Bigr]} } \, \right].
\end{equation}
Applying the zero-temperature rescaling in Eq.~\eqref{eq: appendix - rescaling for zero temperature} and taking the limit $\beta \to +\infty$, we obtain
\begin{equation}
    \lim_{\beta \to +\infty} \frac{\partial f_{\beta, h}}{\partial h} \Bigg|_{h=0} = \frac{\alpha}{\kappa} \, \mathbb{E}_{\xi \sim \mathcal{N}(0,1)}\!\left[ \, \int_\mathbb{R} \dd y \, P_\mathrm{mem}(y) \, \theta(- y \mathcal{P}_{V \ell(y, \cdot)}(B + \sqrt{q} \xi)) \, \right].
\end{equation}
It then follows from Eq.~\eqref{eq: appendix - mem error from free energy density} that the asymptotic memorization error is
\begin{equation}
    \mathcal{E}_\mathrm{mem} = \mathbb{E}_{\xi \sim \mathcal{N}(0,1)}\!\biggl[ \, \int_\mathbb{R} \dd y \, P_\mathrm{mem}(y) \, \theta(- y \mathcal{P}_{V \ell(y, \cdot)}(B + \sqrt{q} \xi)) \, \biggr].
\end{equation}
Substituting Eq.~\eqref{eq: appendix - mem output channel} for $P_\mathrm{mem}(y)$ yields
\begin{equation} \label{eq: appendix - memorization error formula mu0 not zero}
    \mathcal{E}_\mathrm{mem} = \frac{1}{2} \, \mathbb{E}_{\xi \sim \mathcal{N}(0,1)}\!\left[ \theta\!\left(\mathcal{P}_{V \ell(-1,\cdot)}(B + \sqrt{q}\xi)\right) + \theta\!\left(-\mathcal{P}_{V \ell(1,\cdot)}(B + \sqrt{q}\xi)\right) \right].
\end{equation}

For $\mu_0 = 0$, an analogous calculation gives
\begin{equation} \label{eq: appendix - memorization error formula}
    \mathcal{E}_\mathrm{mem} = \frac{1}{2} \, \mathbb{E}_{\xi \sim \mathcal{N}(0,1)}\!\left[ \theta\!\left(\mathcal{P}_{V \ell(-1,\cdot)}(\sqrt{q}\xi)\right) + \theta\!\left(-\mathcal{P}_{V \ell(1,\cdot)}(\sqrt{q}\xi)\right) \right].
\end{equation}
Here, $q$ and $V$ are the solutions of the state equations \eqref{eq: appendix - SE random features overlaps} and \eqref{eq: appendix - SE random features conjugates}.

Equation~\eqref{eq: appendix - memorization error formula} is precisely Eq.~\eqref{eq:mmem-error} presented in the main text. In the next section, we show that, for the RAF output channel, Eq.~\eqref{eq: appendix - memorization error formula mu0 not zero} formally reduces to the same expression.

\subsection{Sign-symmetric output channels and losses}
Consider a generic output channel $P_\mathrm{out^\star}(y \mid \nu)$, so that the labels are generated according to
\begin{equation}
    y^\mu \sim P_\mathrm{out^\star}(\cdot \mid \nu_\mu), \qquad \nu_\mu = \frac{\boldsymbol{w}_\star \cdot \boldsymbol{x}^\mu}{\sqrt{d}}.
\end{equation}
Assume that the output channel is invariant under the simultaneous sign reversal of the label and the teacher preactivation, namely
\begin{equation} \label{eq:generic-channel-sign-symmetry}
  P_\mathrm{out^\star}(-y \mid -\nu) = P_\mathrm{out^\star}(y \mid \nu).
\end{equation}
This symmetry is inherited by its Gaussian convolution:
\begin{equation} \label{eq:generic-BO-denoising-function-symmetry}
  \cZ_{\rm out^\star}(-y, -\omega, \tau) = \cZ_{\rm out^\star}(y, \omega, \tau).
\end{equation}

Suppose, in addition, that the loss satisfies the analogous symmetry
\begin{equation} \label{eq:sign-symmetric-loss}
  \ell(-y,-z)=\ell(y,z).
\end{equation}
The corresponding proximal operator then obeys
\begin{equation} \label{eq:generic-proximal-sign-symmetry}
  \mathcal P_{\tau\ell(-y,\cdot)}(-\omega) = - \mathcal P_{\tau\ell(y,\cdot)}(\omega).
\end{equation}
which implies
\begin{equation} \label{eq:generic-response-sign-symmetry}
  f_\mathrm{out}(-y,-\omega,\tau) = - f_\mathrm{out}(y,\omega,\tau).
\end{equation}

Under these assumptions, we can identify a solution for the order parameter $B$ directly from its saddle-point equation in ~\eqref{eq: appendix - SE random features conjugates mu0 not zero}. Setting $B=0$, denote the expectation appearing in this equation by
\begin{equation}
    \mathcal{I}_{B = 0} = \mathbb{E}_{\xi \sim \mathcal{N}(0,1)} \left[ \int_\mathbb{R} \dd y \, \mathcal{Z}_{\mathrm{out}^\star}(y, \sqrt{\eta} \xi, 1 - \eta) \, f_{\mathrm{out}}(y, \sqrt{q} \xi, V) \right]
\end{equation}
Performing the simultaneous change of variables
$(\xi,y) \mapsto (-\xi,-y)$ leaves the standard Gaussian measure of $\xi$ and the measure $\dd y$ invariant. Using Eqs.~\eqref{eq:generic-BO-denoising-function-symmetry} and \eqref{eq:generic-response-sign-symmetry}, we therefore obtain
\begin{equation}
\begin{split}
    \mathcal{I}_{B = 0} &= \mathbb{E}_{\xi \sim \mathcal{N}(0,1)} \left[ \int_\mathbb{R} \dd y \, \mathcal{Z}_{\mathrm{out}^\star}(-y, - \sqrt{\eta} \xi, 1 - \eta) \, f_{\mathrm{out}}(-y, - \sqrt{q} \xi, V) \right]
    \\[4pt]
    &= - \mathbb{E}_{\xi \sim \mathcal{N}(0,1)} \left[ \int_\mathbb{R} \dd y \, \mathcal{Z}_{\mathrm{out}^\star}(y, \sqrt{\eta} \xi, 1 - \eta) \, f_{\mathrm{out}}(y, \sqrt{q} \xi, V) \right]
    \\[4pt]
    &= - \mathcal{I}_{B = 0}
\end{split}
\end{equation}
It follows that $\mathcal{I}_{B=0} = 0$, and hence $B=0$ satisfies the corresponding saddle-point equation.

At the solution $B=0$, the state equations \eqref{eq: appendix - SE random features overlaps}--\eqref{eq: appendix - SE random features conjugates mu0 not zero} for $\mu_0 \neq 0$ formally reduce to those for $\mu_0 = 0$, given by Eqs.~\eqref{eq: appendix - SE random features overlaps} and \eqref{eq: appendix - SE random features conjugates}. Likewise, the memorization error in Eq.~\eqref{eq: appendix - memorization error formula mu0 not zero} reduces to Eq.~\eqref{eq: appendix - memorization error formula}. Moreover, the asymptotic generalization error of the empirical risk minimizer can be evaluated using the formula already derived in \cite{gerace2020generalisation}, which we report here for completeness:
\begin{equation} \label{eq: appendix - generalization error formula}
    \mathcal{E}_\mathrm{gen} = \frac{1}{\pi} \arccos\!\left(\frac{m}{\sqrt{q}}\right),
\end{equation}
where $m = \mu_1 m_s$ and $q = \mu_1^2 q_s + \mu_\star^2 q_w$, with $m_s$, $q_s$, and $q_w$ determined by the common state equations \eqref{eq: appendix - SE random features overlaps} and \eqref{eq: appendix - SE random features conjugates}.

The RAF output channel in Eq.~\eqref{eq: appendix - RAF output channel} satisfies the sign-reversal symmetry \eqref{eq:generic-channel-sign-symmetry}, while both the square and hinge losses considered in this work satisfy Eq.~\eqref{eq:sign-symmetric-loss}. Accordingly, throughout the remainder of the appendices, we implicitly work with the solution $B=0$ and use the common form of the state equations and memorization error derived above, without explicitly distinguishing between the cases $\mu_0 = 0$ and $\mu_0 \neq 0$.

%% file: Appendices/Appendix_Kernel_review.tex
\section{Random features regression in the highly-overparametrized regime: the kernel limit}
\label{app: eqs_kernel_limit}
In the limit $\kappa \to \infty$ with $\alpha$ and $\lambda$ held finite, the fixed point of Eqs.~\eqref{eq: appendix - SE random features overlaps} and \eqref{eq: appendix - SE random features conjugates} yields the overlap order parameters that characterize the asymptotic performance of dot-product kernels. In this appendix, we first show how to take the limit and then justify why the resulting equations describe regression with dot-product kernels. We also highlight notable symmetry properties of their solutions. We emphasize that this limit was not carried out in \cite{gerace2020generalisation}.

\subsection{State equations for kernel regression}
Assume that, as $\kappa \to \infty$ with $\alpha$ and $\lambda$ finite, the conjugate variables scale as
\begin{equation} \label{eq: appendix - scalings for zero aspect ratio}
    \hat{m}_s = \Theta(1),
    \qquad \hat{q}_s = \Theta(1),
    \qquad \hat{V}_s = \Theta(1),
    \qquad \hat{q}_w = o(1),
    \qquad \hat{V}_w = o(1).
\end{equation}
Under this ansatz, we expand the right-hand sides of Eqs.~\eqref{eq: appendix - SE random features overlaps}. A key subtlety is to avoid ill-defined ratios in the $\kappa \to \infty$ limit. To this end, write $z = \frac{\lambda}{\hat{V}_s} + \frac{1}{\kappa} \frac{\mu_\star^2}{\mu_1^2}$ and expand $\Delta$ as
\begin{equation}
    \Delta \underset{\kappa \to \infty}{\simeq} 1 + \frac{\lambda}{\hat{V}_s} + \frac{1}{\kappa} \frac{\left( 1 + \frac{\lambda}{\hat{V}_s} \right) \left( 1 + \frac{\mu_\star^2}{\mu_1^2} \right) - 2}{1 + \frac{\lambda}{\hat{V}_s}}.
\end{equation}
Substituting this expansion and simplifying the resulting expressions, the equations for the overlaps reduce to the remarkably simple form
\begin{equation} \label{eq: appendix - zero aspect ratio equations for the random features overlaps}
    m_s = \frac{\hat{m}_s}{\lambda + \hat{V}_s}, \qquad
    q_s = \frac{\hat{m}_s^2 + \hat{q}_s}{(\lambda + \hat{V}_s)^2},
    \qquad
    V_s = \frac{1}{\lambda + \hat{V}_s},
    \qquad
    q_w = 0,
    \qquad
    V_w = \frac{1}{\lambda}.
\end{equation}
It remains to verify that the conjugate equations \eqref{eq: appendix - SE random features conjugates} are consistent with the scaling assumption \eqref{eq: appendix - scalings for zero aspect ratio}. Provided that the integrals appearing on their right-hand sides are sufficiently regular functions of $(\eta, q, V)$, this consistency condition is satisfied.

As a consequence, the variables $q_w, V_w, \hat{q}_w$ and $\hat{V}_w$ decouple, and the remaining equations reduce to
\begin{equation} \label{eq: appendix - SE kernel from direct limit gamma to zero}
\left\{
\begin{aligned}
& m_s = \frac{\hat{m}_s}{\lambda + \hat{V}_s} \\[4pt]
& q_s = \frac{\hat{m}_s^2 + \hat{q}_s}{(\lambda + \hat{V}_s)^2} \\[4pt]
& V_s = \frac{1}{\lambda + \hat{V}_s} \\[6pt]
& \hat{m}_s = \mu_1 \alpha \, \mathbb{E}_{\xi \sim \mathcal{N}(0,1)} \left[ \int_\mathbb{R} \dd y \, \mathcal{Z}_\mathrm{out^\star}(y, \sqrt{\eta} \xi, 1 - \eta) \, f_\mathrm{out^\star}(y, \sqrt{\eta} \xi, 1 - \eta) \, f_\mathrm{out}(y, \sqrt{q} \xi, V) \right] \\[6pt]
& \hat{q}_s = \mu_1^2 \alpha \, \mathbb{E}_{\xi \sim \mathcal{N}(0,1)} \left[ \int_\mathbb{R} \dd y \, \mathcal{Z}_\mathrm{out^\star}(y, \sqrt{\eta} \xi, 1 - \eta) \, f_\mathrm{out}(y, \sqrt{q} \xi, V)^2 \right] \\[6pt]
& \hat{V}_s = - \mu_1^2 \alpha \, \mathbb{E}_{\xi \sim \mathcal{N}(0,1)} \left[ \int_\mathbb{R} \dd y \, \mathcal{Z}_\mathrm{out^\star}(y, \sqrt{\eta} \xi, 1 - \eta) \, \partial_\omega f_\mathrm{out}(y, \sqrt{q} \xi, V) \right]
\end{aligned}
\right.
\end{equation}
Here $m = \mu_1 m_s$, $q = \mu_1^2 q_s$ and $V = \mu_1^2 V_s + \frac{\mu_\star^2}{\lambda}$.

In this regime, it is more convenient to express the system of state equations directly in terms of the rescaled order parameters $m, q, V$ rather than the bare ones $m_s, q_s, V_s$. Rewriting the first three equations of \eqref{eq: appendix - SE kernel from direct limit gamma to zero} accordingly yields the equivalent system
\begin{equation}
\left\{
\begin{aligned}
& m = \mu_1 \frac{\hat{m}_s}{\lambda + \hat{V}} \\[4pt]
& q = \mu_1^2 \frac{\hat{m}_s^2 + \hat{q}_s}{(\lambda + \hat{V}_s)^2} \\[4pt]
& V = \mu_1^2 \frac{1}{\lambda + \hat{V}_s} + \frac{\mu_\star^2}{\lambda} \\[6pt]
& \hat{m}_s = \mu_1 \alpha \, \mathbb{E}_{\xi \sim \mathcal{N}(0,1)} \left[ \int_\mathbb{R} \dd y \, \mathcal{Z}_\mathrm{out^\star}(y, \sqrt{\eta} \xi, 1 - \eta) \, f_\mathrm{out^\star}(y, \sqrt{\eta} \xi, 1 - \eta) \, f_\mathrm{out}(y, \sqrt{q} \xi, V) \right] \\[6pt]
& \hat{q}_s = \mu_1^2 \alpha \, \mathbb{E}_{\xi \sim \mathcal{N}(0,1)} \left[ \int_\mathbb{R} \dd y \, \mathcal{Z}_\mathrm{out^\star}(y, \sqrt{\eta} \xi, 1 - \eta) \, f_\mathrm{out}(y, \sqrt{q} \xi, V)^2 \right] \\[6pt]
& \hat{V}_s = - \mu_1^2 \alpha \, \mathbb{E}_{\xi \sim \mathcal{N}(0,1)} \left[ \int_\mathbb{R} \dd y \, \mathcal{Z}_\mathrm{out^\star}(y, \sqrt{\eta} \xi, 1 - \eta) \, \partial_\omega f_\mathrm{out}(y, \sqrt{q} \xi, V) \right]
\end{aligned}
\right.
\end{equation}
At this stage, the subscript $s$ becomes redundant and can be dropped for notational simplicity. This leads to the final form of the random features state equations in the highly-overparametrized limit:
\begin{equation} \label{eq: appendix - SE kernel}
\left\{
\begin{aligned}
& m = \mu_1 \frac{\hat{m}}{\lambda + \hat{V}} \\[4pt]
& q = \mu_1^2 \frac{\hat{m}^2 + \hat{q}}{(\lambda + \hat{V})^2} \\[4pt]
& V = \mu_1^2 \frac{1}{\lambda + \hat{V}} + \frac{\mu_\star^2}{\lambda} \\[6pt]
& \hat{m} = \mu_1 \alpha \, \mathbb{E}_{\xi \sim \mathcal{N}(0,1)} \left[ \int_\mathbb{R} \dd y \, \mathcal{Z}_\mathrm{out^\star}(y, \sqrt{\eta} \xi, 1 - \eta) \, f_\mathrm{out^\star}(y, \sqrt{\eta} \xi, 1 - \eta) \, f_\mathrm{out}(y, \sqrt{q} \xi, V) \right] \\[6pt]
& \hat{q} = \mu_1^2 \alpha \, \mathbb{E}_{\xi \sim \mathcal{N}(0,1)} \left[ \int_\mathbb{R} \dd y \, \mathcal{Z}_\mathrm{out^\star}(y, \sqrt{\eta} \xi, 1 - \eta) \, f_\mathrm{out}(y, \sqrt{q} \xi, V)^2 \right] \\[6pt]
& \hat{V} = - \mu_1^2 \alpha \, \mathbb{E}_{\xi \sim \mathcal{N}(0,1)} \left[ \int_\mathbb{R} \dd y \, \mathcal{Z}_\mathrm{out^\star}(y, \sqrt{\eta} \xi, 1 - \eta) \, \partial_\omega f_\mathrm{out}(y, \sqrt{q} \xi, V) \right]
\end{aligned}
\right.
\end{equation}

\subsection{The \texorpdfstring{$\kappa \to \infty$}{kappa -> infinity} limit and the link between \texorpdfstring{$\mu_1, \mu_\star$}{(mu1, mustar)} and dot-product kernels}
\label{app:kernel_limit}
Let us now explain why Eqs.~\eqref{eq: appendix - SE kernel} describe dot-product kernel regression in the high-dimensional limit. For a comprehensive review on kernel methods, see \cite{Hofmann2008kernel}.

We start by the random-features predictor
\begin{equation}
f_{\bw}(\bx)=\frac{1}{\sqrt p}\sum_{a=1}^p w_a\,\sigma\!\left(\frac{\boldsymbol{f}_a \cdot \bx}{\sqrt d}\right),
\end{equation}
where $\boldsymbol{f}_a\in\mathbb{R}^d$ denotes the $a$-th column of $F$. For fixed training dataset $\mathcal{D} = \{(\bx^\mu,y^\mu)\}_{\mu=1}^n$ and fixed dimension $d$, this is a linear model in the $p$ random features
\begin{equation}
\phi_a(\bx)=\sigma\!\left(\frac{\boldsymbol{f}_a \cdot \bx}{\sqrt d}\right).
\end{equation}
Its empirical-risk minimizer hence depends on the data only through the Gram matrix of the random features, namely
\begin{equation}
\label{eq: appendix - random features Gram matrix}
\frac{1}{p}\sum_{a=1}^p \phi_a(\bx^\mu)\phi_a(\bx^\nu)
=
\frac{1}{p}\sum_{a=1}^p
\sigma\!\left(\frac{\boldsymbol{f}_a \cdot \bx^\mu}{\sqrt d}\right)
\sigma\!\left(\frac{\boldsymbol{f}_a \cdot \bx^\nu}{\sqrt d}\right).
\end{equation}
Since the columns of $F$ are i.i.d., the law of large numbers implies that, for fixed $n$ and $d$,
\begin{equation} \label{eq: appendix - random fetures to kernel at fixed n and d}
\frac{1}{p}\sum_{a=1}^p \phi_a(\bx^\mu)\phi_a(\bx^\nu)
\underset{p\to\infty}{\longrightarrow}
K(\bx^\mu,\bx^\nu)
=
\mathbb{E}_{\boldsymbol{f}}
\!\left[
\sigma\!\left(\frac{\boldsymbol{f} \cdot \bx^\mu}{\sqrt d}\right)
\sigma\!\left(\frac{\boldsymbol{f} \cdot \bx^\nu}{\sqrt d}\right)
\right]
\end{equation}
where $\boldsymbol{f}$ is drawn from the common distribution of the columns of $F$. Therefore, for any finite $n$ and $d$, the random-features model in the infinite-width limit is exactly equivalent to kernel regression with the kernel defined above.

Our analysis is conducted in the high-dimensional regime
\begin{equation}
    n,d,p\to\infty, \qquad \alpha=\frac{n}{d}=\Theta(1), \qquad \kappa=\frac{p}{d}=\Theta(1).
\end{equation}
In this regime, the ordinary infinite-width condition $p \to \infty$ is already satisfied even at finite $\kappa$. The parameter that distinguishes a finite-width random-features model from its infinite-width counterpart is therefore the relative width $\kappa = p / d$. Finite $\kappa$ corresponds to a number of random features proportional to the input dimension, whereas $\kappa \to \infty$ makes the number of features arbitrarily large relative to the high-dimensional scale. Moreover, since $\alpha = n / d$ is fixed, $p / n = \kappa / \alpha \to \infty$. In this precise scaling sense, the limit $\kappa \to \infty$ is the high-dimensional counterpart of the usual infinite-width limit $p \to \infty$ at fixed $n$ and $d$. At the level of the asymptotic theory, taking $\kappa \to \infty$ in the random-features state equations \eqref{eq: appendix - SE random features overlaps}--\eqref{eq: appendix - SE random features conjugates} yields Eqs.~\eqref{eq: appendix - SE kernel}. These equations therefore describe the high-dimensional asymptotic performance of regression with the kernel induced by the corresponding infinite-width random-features construction.

It remains to show that, in our setting, the kernel defined above is a dot-product kernel, and to identify its connection with the parameters $\mu_1$ and $\mu_\star$ appearing in Eqs.~\eqref{eq: appendix - SE kernel}.

In the random-features construction with activation function $\sigma$ and Gaussian first-layer weights, the infinite-width limit induces the kernel
\begin{equation}
\label{eq: appendix - kernel random-features definition}
K(\bx^\mu,\bx^\nu)
=
\mathbb{E}_{(g,g')\sim\mathcal{N}(\boldsymbol{0},\Sigma(\bx^\mu, \bx^\nu))}
\bigl[\sigma(g)\sigma(g')\bigr],
\end{equation}
where
\begin{equation}
\Sigma(\bx^\mu, \bx^\nu)=
\begin{pmatrix}
\frac{\| \bx^\mu \|^2}{d} & \rho\\
\rho & \frac{\| \bx^\nu\|^2}{d}
\end{pmatrix},
\qquad
\rho=\frac{\bx^{\mu} \cdot \bx^\nu}{d}.
\end{equation}
For inputs drawn uniformly on the sphere of radius $\sqrt{d}$, the kernel in Eq.~\eqref{eq: appendix - kernel random-features definition} becomes a dot-product kernel:
\begin{equation}
\label{eq: appendix - kernel random-features dot-product}
K(\rho)
=
\mathbb{E}_{(g,g')\sim\mathcal{N}(\boldsymbol{0},\Sigma(\rho))}
\bigl[\sigma(g)\sigma(g')\bigr],
\end{equation}
where
\begin{equation} \label{eq: appendix - covariance matrix dot-product}
\Sigma(\rho)=
\begin{pmatrix}
1 & \rho\\
\rho & 1
\end{pmatrix},
\qquad
\rho=\frac{\bx^{\mu} \cdot \bx^\nu}{d}.
\end{equation}
The same reduction holds asymptotically for Gaussian inputs $\bx^\mu \sim \mathcal{N}(\boldsymbol{0}, \mathbb{I}_d)$, which is the setting considered here. Indeed, as $d \to \infty$, the norm of a sample approaches $\sqrt{d}$, while for distinct samples the normalized inner product $\rho$ approaches zero. Notice that this conclusion is unchanged in the proportional limit $n, d \to \infty$ with $\alpha = n / d$ finite.

Let us now establish the connection between the dot-product kernel in Eq.~\eqref{eq: appendix - kernel random-features dot-product} and the parameters $\mu_1, \mu_\star$. Assuming that $\sigma \in L^2(\mathcal{N}(0,1))$, we can expand the activation function in the basis of the Hermite polynomials:
\begin{equation}
\label{eq: appendix - expansion sigma hermite}
\sigma(g)=\sum_{\ell\geq 0} c_\ell H_\ell(g),
\qquad
c_\ell=\frac{1}{\ell!}\,\mathbb{E}_{g \sim \mathcal{N}(0,1)}\!\left[\sigma(g)H_\ell(g)\right].
\end{equation}
Here $\{ H_\ell \}_{\ell \geq 0}$ denote the probabilists' Hermite polynomials, normalized such that
\begin{equation}
\mathbb{E}_{g \sim \mathcal{N}(0,1)}\!\left[H_\ell(g)H_m(g)\right]=\ell!\,\delta_{\ell m}.
\end{equation}
If $(g,g')$ is a centered jointly Gaussian pair with correlation $\rho$, then the following identity holds:
\begin{equation}
\label{eq: appendix - cross hermite}
\mathbb{E}_{(g,g')\sim\mathcal{N}(\boldsymbol{0},\Sigma(\rho))}\!\left[H_\ell(g)H_m(g')\right] = \ell!\,\rho^\ell\,\delta_{\ell m},
\end{equation}
where $\Sigma(\rho)$ is as in Eq.~\eqref{eq: appendix - covariance matrix dot-product}. Using \eqref{eq: appendix - expansion sigma hermite} together with \eqref{eq: appendix - cross hermite}, the dot-product kernel in Eq.~\eqref{eq: appendix - kernel random-features dot-product} can be expanded as
\begin{equation}
\label{eq: appendix - kernel expansion}
K(\rho)
=
\sum_{\ell\geq 0} a_\ell \rho^\ell,
\qquad
a_\ell=\ell!\,c_\ell^2
=
\frac{1}{\ell!}
\left(\mathbb{E}_{g \sim \mathcal{N}(0,1)}\!\left[\sigma(g)H_\ell(g)\right]\right)^2
\geq 0.
\end{equation}
In particular, we have
\begin{equation}
\label{eq: appendix - kernel coefficients}
a_0=\left(\mathbb{E}_{g \sim \mathcal{N}(0,1)}[\sigma(g)]\right)^2,
\qquad
a_1=\left(\mathbb{E}_{g \sim \mathcal{N}(0,1)}[g\,\sigma(g)]\right)^2,
\qquad
\sum_{\ell\geq 0} a_\ell
=
\mathbb{E}_{g \sim \mathcal{N}(0,1)}[\sigma(g)^2].
\end{equation}
Recall now the definitions of $\mu_0, \mu_1$ and $\mu_\star$ introduced in the random-features analysis:
\begin{equation}
\label{eq: appendix - definition mus random features}
\mu_0=\mathbb{E}_{g \sim \mathcal{N}(0,1)}[\sigma(g)],
\qquad
\mu_1=\mathbb{E}_{g \sim \mathcal{N}(0,1)}[g\,\sigma(g)],
\qquad
\mu_\star^2=\mathbb{E}_{g \sim \mathcal{N}(0,1)}[\sigma(g)^2]-\mu_0^2-\mu_1^2.
\end{equation}
Comparing \eqref{eq: appendix - kernel coefficients} and \eqref{eq: appendix - definition mus random features}, we obtain
\begin{equation}
\mu_0^2=a_0=K(0),
\qquad
\mu_1^2=a_1=K'(0),
\qquad
\mu_\star^2=\sum_{\ell\geq 2} a_\ell
=K(1)-K(0)-K'(0).
\label{eq: appendix - mus from kernel}
\end{equation}
This proves the identities~\eqref{eq:activations_kernel} reported in the main text. For more details on the Hermite polynomials, we defer to~\cite{szeg1939orthogonal}.

Thus, in the highly-overparameterized regime, the state equations for Gaussian random-features regression with activation $\sigma \in L^2(\mathcal{N}(0,1))$ reduce to those characterizing the asymptotic performance of kernel regression with a dot-product kernel admitting an expansion of the form \eqref{eq: appendix - kernel expansion}. In particular, the asymptotic performance depends on the kernel only through the two parameters $\mu_1$ and $\mu_\star$, which are obtained directly from its expansion coefficients via the identities \eqref{eq: appendix - mus from kernel}.

\subsection{The linear model as a special kernel}
If we set $\mu_1 = 1$ and $\mu_\star = 0$, Eqs.~\eqref{eq: appendix - SE kernel} become exactly the state equations describing the asymptotic performance of a linear model trained by minimizing an empirical risk with convex loss and ridge regularization, which were derived in \cite{aubin2020generalization}. Indeed, the combination $\mu_1 = 1$ and $\mu_\star = 0$ corresponds to the linear kernel $K(\rho) = a_0 + \rho$, thus to standard linear regression.

\subsection{Symmetry properties of kernel regression solutions}
\label{app: kernel symmetries}
The fixed point of Eqs.~\eqref{eq: appendix - SE kernel} exhibits notable symmetries when the kernel parameters $\mu_1$ and $\mu_\star$ are appropriately rescaled. In this section, we first derive a reduced system of state equations from Eqs.~\eqref{eq: appendix - SE kernel}, and then emphasize the key symmetries satisfied by their solutions.

\subsubsection{Reduced kernel regression state equations}
\label{app:kernel-reduction}
Introducing
\begin{equation}
\ell = \frac{\lambda}{\mu_1^2},
\qquad
t = \frac{\mu_1}{\mu_\star}.
\label{eq:app_defs_l_t}
\end{equation}
and defining the rescaled variables
\begin{equation}
\tilde{m} = \frac{\hat m}{\mu_1},
\qquad
\tilde{q} = \frac{\hat q}{\mu_1^2},
\qquad
\tilde{V} = \frac{\hat V}{\mu_1^2}.
\label{eq:app_hinge_defs_u_s_t}
\end{equation}
the system of Eqs.~\eqref{eq: appendix - SE kernel} can be recast as
\begin{equation} \label{eq: appendix - SE kernel reduced}
\left\{
\begin{aligned}
& m = \frac{\tilde{m}}{\ell+\tilde{V}} \\[4pt]
& q = \frac{\tilde{m}^2+\tilde{q}}{(\ell+\tilde{V})^2} \\[4pt]
& V = \frac{1}{\ell+\tilde{V}} + \frac{1}{t^2 \ell} \\[6pt]
& \tilde{m} = \alpha \, \mathbb{E}_{\xi \sim \mathcal{N}(0,1)} \left[ \int_\mathbb{R} \dd y \, \mathcal{Z}_\mathrm{out^\star}(y, \sqrt{\eta} \xi, 1 - \eta) \, f_\mathrm{out^\star}(y, \sqrt{\eta} \xi, 1 - \eta) \, f_\mathrm{out}(y, \sqrt{q} \xi, V) \right] \\[6pt]
& \tilde{q} = \alpha \, \mathbb{E}_{\xi \sim \mathcal{N}(0,1)} \left[ \int_\mathbb{R} \dd y \, \mathcal{Z}_\mathrm{out^\star}(y, \sqrt{\eta} \xi, 1 - \eta) \, f_\mathrm{out}(y, \sqrt{q} \xi, V)^2 \right] \\[6pt]
& \tilde{V} = - \alpha \, \mathbb{E}_{\xi \sim \mathcal{N}(0,1)} \left[ \int_\mathbb{R} \dd y \, \mathcal{Z}_\mathrm{out^\star}(y, \sqrt{\eta} \xi, 1 - \eta) \, \partial_\omega f_\mathrm{out}(y, \sqrt{q} \xi, V) \right]
\end{aligned}
\right.
\end{equation}
Eqs.~\eqref{eq: appendix - SE kernel reduced} form a closed reduced system for $(m, q, V, \tilde{m}, \tilde{q}, \tilde{V})$ that depends on the kernel only through the two scalars $(\ell, t)$ and on the data only through $(\alpha, \varepsilon)$. In particular, there exists a reduced solution map, namely
\begin{equation} \label{eq: appendix - kernel reduced solution map}
m = m(\alpha, \varepsilon; \ell, t),
\qquad
q = q(\alpha, \varepsilon; \ell, t),
\qquad
V = V(\alpha, \varepsilon; \ell, t).
\end{equation}

\subsubsection{Scaling invariance and angular symmetry in the plane \texorpdfstring{$(\mu_\star,\mu_1)$}{(mu1, mustar)}}

The reduced equations \eqref{eq: appendix - SE kernel reduced} are exactly invariant under the rescaling
\begin{equation}
(\mu_\star, \mu_1, \lambda)
\;\mapsto\;
(r\,\mu_\star,\, r\,\mu_1,\, r^2\,\lambda),
\qquad r > 0.
\label{eq: appendix - kernel invariant scaling}
\end{equation}
Indeed, since $\ell$ and $t$ remain unchanged under this transformation, it follows that the order parameters \eqref{eq: appendix - kernel reduced solution map} are also unchanged. 

This invariance has a simple geometric interpretation. Provided that the regularization strength is scaled as $\lambda \propto \mu_1^2$, the solutions of the reduced state equations organize along radial directions in the $(\mu_\star, \mu_1)$ plane. In other words, fixing the angle
\begin{equation}
\gamma = \arctan(t) = \arctan\!\left(\frac{\mu_1}{\mu_\star}\right)
\end{equation}
and moving along the corresponding straight line through the origin leaves the solution to Eqs.~\eqref{eq: appendix - SE kernel reduced} invariant. 

If instead $\lambda$ is held fixed, the invariance under radial rescaling $(\mu_\star,\mu_1) \mapsto (r\,\mu_\star, r\,\mu_1)$ is broken in general, as $\ell$ no longer remains unchanged. There are, however, two relevant exceptions.
\begin{itemize}
\item[(i)] \textbf{Extreme-regularization limits.} Radial invariance is recovered in both strict limits $\lambda \to 0^+$ and $\lambda \to +\infty$. Whenever the corresponding limits of the order parameters \eqref{eq: appendix - kernel reduced solution map} exist, possibly in the extended sense, the strict limiting solution of the reduced state equations is independent of the radial scale. This statement does not imply that the leading corrections at small but finite $\ell$, or at large but finite $\ell$, are themselves radially invariant.
\item[(ii)] \textbf{Optimal regularization.} Radial invariance is also recovered when $\lambda$ is chosen to optimize any function of the order parameters \eqref{eq: appendix - kernel reduced solution map}. Indeed, optimizing with respect to $\lambda$ at fixed $\mu_1, \mu_\star$ is then equivalent to optimizing with respect to $\ell$. The optimal reduced regularization strength therefore takes the form $\ell_{\mathrm{opt}} = \ell_{\mathrm{opt}}(\alpha,\varepsilon,t)$, and depends on $\mu_1, \mu_\star$ only through their ratio $t$. The corresponding regularization strength is $\lambda_\mathrm{opt} = \mu_1^2 \ell_\mathrm{opt}(\alpha, \varepsilon, t)$, which under radial rescaling transforms as $\lambda_\mathrm{opt} \mapsto r^2 \lambda_\mathrm{opt}$. As a consequence, the order parameters evaluated at $\ell_\mathrm{opt}$ depend on $\mu_1, \mu_\star$ only via $t$ and are thus preserved along radial directions.
\end{itemize}

Finally, in the two extreme-regularization limits, the same radial invariance holds for observables determined solely by the solution of the reduced state equations. Their limiting values therefore depend on $\mu_1, \mu_\star$ only through their ratio $t$.

%% file: Appendices/Appendix_KRR-SVM-RAF_review.tex
\section{Performance analysis of kernel ridge regression and support vector machines in the RAF task model}
\label{app:main_novel_analysis}
In this appendix, we specialize the general state equations derived for dot-product kernel regression in Appendix \ref{app: eqs_kernel_limit} to the output distribution of the RAF data model. We then provide a detailed analysis of the performance of dot-product kernel regression on the RAF tasks under two loss functions: the square loss, corresponding to kernel ridge regression (KRR), and the hinge loss, corresponding to support vector machines (SVMs).

The analysis in this appendix covers all the theoretical results and phenomenology for non-linear kernels presented in Sections \ref{sec: main equations} and \ref{sec:results} of the main text. At finite regularization, the linear model is recovered by setting $\mu_1 = 1$ and $\mu_\star = 0$. The regime $\lambda \to 0^+$ is instead treated separately, because taking the linear-model limit $\mu_\star \to 0$ need not commute with the ridgeless limit.

\subsection{Loss-dependent kernel state equations for the RAF data model} \label{app:loss-dep kernel SE RAF}
The output distribution $P_\mathrm{out^\star}$ and the loss $\ell$ enter the kernel state equations \eqref{eq: appendix - SE kernel} only through the right-hand sides of the equations for $\hat{m}, \hat{q}$ and $\hat{V}$: $P_\mathrm{out^\star}$ enters through $\mathcal{Z}_\mathrm{out^\star}$ and $f_\mathrm{out^\star}$, while $\ell$ enters through $f_\mathrm{out}$. For convenience, we recall these equations:
\begin{align}
    \hat{m} &= \mu_1 \alpha \, \mathbb{E}_{\xi \sim \mathcal{N}(0,1)} \left[ \int_\mathbb{R} \dd y \, \mathcal{Z}_\mathrm{out^\star}(y, \sqrt{\eta} \xi, 1 - \eta) \, f_\mathrm{out^\star}(y, \sqrt{\eta} \xi, 1 - \eta) \, f_\mathrm{out}(y, \sqrt{q} \xi, V) \right],
    \\
    \hat{q} &= \mu_1^2 \alpha \, \mathbb{E}_{\xi \sim \mathcal{N}(0,1)} \left[ \int_\mathbb{R} \dd y \, \mathcal{Z}_\mathrm{out^\star}(y, \sqrt{\eta} \xi, 1 - \eta) \, f_\mathrm{out}(y, \sqrt{q} \xi, V)^2 \right],
    \\
    \hat{V} &= - \mu_1^2 \alpha \, \mathbb{E}_{\xi \sim \mathcal{N}(0,1)} \left[ \int_\mathbb{R} \dd y \, \mathcal{Z}_\mathrm{out^\star}(y, \sqrt{\eta} \xi, 1 - \eta) \, \partial_\omega f_\mathrm{out}(y, \sqrt{q} \xi, V) \right].
\end{align}

Let us specialize them for the RAF data model. Plugging the RAF output distribution \eqref{eq: appendix - RAF output channel} into Eq.~\eqref{eq: appendix - output channel partition function} for $\mathcal{Z}_\mathrm{out^\star}$, we find
\begin{equation} \label{app:z_out_denoising RAF}
    \mathcal{Z}_\mathrm{out^\star}(y, \omega, \tau) = \frac{1}{2} \delta(y + 1) + \frac{1}{2} \delta(y - 1) + \frac{1 - \varepsilon}{2} \left[ \delta(y - 1) - \delta(y + 1) \right] \operatorname{erf}\!\left( \frac{\omega}{\sqrt{2 \tau}} \right).
\end{equation}
Then, $f_{\mathrm{out}^{\star}}$ follows from Eq.~\eqref{eq: appendix - Bayes-optimal denoising function}:
\begin{equation} \label{app:f_out_denoising RAF}
    f_\mathrm{out^\star}(y, \omega, \tau) = \frac{1}{\mathcal{Z}_{\mathrm{out}^\star}(y, \omega, \tau)} (1 - \varepsilon) \left[ \delta(y - 1) - \delta(y + 1) \right] \frac{e^{- \frac{\omega^2}{2 \tau}}}{\sqrt{2 \pi \tau}}.
\end{equation}
Finally, substituting the expressions of $\mathcal{Z}_\mathrm{out^\star}$ and $f_\mathrm{out^\star}$ from Eqs.~\eqref{app:z_out_denoising RAF} and \eqref{app:f_out_denoising RAF}, respectively, we obtain the loss-dependent kernel state equations for the RAF data model:
\begin{align}
    \hat{m} &= \mu_1 \alpha (1 - \varepsilon) \int_\mathbb{R} \frac{\dd \xi}{\sqrt{2\pi}} \frac{e^{- \frac{\xi^2}{2 (1 - \eta)}}}{\sqrt{2 \pi (1 - \eta)}} \left[ f_\mathrm{out}(1, \sqrt{q} \xi, V) - f_\mathrm{out}(-1, \sqrt{q} \xi, V) \right], \label{eq: appendix - kernel SE of m_hat for the RAF data model}
    \\
    \hat{q} &= \frac{\mu_1^2 \alpha}{2} \int_\mathbb{R} \frac{\dd \xi}{\sqrt{2\pi}} e^{-\xi^2/2} \left[ f_\mathrm{out}(-1, \sqrt{q} \xi, V)^2 + f_\mathrm{out}(1, \sqrt{q} \xi, V)^2 \right] \nonumber
    \\
    &\quad + \frac{\mu_1^2 \alpha (1 - \varepsilon)}{2} \int_\mathbb{R} \frac{\dd \xi}{\sqrt{2\pi}} e^{-\xi^2/2} \operatorname{erf}\!\left( \frac{\sqrt{\eta} \xi}{\sqrt{2 (1 - \eta)}} \right) \left[ f_\mathrm{out}(1, \sqrt{q} \xi, V)^2 - f_\mathrm{out}(-1, \sqrt{q} \xi, V)^2 \right], \label{eq: appendix - kernel SE of q_hat for the RAF data model}
    \\
    \hat{V} &= - \frac{\mu_1^2 \alpha}{2} \int_\mathbb{R} \frac{\dd \xi}{\sqrt{2\pi}} e^{-\xi^2/2} \left[ \partial_\omega f_\mathrm{out}(-1, \sqrt{q} \xi, V) + \partial_\omega f_\mathrm{out}(1, \sqrt{q} \xi, V) \right] \nonumber
    \\ 
    &\quad - \frac{\mu_1^2 \alpha (1 - \varepsilon)}{2} \int_\mathbb{R} \frac{\dd \xi}{\sqrt{2\pi}} e^{-\xi^2/2} \operatorname{erf}\!\left( \frac{\sqrt{\eta} \xi}{\sqrt{2 (1 - \eta)}} \right) \left[ \partial_\omega f_\mathrm{out}(1, \sqrt{q} \xi, V) - \partial_\omega f_\mathrm{out}(-1, \sqrt{q} \xi, V) \right]. \label{eq: appendix - kernel SE of V_hat for the RAF data model}
\end{align}

\subsection{Kernel state equations for the RAF data model}
\label{app: kernel SE RAF}
For reference in the subsequent sections, we collect here the full set of kernel state equations for the RAF data model:
\begin{equation} \label{eq: appendix - SE kernel RAF}
\left\{
\begin{aligned}
& m = \mu_1 \frac{\hat{m}}{\lambda + \hat{V}} \\[4pt]
& q = \mu_1^2 \frac{\hat{m}^2 + \hat{q}}{(\lambda + \hat{V})^2} \\[4pt]
& V = \mu_1^2 \frac{1}{\lambda + \hat{V}} + \frac{\mu_\star^2}{\lambda} \\[4pt]
& \hat{m} = \mu_1 \alpha (1 - \varepsilon) \int_\mathbb{R} \frac{\dd \xi}{\sqrt{2\pi}} \frac{e^{- \frac{\xi^2}{2 (1 - \eta)}}}{\sqrt{2 \pi (1 - \eta)}} \left[ f_\mathrm{out}(1, \sqrt{q} \xi, V) - f_\mathrm{out}(-1, \sqrt{q} \xi, V) \right] \\[6pt]
& \hat{q} = \frac{\mu_1^2 \alpha}{2} \int_\mathbb{R} \frac{\dd \xi}{\sqrt{2\pi}} e^{-\xi^2/2} \left[ f_\mathrm{out}(-1, \sqrt{q} \xi, V)^2 + f_\mathrm{out}(1, \sqrt{q} \xi, V)^2 \right] \\ & \qquad + \frac{\mu_1^2 \alpha (1 - \varepsilon)}{2} \int_\mathbb{R} \frac{\dd \xi}{\sqrt{2\pi}} e^{-\xi^2/2} \operatorname{erf}\!\left( \frac{\sqrt{\eta} \xi}{\sqrt{2 (1 - \eta)}} \right) \left[ f_\mathrm{out}(1, \sqrt{q} \xi, V)^2 - f_\mathrm{out}(-1, \sqrt{q} \xi, V)^2 \right] \\[6pt]
& \hat{V} = - \frac{\mu_1^2 \alpha}{2} \int_\mathbb{R} \frac{\dd \xi}{\sqrt{2\pi}} e^{-\xi^2/2} \left[ \partial_\omega f_\mathrm{out}(-1, \sqrt{q} \xi, V) + \partial_\omega f_\mathrm{out}(1, \sqrt{q} \xi, V) \right] \\ & \qquad - \frac{\mu_1^2 \alpha (1 - \varepsilon)}{2} \int_\mathbb{R} \frac{\dd \xi}{\sqrt{2\pi}} e^{-\xi^2/2} \operatorname{erf}\!\left( \frac{\sqrt{\eta} \xi}{\sqrt{2 (1 - \eta)}} \right) \left[ \partial_\omega f_\mathrm{out}(1, \sqrt{q} \xi, V) - \partial_\omega f_\mathrm{out}(-1, \sqrt{q} \xi, V) \right]
\end{aligned}
\right.
\end{equation}
Upon identifying $\eta = m^2 / q$, these equations coincide with those given in \eqref{eq:rf-MQSigma-def} and \eqref{eq:RF-saddle-kernel} in the main text.

We observe that, to specialize the analysis to a given loss, in addition to computing, for the corresponding $f_\mathrm{out}$, the right-hand sides of Eqs.~\eqref{eq: appendix - kernel SE of m_hat for the RAF data model}--\eqref{eq: appendix - kernel SE of V_hat for the RAF data model}, one must also derive the explicit expression of the memorization error. Indeed, unlike the generalization error in \eqref{eq: appendix - generalization error formula}, the memorization error in \eqref{eq: appendix - memorization error formula} depends explicitly on the loss through the associated proximal operator.

The following sections carry out this program for the square and hinge losses, providing a detailed analysis of the resulting state equations and of the corresponding memorization and generalization errors evaluated at their solutions.

\subsection{Kernel ridge regression in the RAF task model}
\label{app: KRR RAF}
For the square loss $\ell_\mathrm{square}(y,z) = \frac{(y - z)^2}{2}$, the proximal operator admits the simple closed-form expression
\begin{equation}
    \mathcal{P}_{\tau \ell_\mathrm{square}(y, \cdot)}(\omega) = \frac{\omega + \tau y}{1 + \tau}.
\end{equation}
Consequently, both $f_\mathrm{out}^\mathrm{square}$ and its derivative $\partial_\mathrm{\omega} f_\mathrm{out}^\mathrm{square}$ simplify to
\begin{equation}
    f_\mathrm{out}^\mathrm{square}(y, \omega, \tau) = \frac{y - \omega}{1 + \tau}, \qquad
    \partial_\omega f_\mathrm{out}^\mathrm{square}(y, \omega, \tau) = - \frac{1}{1 + \tau}.
\end{equation}
Substituting these expressions into Eqs.~\eqref{eq: appendix - SE kernel RAF} and evaluating the resulting Gaussian integrals yields the full set of state equations describing kernel ridge regression on the RAF data model in the high-dimensional limit:
\begin{align}
  m &= \mu_1 \frac{\hat{m}}{\lambda + \hat{V}},
  &\quad
  q &= \mu_1^2 \frac{\hat{m}^2 + \hat{q}}{(\lambda +\hat{V})^2},
  &\quad
  V &= \frac{\mu_1^2}{\lambda + \hat{V}} + \frac{\mu_\star^2}{\lambda},
  \\
  \hat{m} &= \frac{\mu_1 \alpha (1 - \varepsilon) \sqrt{\frac{2}{\pi}}}{1 + V},
  &\quad
  \hat{q} &= \mu_1^2 \alpha \frac{1 + q - 2 \sqrt{\frac{2}{\pi}} (1-\varepsilon) m}{(1 + V)^2},
  &\quad
  \hat{V} &= \frac{\mu_1^2 \alpha}{1+V}.
\end{align}

Introducing the parameters $\ell = \lambda / \mu_1^2$ and $t = \mu_1 / \mu_\star$, and applying rescalings analogous to those in Appendix \ref{app:kernel-reduction}, the above system can be rewritten in the reduced form
\begin{align}
  m &= \frac{\tilde{m}}{\ell + \tilde{V}},
  &\quad
  q &= \frac{\tilde{m}^2 + \tilde{q}}{(\ell +\tilde{V})^2},
  &\quad
  V &= \frac{1}{\ell + \tilde{V}} + \frac{1}{t^2 \ell},
  \\
  \tilde{m} &= \frac{\alpha (1 - \varepsilon) \sqrt{\frac{2}{\pi}}}{1 + V},
  &\quad
  \tilde{q} &= \alpha \frac{1 + q - 2 \sqrt{\frac{2}{\pi}} (1-\varepsilon) m}{(1 + V)^2},
  &\quad
  \tilde{V} &= \frac{\alpha}{1 + V}.
\end{align}
These KRR state equations are sufficiently simple to admit an analytical solution. Combining the equations for $V$ and $\tilde{V}$ yields
\begin{equation} \label{eq: appendix - KRR closed equation for V}
    V = \frac{1 + \frac{1}{t^2} - \alpha - \ell + \sqrt{\left( \alpha - 1 - \ell - \frac{1}{t^2} \right)^2 + 4 \alpha \left( \ell + \frac{1}{t^2} \right)}}{2 \ell}.
\end{equation}
All remaining variables can then be expressed in terms of $V$. In particular, the overlaps $m$ and $q$ are given by
\begin{align}
    \label{eq: appendix - KRR m}
    m &= \frac{\alpha (1 - \varepsilon) \sqrt{\frac{2}{\pi}}}{\alpha + \ell (1 + V)},
    \\[6pt]
    \label{eq: appendix - KRR q}
    q &= \frac{\alpha \left[ 1 + \frac{2 \alpha (1 - \varepsilon)^2}{\pi} \frac{\alpha + \ell (1 + V) - 2}{\alpha + \ell (1 + V)} \right]}{\left[ \alpha + \ell (1 + V) \right]^2 - \alpha} = \frac{\left[ 1 + \frac{2 \alpha (1 - \varepsilon)^2}{\pi} - 2 \sqrt{\frac{2}{\pi}} (1 - \varepsilon) m \right] m^2}{\frac{2 \alpha (1 - \varepsilon)^2}{\pi} - m^2}.
\end{align}

For the square loss, the formula for the memorization error also simplifies considerably. Since $1 + V > 0$, we have
\begin{equation}
    \mathcal{P}_{V \ell_\mathrm{square}(-1,\cdot)}(\sqrt{q}\xi) > 0 \quad \Longleftrightarrow \quad \xi > V / \sqrt{q}, \qquad\mathcal{P}_{V \ell_\mathrm{square}(1,\cdot)}(\sqrt{q}\xi) < 0 \quad \Longleftrightarrow \quad \xi < - V / \sqrt{q}.
\end{equation}
Thus, evaluating the right-hand side of Eq.~\eqref{eq: appendix - memorization error formula} amounts to computing a Gaussian tail integral, leading to
\begin{equation} \label{eq: appendix - square memorization error}
    \mathcal{E}_\mathrm{mem}^\mathrm{square} = \int_{\frac{V}{\sqrt{q}}}^\infty \frac{\dd \xi}{\sqrt{2 \pi}} e^{-\xi^2 / 2} = \frac{1}{2}\,\operatorname{erfc}\!\left( \frac{V}{\sqrt{2q}} \right).
\end{equation}

\subsubsection{Regularization limits and optimal regularization in KRR}
\label{app: reg limits and opt lambda KRR}
The closed-form solution for the overlap order parameters in \eqref{eq: appendix - KRR closed equation for V}--\eqref{eq: appendix - KRR q} permits an explicit evaluation of the memorization and generalization errors of KRR in notable limiting regimes of the regularization strength $\lambda$.

\paragraph{Ridgeless limit for non-linear kernels}
We begin by considering the limit $\lambda \to 0^+$ for any $\mu_\star > 0$. Expanding the right-hand side of Eq.~\eqref{eq: appendix - KRR closed equation for V} in the limit $\ell \to 0^+$ (noting that $\lambda$ enters only through $\ell = \lambda / \mu_1^2$), we find that the order parameter $V$ behaves as
\begin{equation}
    V \underset{\ell \to 0^+}{\simeq} \frac{1 + \frac{1}{t^2} - \alpha + \sqrt{\left(\alpha - 1 - \frac{1}{t^2} \right)^2 + \frac{4 \alpha}{t^2}}}{2 \ell} = \frac{V_0(\alpha, t)}{\ell}
\end{equation}
with
\begin{equation}
    V_0(\alpha, t) = \frac{1 + \frac{1}{t^2} - \alpha + \sqrt{\left(\alpha - 1 - \frac{1}{t^2} \right)^2 + \frac{4 \alpha}{t^2}}}{2}.
\end{equation}
The corresponding limiting expressions for $m$ and $q$ then follow from Eqs.~\eqref{eq: appendix - KRR m} and \eqref{eq: appendix - KRR q}:
\begin{align}
    m &\underset{\ell \to 0^+}{\longrightarrow} \frac{\alpha (1 - \varepsilon) \sqrt{\frac{2}{\pi}}}{\alpha + V_0(\alpha, t)}, \label{eq: appendix - KRR m zero lambda}
    \\[4pt]
    q &\underset{\ell \to 0^+}{\longrightarrow} \frac{\alpha \left[ 1 + \frac{2 \alpha (1 - \varepsilon)^2}{\pi} \frac{\alpha + V_0(\alpha, t) - 2}{\alpha + V_0(\alpha, t)} \right]}{\left( \alpha + V_0(\alpha, t) \right)^2 - \alpha}. \label{eq: appendix - KRR q zero lambda}
\end{align}

Since $q$ remains finite as $\lambda \to 0^+$ while $V$ diverges, the memorization error in Eq.~\eqref{eq: appendix - square memorization error} vanishes for all $\alpha$: non-linear kernels achieve perfect memorization in the ridgeless limit at any sample complexity.

For the generalization error, the limiting expression is less immediate. Plugging the limiting forms of $m$ and $q$ in Eqs.~\eqref{eq: appendix - KRR m zero lambda} and \eqref{eq: appendix - KRR q zero lambda} into Eq.~\eqref{eq: appendix - generalization error formula}, we obtain
\begin{equation}
\begin{split}
    &\lim_{\ell \to 0^+} \mathcal{E}_\mathrm{gen}^\mathrm{square}(\alpha, \varepsilon, \ell, t) =
    \\
    &\quad = \frac{1}{\pi} \arccos\!\left[ (1 - \varepsilon) \sqrt{\frac{\frac{2 \alpha}{\pi} ( (\alpha + V_0(\alpha, t))^2 - \alpha )}{(\alpha + V_0(\alpha, t)) \left[ \alpha + V_0(\alpha, t) + \frac{2 \alpha (1 - \varepsilon)^2}{\pi} (\alpha + V_0(\alpha, t) - 2) \right]}} \, \right].
\end{split}
\end{equation}
Upon substituting the explicit expression for $V_0(\alpha, t)$, this becomes a function of only $\alpha$, $\varepsilon$, and $t = \mu_1 / \mu_\star$. A lengthy but straightforward calculation shows that the value of $t$ minimizing the ridgeless KRR generalization error is
\begin{equation} \label{eq: appendix - t opt mem}
    t_\mathrm{mem}^\mathrm{opt}(\varepsilon) = \left( \frac{\pi}{2(1-\varepsilon)^2}-1 \right)^{-1/2},
\end{equation}
for $0 \leq \varepsilon < 1$. This value determines the optimal angle \eqref{eq:angle_opt_mem} introduced in the main text:
\begin{equation} \label{eq: appendix - gamma opt mem}
    \gamma_\mathrm{mem}^\mathrm{opt}(\varepsilon) = \arctan\!\left[ t_\mathrm{mem}^\mathrm{opt}(\varepsilon) \right] = \arctan\!\left[ \left( \frac{\pi}{2(1-\varepsilon)^2}-1 \right)^{-1/2} \right].
\end{equation}
Importantly, this angle depends on $\varepsilon$ but not on $\alpha$.

\paragraph{Ridgeless limit for the linear model}
Setting $\mu_1 = 1$ and $\mu_\star = 0$ in Eqs.~\eqref{eq: appendix - KRR closed equation for V}--\eqref{eq: appendix - KRR q} yields the order parameters characterizing the asymptotic performance of ridge regression. In particular, the order parameter $V$ is given by
\begin{equation}
    \label{eq: appendix - V finite lambda perceptron square loss}
    V = \frac{1 - \alpha - \lambda + \sqrt{(\alpha - 1 - \lambda)^2 + 4 \alpha \lambda}}{2 \lambda}.
\end{equation}
Expanding in the limit $\lambda \to 0^+$, we obtain
\begin{equation}
    V \underset{\lambda \to 0^+}{\simeq} \frac{|\alpha - 1| + 1 - \alpha}{2 \lambda} + \frac{1}{2} \left( \frac{\alpha + 1}{|\alpha - 1|} - 1 \right).
\end{equation}
The resulting behavior depends on whether $0 < \alpha < 1$ or $\alpha > 1$. For $0 < \alpha < 1$, we have
\begin{equation}
    V \underset{\lambda \to 0^+}{\simeq} \frac{1 - \alpha}{\lambda} + \frac{\alpha}{1 - \alpha}.
\end{equation}
This yields the following closed expressions for the overlap order parameters:
\begin{equation} 
\begin{split}
    V &\underset{\lambda \to 0^+}{\simeq} \frac{1 - \alpha}{\lambda},
    \\[4pt]
    m &\underset{\lambda \to 0^+}{\longrightarrow} \alpha (1 - \varepsilon) \sqrt{\frac{2}{\pi}},
    \\[4pt]
    q &\underset{\lambda \to 0^+}{\longrightarrow} \alpha \, \frac{\pi - 2 \alpha (1 - \varepsilon)^2}{\pi (1 - \alpha)}.
\end{split}
\end{equation}
For $\alpha > 1$, instead, we obtain
\begin{equation}
    V \underset{\lambda \to 0^+}{\longrightarrow} \frac{1}{\alpha - 1}.
\end{equation}
The corresponding overlap order parameters are
\begin{equation} 
\begin{split}
    V &\underset{\lambda \to 0^+}{\longrightarrow} \frac{1}{\alpha - 1},
    \\[4pt]
    m &\underset{\lambda \to 0^+}{\longrightarrow} (1 - \varepsilon) \sqrt{\frac{2}{\pi}},
    \\[4pt]
    q &\underset{\lambda \to 0^+}{\longrightarrow} \frac{1 + \frac{2}{\pi} (1 - \varepsilon)^2 (\alpha - 2)}{\alpha - 1}.
\end{split}
\end{equation}

Combining these results, we obtain the limiting memorization and generalization errors as $\lambda \to 0^+$, depending on the value of $\alpha$:
\begin{align}
    \lim_{\lambda \to 0^+} \mathcal{E}_\mathrm{mem}^\mathrm{square, lin}(\alpha, \varepsilon, \lambda) &= \begin{cases}
    0 & \text{if } 0 < \alpha \le 1, \\[4pt]
    \frac{1}{2}\operatorname{erfc}\!\left(\sqrt{\frac{\pi}{2 (\alpha - 1)\left( \pi + 2 (1 - \varepsilon)^2 (\alpha - 2)\right)}} \right) & \text{if } \alpha > 1,
    \end{cases} \label{eq: memorization error perceptron pseudo-inverse}
    \\[10pt]
    \lim_{\lambda \to 0^+} \mathcal{E}_\mathrm{gen}^\mathrm{square, lin}(\alpha, \varepsilon, \lambda) &= \begin{cases}
    \frac{1}{\pi}\arccos\!\left((1 - \varepsilon)\sqrt{\frac{2 \alpha (1 - \alpha)}{\pi - 2 \alpha (1 - \varepsilon)^2}} \,\right) & \text{if } 0 < \alpha \le 1, \\[8pt]
    \frac{1}{\pi}\arccos\!\left( (1 - \varepsilon)\sqrt{\frac{2 (\alpha - 1)}{\pi + 2 (\alpha - 2) (1 - \varepsilon)^2}} \,\right) & \text{if } \alpha > 1.
    \end{cases} \quad \label{eq: generalization error perceptron pseudo-inverse}
\end{align}

Eq.~\eqref{eq: memorization error perceptron pseudo-inverse} shows that, for any $0 < \varepsilon \leq 1$, ridgeless linear regression perfectly memorizes the randomly labeled training samples when $\alpha \leq 1$. By contrast, for $\alpha > 1$, memorization deteriorates: the memorization error increases with $\alpha$ and approaches $1/2$ in the limit $\alpha \to \infty$. The critical sample complexity $\alpha_c = 1$ coincides with the interpolation threshold of linear regression with square loss. Notably, this threshold is independent of $\varepsilon$.

Eq.~\eqref{eq: generalization error perceptron pseudo-inverse} shows that the generalization error of the ridgeless linear estimator reaches the random-guessing value $1/2$ at the interpolation threshold $\alpha_c = 1$. For $0 \leq \varepsilon < 1$, it subsequently decreases for $\alpha > 1$ and vanishes as $\alpha \to \infty$.

\paragraph{Optimal regularization}
The optimal regularization strength $\lambda_\mathrm{opt}$, defined as the value that minimizes the generalization error in Eq.~\eqref{eq: appendix - generalization error formula} at fixed $\alpha$, $0 \leq \varepsilon < 1$, $\mu_1$ and $\mu_\star$, can also be determined analytically for KRR. Since $m$ and $q$ depend on $\lambda$ only through $\ell = \lambda / \mu_1^2$, minimizing the generalization error with respect to $\lambda$ is equivalent to minimizing it with respect to $\ell$. A direct calculation yields the unique stationary point
\begin{equation} \label{eq: appendix - KRR ell optimal}
    \ell_\mathrm{stat}(\varepsilon, t) = \frac{\pi}{2 (1 - \varepsilon)^2} - 1 - \frac{1}{t^2}.
\end{equation}
Moreover,
\begin{equation}
    \partial_\ell \left( \frac{m}{\sqrt{q}} \right) \gtrless 0 \quad\Longleftrightarrow \quad \ell \lessgtr \ell_\mathrm{stat}(\varepsilon, t).
\end{equation}
Therefore, $m / \sqrt{q}$ is maximized, and hence the generalization error is minimized, at $\ell = \ell_\mathrm{stat}(\varepsilon, t)$.

The regularization strength is, however, constrained to satisfy $\lambda > 0$ and therefore $\ell$ must also be positive. If $\ell_\mathrm{stat}(\varepsilon, t) \leq 0$, the stationary point lies outside the admissible interval $(0, \infty)$. In this case, the infimum of the generalization error is approached as $\ell \to 0^+$, or equivalently in the ridgeless limit $\lambda \to 0^+$. We thus define the constrained optimal value of $\ell$ as
\begin{equation}
    \ell_\mathrm{opt}(\varepsilon, t) = \max\!\left\{ \ell_\mathrm{stat}(\varepsilon, t), \; 0^+ \right\}.
\end{equation}
The corresponding optimal regularization strength is
\begin{equation}
\label{Eq:app-lambda-opt}
\lambda_\mathrm{opt}(\varepsilon, \mu_1, \mu_\star) = \mu_1^2\ell_\mathrm{opt}(\varepsilon,t) = \max\!\left\{ \mu_1^2 \left( \frac{\pi}{2 (1 - \varepsilon)^2} - 1 \right) - \mu_\star^2, \; 0^+ \right\}.
\end{equation}
Here $\max\{a,0^+\}$ is understood in the limiting sense:
\begin{equation}
    \max \{a, 0^+\} = \begin{cases}
        a &\text{ if } a > 0, \\
        0^+ &\text{ if } a \leq 0,
    \end{cases}
\end{equation}
where $0^+$ denotes the limiting choice $\lambda \to 0^+$.

The optimal regularization strength is strictly positive if and only if
\begin{equation}
    \mu_1 > \left( \frac{\pi}{2(1-\varepsilon)^2}-1 \right)^{-1/2} \mu_\star.
\end{equation}
Equivalently, using \eqref{eq: appendix - t opt mem}, this condition can be written as
\begin{equation}
    t = \frac{\mu_1}{\mu_\star} > t_\mathrm{mem}^\mathrm{opt}(\varepsilon),
\end{equation}
or, in terms of the angle $\gamma_\mathrm{mem}^\mathrm{opt}(\varepsilon)$ in \eqref{eq: appendix - gamma opt mem}, as
\begin{equation}
    \gamma = \arctan\!\left( \frac{\mu_1}{\mu_\star} \right) > \gamma_\mathrm{mem}^\mathrm{opt}(\varepsilon).
\end{equation}
Thus, at fixed $\varepsilon$, kernels whose angle $\gamma$ in the $(\mu_\star, \mu_1)$ plane exceeds $\gamma_\mathrm{mem}^\mathrm{opt}(\varepsilon)$ have a strictly positive optimal regularization strength, given by $\lambda_\mathrm{opt} = \mu_1^2 \ell_\mathrm{stat}$. Evaluating the order parameters \eqref{eq: appendix - KRR closed equation for V}--\eqref{eq: appendix - KRR q} at this stationary point yields
\begin{align}
    V(\alpha, \varepsilon, \ell_\mathrm{stat}(\varepsilon, t), t) &= \frac{2 \left(1 + \frac{1}{t^2} \right) - \alpha - \frac{\pi}{2 (1 - \varepsilon)^2} + \sqrt{\left( \alpha + \frac{\pi}{2 (1 - \varepsilon)^2} \right)^2 - 4 \alpha}}{2 \left( \frac{\pi}{2 (1 - \varepsilon)^2} - 1 - \frac{1}{t^2} \right)},
    \\[6pt]
    m(\alpha, \varepsilon, \ell_\mathrm{stat}(\varepsilon, t), t) &= \frac{1 - \varepsilon}{\sqrt{2 \pi}} \left[ \alpha + \frac{\pi}{2 (1 - \varepsilon)^2} - \sqrt{\left( \alpha + \frac{\pi}{2 (1 - \varepsilon)^2} \right)^2 - 4 \alpha} \, \right],
    \\[6pt]
    q(\alpha, \varepsilon, \ell_\mathrm{stat}(\varepsilon, t), t) &= \frac{(1 - \varepsilon)^2}{\pi} \left[ \alpha + \frac{\pi}{2 (1 - \varepsilon)^2} - \sqrt{\left( \alpha + \frac{\pi}{2 (1 - \varepsilon)^2} \right)^2 - 4 \alpha} \, \right].
\end{align}
At $\ell = \ell_\mathrm{stat}$, $V$ retains an explicit dependence on $t$, while $m$ and $q$ depend only on $\alpha$ and $\varepsilon$. Consequently, the generalization error evaluated at $\ell_\mathrm{stat}$ is independent of $t$:
\begin{equation}
    \mathcal{E}_\mathrm{gen}^\mathrm{square}(\alpha, \varepsilon, \ell_\mathrm{stat}(\varepsilon, t), t) = \frac{1}{\pi} \arccos\!\left[ \sqrt{ \frac{1}{2} \left( \alpha + \frac{\pi}{2 (1 - \varepsilon)^2} - \sqrt{\left( \alpha + \frac{\pi}{2 (1 - \varepsilon)^2} \right)^2 - 4 \alpha} \, \right) } \, \right].
\end{equation}
Moreover, this constant is exactly equal to $\mathcal{E}_\mathrm{gen}^\mathrm{square}(\alpha, \varepsilon, 0^+, t_\mathrm{mem}^\mathrm{opt}(\varepsilon))$, as expected from the continuity of $\mathcal{E}_\mathrm{gen}^\mathrm{square}(\alpha, \varepsilon, \ell_\mathrm{opt}(\varepsilon, t), t)$ at $t = t_\mathrm{mem}^\mathrm{opt}(\varepsilon)$.

The memorization and generalization errors at the constrained optimum can therefore be summarized as
\begin{equation}
    \mathcal{E}_\mathrm{mem}^\mathrm{square}(\alpha, \varepsilon, \ell_\mathrm{opt}(\varepsilon, t), t) = \begin{cases}
        0 &\text{ if } t \leq t_\mathrm{mem}^\mathrm{opt}(\varepsilon), \\[4pt]
        \mathcal{E}_\mathrm{mem}^\mathrm{square}(\alpha, \varepsilon, \ell_\mathrm{stat}(\varepsilon, t), t) &\text{ if } t > t_\mathrm{mem}^\mathrm{opt}(\varepsilon),
    \end{cases}
\end{equation}
and
\begin{equation}
    \mathcal{E}_\mathrm{gen}^\mathrm{square}(\alpha, \varepsilon, \ell_\mathrm{opt}(\varepsilon, t), t) = \begin{cases}
        \mathcal{E}_\mathrm{gen}^\mathrm{square}(\alpha, \varepsilon, 0^+, t) &\text{ if } t \leq t_\mathrm{mem}^\mathrm{opt}(\varepsilon), \\[4pt]
        \mathcal{E}_\mathrm{gen}^\mathrm{square}(\alpha, \varepsilon, 0^+, t_\mathrm{mem}^\mathrm{opt}(\varepsilon)) &\text{ if } t > t_\mathrm{mem}^\mathrm{opt}(\varepsilon).
    \end{cases}
\end{equation}

\paragraph{Infinite-regularization limit}
We now derive the leading-order behavior of the order parameters in the infinite-regularization limit and the corresponding limiting values of the memorization and generalization errors. Starting from the finite-$\ell$ state equations, we expand the right-hand side of Eq.~\eqref{eq: appendix - KRR closed equation for V} for $\ell \to +\infty$ and then use Eqs.~\eqref{eq: appendix - KRR m}–\eqref{eq: appendix - KRR q}. This gives
\begin{equation}
\begin{split}
    V &\underset{\ell \to +\infty}{\simeq} \frac{1 + \frac{1}{t^2}}{\ell},
    \\[4pt]
    m &\underset{\ell \to +\infty}{\simeq} \frac{\alpha (1 - \varepsilon) \sqrt{\frac{2}{\pi}}}{\ell},
    \\[4pt]
    q &\underset{\ell \to +\infty}{\simeq} \frac{(1 - \varepsilon)^2 \frac{2 \alpha^2}{\pi} + \alpha}{\ell^2}.
\end{split}
\end{equation}
The corresponding limiting memorization and generalization errors are therefore
\begin{equation}
    \lim_{\ell \to +\infty} \mathcal{E}_\mathrm{mem}(\alpha, \varepsilon, \ell, t) = \frac{1}{2} \operatorname{erfc}\!\left( \frac{1 + \frac{1}{t^2}}{\sqrt{2 \left( \alpha (1 - \varepsilon) \sqrt{\frac{2}{\pi}} \right)^2 + 2 \alpha}} \right),
\end{equation}
and
\begin{equation}
    \lim_{\ell \to +\infty} \mathcal{E}_\mathrm{gen}(\alpha, \varepsilon, \ell, t) = \frac{1}{\pi} \arccos\!\left( \frac{(1 - \varepsilon) \sqrt{\frac{2 \alpha}{\pi}}}{\sqrt{1 + (1 - \varepsilon)^2 \frac{2 \alpha}{\pi}}} \right).
\end{equation}

Although these expressions have been obtained from the finite-regularization state equations of kernel ridge regression, the resulting limiting classification errors are the same for the hinge loss. More generally, this loss independence is expected to hold for sufficiently regular classification losses whose leading response around a vanishing prediction differs only by an overall positive factor. Indeed, as the regularization strength diverges, the estimator converges to zero because any nonzero norm is increasingly penalized. Its leading-order direction, however, remains data dependent and is determined by the direction along which the empirical loss initially decreases away from the origin. Changing the loss can modify the overall magnitude of this leading response, but not its direction. Since classification errors depend only on the signs of the predicted scores, this loss-dependent positive factor cancels. The scores therefore become arbitrarily small while retaining the same nontrivial pattern of signs, which explains both why the limiting errors remain nontrivial and why their values are common to all the losses considered here.

\subsubsection{Large-\texorpdfstring{$\alpha$}{alpha} asymptotics of the KRR generalization error at fixed \texorpdfstring{$\lambda$}{lambda}}
\label{app:kernel_large_alpha}
We now derive the large-$\alpha$ behavior of the KRR generalization error at fixed $0 \leq \varepsilon < 1$, fixed regularization strength $\lambda > 0$, and fixed kernel parameters $\mu_1$ and $\mu_\star$.

The order parameters $m$ and $q$ have the asymptotic expansions
\begin{equation}
    m \underset{\alpha \to +\infty}{\simeq} (1 - \varepsilon) \sqrt{\frac{2}{\pi}} - (1 - \varepsilon) \sqrt{\frac{2}{\pi}} \left( \ell + \frac{1}{t^2} \right) \alpha^{-1}
\end{equation}
and
\begin{equation}
    q \underset{\alpha \to +\infty}{\simeq} \frac{2 (1 - \varepsilon)^2}{\pi} + \left[ 1 - \frac{2 (1 - \varepsilon)^2}{\pi} - \frac{4 (1 - \varepsilon)^2}{\pi} \left( \ell + \frac{1}{t^2} \right) \right] \alpha^{-1}.
\end{equation}
Although the first corrections to both $m$ and $q$ depend on the regularization strength and the kernel parameters through $\ell = \lambda / \mu_1^2$ and $t = \mu_1 / \mu_\star$, this dependence cancels in the leading correction to the normalized overlap. Indeed,
\begin{equation}
    \frac{m}{\sqrt{q}} \underset{\alpha \to +\infty}{\simeq} 1 - \frac{\pi - 2 (1 - \varepsilon)^2}{4 (1 - \varepsilon)^2} \, \alpha^{-1}.
\end{equation}
It follows that
\begin{equation}
    \mathcal{E}_\mathrm{gen}^\mathrm{square}(\alpha, \varepsilon, \ell, t) = \frac{1}{\pi} \arccos\!\left( \frac{m}{\sqrt{q}} \right) \underset{\alpha \to +\infty}{\simeq} \frac{\sqrt{\pi - 2 (1 - \varepsilon)^2}}{\sqrt{2} \pi (1 - \varepsilon)} \, \alpha^{- 1 / 2}.
\end{equation}
This establishes Eq.~\eqref{eq:square_large_alpha_rate_results} in the main text.

For any fixed $\lambda > 0$ and fixed kernel parameters $\mu_1, \mu_\star$, the leading large-$\alpha$ behavior of the KRR generalization error is independent of the regularization strength and the kernel parameters and decays universally as $\alpha^{-1/2}$. In particular, for $\varepsilon = 0$, the resulting $\alpha^{-1/2}$ decay agrees with the classical large-sample asymptotics of the ADALINE rule for the perceptron \cite{engel2001statistical}, which is related to the ridgeless linear model trained with the square loss considered here.

\subsection{Support vector machines in the RAF task model}
\label{app: SVM RAF}
For the hinge loss $\ell_\mathrm{hinge}(y, z) = \max(0, 1 - yz)$, the proximal operator is a piecewise function:
\begin{equation}\label{eq: appendix - hinge proximal}
\mathcal{P}_{\tau \ell_{\mathrm{hinge}}(y, \cdot)}(\omega) =
\begin{cases}
\omega + \tau y & \text{if } y \omega < 1 - \tau, \\[4pt]
y & \text{if } 1 - \tau < y \omega < 1, \\[4pt]
\omega & \text{if } y \omega > 1.
\end{cases}
\end{equation}
Therefore, $f_\mathrm{out}^\mathrm{hinge}$ and its derivative $\partial_\mathrm{\omega} f_\mathrm{out}^\mathrm{hinge}$ are also piecewise:
\begin{equation}
\begin{split}
    f_\mathrm{out}^\mathrm{hinge}(y, \omega, \tau) &= \begin{cases}
    y & \text{if } y \omega < 1 - \tau, \\[4pt]
    \frac{y - \omega}{\tau} & \text{if } 1 - \tau < y \omega < 1, \\[4pt]
    0 & \text{if } y \omega > 1,
    \end{cases}
    \\[6pt]
    \partial_\omega f_\mathrm{out}^\mathrm{hinge}(y, \omega, \tau) &= \begin{cases}
    - \frac{1}{\tau} & \text{if } 1 - \tau < y \omega < 1, \\[4pt]
    0 & \text{otherwise}.
    \end{cases}
\end{split}
\end{equation}
Substituting these expressions into Eqs.~\eqref{eq: appendix - SE kernel RAF} and using standard Gaussian integration results, we obtain the full set of state equations describing support vector machines in the RAF model in the high-dimensional limit:
\begin{equation} \label{eq: appendix - SE SVM overlaps}
  m = \mu_1 \frac{\hat{m}}{\lambda + \hat{V}},
  \qquad
  q = \mu_1^2 \frac{\hat{m}^2 + \hat{q}}{(\lambda +\hat{V} )^2},
  \qquad
  V =  \frac{\mu_1^2}{\lambda + \hat{V}} + \frac{\mu_\star^2}{\lambda},
\end{equation}

\begin{align} \label{eq: appendix - SE SVM hat variables}
    \hat{m} &= \frac{\mu_1 \alpha (1 - \varepsilon)}{\sqrt{2\pi}} \Bigg[ 1 + \frac{1}{V} \left( \operatorname{erf}\!\left( \frac{1}{\sqrt{2 q (1 - \eta)}} \right) - (1 - V) \operatorname{erf}\!\left( \frac{1 - V}{\sqrt{2 q (1 - \eta)}} \right) \right) \nonumber
    \\
    &\qquad + \frac{1}{V} \sqrt{\frac{2 q (1 - \eta)}{\pi}} \left( e^{- \frac{1}{2 q (1 - \eta)}} - e^{- \frac{(1 - V)^2}{2 q (1 - \eta)}} \right) \Biggr], \nonumber
    \\ \nonumber
    \\
    \hat{q} &= \frac{\mu_1^2 \alpha}{2} \left( 1 + \operatorname{erf}\!\left( \frac{1 - V}{\sqrt{2 q}} \right) \right) \nonumber
    \\
    &\quad + \frac{\mu_1^2 \alpha}{V^2} \Biggl[ \frac{1 + q}{2} \left( \operatorname{erf}\!\left( \frac{1}{\sqrt{2 q}} \right) - \operatorname{erf}\!\left( \frac{1 - V}{\sqrt{2 q}} \right) \right) + \sqrt{\frac{q}{2 \pi}} \left( e^{- \frac{1}{2 q}} - (1 + V) e^{- \frac{(1 - V)^2}{2 q}} \right) \Biggr]
    \\
    &\quad + \mu_1^2 \alpha (1 - \varepsilon) \Biggl[ - \frac{1}{\pi} \arctan\!\left( \sqrt{\frac{\eta}{1 - \eta}} \right) - \int_{- \frac{1 - V}{\sqrt{q}}}^0 \frac{\dd \xi}{\sqrt{2\pi}} e^{-\xi^2/2} \operatorname{erf}\!\left( \frac{\sqrt{\eta} \xi}{\sqrt{2 (1 - \eta)}} \right) \nonumber
    \\
    &\quad + \frac{1}{V^2} \int_{\frac{1 - V}{\sqrt{q}}}^{\frac{1}{\sqrt{q}}} \frac{\dd \xi}{\sqrt{2\pi}} e^{-\xi^2/2} \operatorname{erf}\!\left( \frac{\sqrt{\eta} \xi}{\sqrt{2 (1 - \eta)}} \right) (1 - \sqrt{q} \xi)^2 \Biggr], \nonumber
    \\ \nonumber
    \\
    \hat{V} &= \frac{\mu_1^2 \alpha}{V} \left[ \frac{1}{2} \left( \operatorname{erf}\!\left( \frac{1}{\sqrt{2 q}} \right) - \operatorname{erf}\!\left( \frac{1 - V}{\sqrt{2 q}} \right) \right) + (1 - \varepsilon) \int_{\frac{1 - V}{\sqrt{q}}}^{\frac{1}{\sqrt{q}}} \dd \xi \, \frac{e^{-\xi^2 / 2}}{\sqrt{2 \pi}} \operatorname{erf}\!\left( \frac{\sqrt{\eta} \xi}{\sqrt{2 (1 - \eta)}} \right) \right], \nonumber
\end{align}
where, as a reminder, $\eta = m^2 / q$.

We can recast these equations in their reduced form introducing $\ell = \lambda/\mu_1^2$ and $t=\mu_1/\mu_\star$, and applying rescalings analogous to those in Appendix \ref{app:kernel-reduction}:
\begin{equation}
    m = \frac{\tilde{m}}{\ell+\tilde{V}},
    \qquad
    q = \frac{\tilde{m}^2+\tilde{q}}{(\ell+\tilde{V})^2},
    \qquad
    V = \frac{1}{\ell+\tilde{V}} + \frac{1}{t^2 \ell},
\end{equation}

\begin{equation}
\begin{split}
    \tilde{m} &= \frac{\alpha (1 - \varepsilon)}{\sqrt{2\pi}} \Bigg[ 1 + \frac{1}{V} \left( \operatorname{erf}\!\left( \frac{1}{\sqrt{2 q (1 - \eta)}} \right) - (1 - V) \operatorname{erf}\!\left( \frac{1 - V}{\sqrt{2 q (1 - \eta)}} \right) \right)
    \\
    &\qquad + \frac{1}{V} \sqrt{\frac{2 q (1 - \eta)}{\pi}} \left( e^{- \frac{1}{2 q (1 - \eta)}} - e^{- \frac{(1 - V)^2}{2 q (1 - \eta)}} \right) \Biggr],
    \\
    \\
    \tilde{q} &= \frac{\alpha}{2} \left( 1 + \operatorname{erf}\!\left( \frac{1 - V}{\sqrt{2 q}} \right) \right)
    \\
    &\quad + \frac{\alpha}{V^2} \Biggl[ \frac{1 + q}{2} \left( \operatorname{erf}\!\left( \frac{1}{\sqrt{2 q}} \right) - \operatorname{erf}\!\left( \frac{1 - V}{\sqrt{2 q}} \right) \right) + \sqrt{\frac{q}{2 \pi}} \left( e^{- \frac{1}{2 q}} - (1 + V) e^{- \frac{(1 - V)^2}{2 q}} \right) \Biggr]
    \\
    &\quad + \alpha (1 - \varepsilon) \Biggl[ - \frac{1}{\pi} \arctan\!\left( \sqrt{\frac{\eta}{1 - \eta}} \right) - \int_{- \frac{1 - V}{\sqrt{q}}}^0 \frac{\dd \xi}{\sqrt{2\pi}} e^{-\xi^2/2} \operatorname{erf}\!\left( \frac{\sqrt{\eta} \xi}{\sqrt{2 (1 - \eta)}} \right)
    \\
    &\quad + \frac{1}{V^2} \int_{\frac{1 - V}{\sqrt{q}}}^{\frac{1}{\sqrt{q}}} \frac{\dd \xi}{\sqrt{2\pi}} e^{-\xi^2/2} \operatorname{erf}\!\left( \frac{\sqrt{\eta} \xi}{\sqrt{2 (1 - \eta)}} \right) (1 - \sqrt{q} \xi)^2 \Biggr],
    \\
    \\
    \tilde{V} &= \frac{\alpha}{V} \left[ \frac{1}{2} \left( \operatorname{erf}\!\left( \frac{1}{\sqrt{2 q}} \right) - \operatorname{erf}\!\left( \frac{1 - V}{\sqrt{2 q}} \right) \right) + (1 - \varepsilon) \int_{\frac{1 - V}{\sqrt{q}}}^{\frac{1}{\sqrt{q}}} \dd \xi \, \frac{e^{-\xi^2 / 2}}{\sqrt{2 \pi}} \operatorname{erf}\!\left( \frac{\sqrt{\eta} \xi}{\sqrt{2 (1 - \eta)}} \right) \right].
\end{split}
\end{equation}
In contrast to KRR, these state equations cannot be closed analytically.

Notably, the memorization error for the hinge loss reduces to the same form as in the square-loss case. The key observation is that
\begin{equation}
    \mathcal{P}_{V \ell_\mathrm{hinge}(-1,\cdot)}(\sqrt{q}\xi) > 0 \quad \Longleftrightarrow \quad \xi > V / \sqrt{q}, \qquad\mathcal{P}_{V \ell_\mathrm{hinge}(1,\cdot)}(\sqrt{q}\xi) < 0 \quad \Longleftrightarrow \quad \xi < - V / \sqrt{q}
\end{equation}
which are precisely the same conditions obtained for the square loss. Hence,
\begin{equation} \label{eq: appendix - hinge memorization error}
    \mathcal{E}_\mathrm{mem}^\mathrm{hinge} = \frac{1}{2}\,\operatorname{erfc}\!\left( \frac{V}{\sqrt{2q}} \right).
\end{equation}

\subsubsection{Regularization limits and optimal regularization in SVMs}
\label{app: reg limits and opt lambda SVM}
Unlike KRR, the SVMs state equations cannot be solved analytically, and explicit expressions for the generalization and memorization errors are therefore not generally available in the different regularization regimes. In particular, it is not possible to determine analytically the value of $\lambda$ that minimizes the generalization error. Nevertheless, the state equations can be evaluated numerically at finite $\lambda$. Moreover, by examining the scaling of their solutions as $\lambda \to 0^+$, one can derive reduced state equations directly characterizing the ridgeless limit and extract relevant information about the corresponding errors, in particular about the memorization error.

\paragraph{Ridgeless limit for non-linear kernels}
We investigate the ridgeless limit numerically by solving the state equations through a fixed-point iteration scheme at progressively smaller values of $\lambda$ and examining the resulting scaling of the overlap order parameters and their conjugates. For fixed $\alpha$, $\varepsilon$, and kernel parameters $\mu_1$ and $\mu_\star > 0$, the numerical solutions obey
\begin{equation} \label{eq: appendix - SVMs overlaps scaling for lambda to zero}
    m = \Theta(1),
    \quad
    q = \Theta(1),
    \quad
    V = \Theta(\lambda^{-1}),
    \quad
    \hat{m} = \Theta(\lambda),
    \quad
    \hat{q} = \Theta(\lambda^2),
    \quad
    \hat{V} = \Theta(\lambda).
\end{equation}
To obtain a non-trivial limiting system, we introduce the rescaled variables
\begin{equation}
    \overline V = \lambda V,
    \qquad
    \overline{\hat m} = \frac{\hat m}{\lambda},
    \qquad
    \overline{\hat q} = \frac{\hat q}{\lambda^2},
    \qquad
    \overline{\hat V} = \frac{\hat V}{\lambda}.
\end{equation}
Dropping the bars for notational simplicity, Eqs.~\eqref{eq: appendix - SE SVM overlaps}--\eqref{eq: appendix - SE SVM hat variables} reduce, as $\lambda\to0^+$, to
\begin{equation}
\begin{alignedat}{2}
    & m = \mu_1 \frac{\hat{m}}{1 + \hat{V}},
    &\qquad\qquad& \hat{m} = \frac{\mu_1 \alpha}{V} \mathcal{I}_{\hat{m}}(q, \eta), \\[6pt]
    & q = \mu_1^2 \frac{\hat{m}^2 + \hat{q}}{(1 + \hat{V})^2},
    && \hat{q} = \frac{\mu_1^2 \alpha}{V^2} \mathcal{I}_{\hat{q}}(q, \eta), \\[6pt]
    & V = \frac{\mu_1^2}{1 + \hat{V}} + \mu_\star^2,
    && \hat{V} = \frac{\mu_1^2 \alpha}{V} \mathcal{I}_{\hat{V}}(q, \eta),
\end{alignedat}
\end{equation}
where $\eta = m^2 / q$ and
\begin{equation} \label{eq: appendix - I hinge integrals lambda zero}
\begin{split}
    \mathcal{I}_{\hat{m}}(q, \eta) &= \frac{1 - \varepsilon}{\sqrt{2 \pi}} \left[ 1 + \operatorname{erf}\!\left( \frac{1}{\sqrt{2 q (1 - \eta)}} \right) + \sqrt{\frac{2 q (1 - \eta)}{\pi}} e^{- \frac{1}{2 q (1 - \eta)}} \right],
    \\
    \mathcal{I}_{\hat{q}}(q, \eta) &= \frac{1 + q}{2} \left( 1 + \operatorname{erf}\!\left( \frac{1}{\sqrt{2q}} \right) \right) + \sqrt{\frac{q}{2 \pi}} e^{-\frac{1}{2q}}
    \\
    &\quad + (1 - \varepsilon) \int_{-\infty}^{\frac{1}{\sqrt{q}}} \frac{\dd \xi}{\sqrt{2 \pi}} e^{-\xi^2/2} \operatorname{erf}\!\left( \frac{\sqrt{\eta} \xi}{\sqrt{2 (1 - \eta)}} \right) (1 - \sqrt{q} \xi)^2,
    \\
    \mathcal{I}_{\hat{V}}(q, \eta) &= \frac{1}{2} \left( 1 + \operatorname{erf}\!\left( \frac{1}{\sqrt{2 q}} \right) \right) + (1 - \varepsilon) \int_{-\infty}^{\frac{1}{\sqrt{q}}} \frac{\dd \xi}{\sqrt{2 \pi}} e^{-\xi^2/2} \operatorname{erf}\!\left( \frac{\sqrt{\eta} \xi}{\sqrt{2 (1 - \eta)}} \right).
\end{split}
\end{equation}
Here, $V$ denotes the rescaled quantity, which remains finite as $\lambda \to 0^+$, whereas the original overlap diverges proportionally to $\lambda^{-1}$. Since $q$ remains finite as $\lambda \to 0^+$ while the original $V$ diverges, the memorization error in \eqref{eq: appendix - hinge memorization error} vanishes for all $\alpha$. Thus, similarly to KRR, non-linear kernels achieve perfect memorization in the ridgeless limit at any sample complexity.

\paragraph{Ridgeless limit for the linear model}
We now consider the linear model, obtained by setting $\mu_1 = 1$ and $\mu_\star = 0$. In this case, the same numerical analysis reveals that, for any $0 < \varepsilon < 1$, there exists a critical sample complexity $\alpha_c(\varepsilon)$ separating two distinct regimes. Below this threshold, the fixed-point solutions obey the same small-$\lambda$ scalings \eqref{eq: appendix - SVMs overlaps scaling for lambda to zero} found for non-linear kernels. Above the threshold, instead, all the overlap order parameters and their conjugates remain finite as $\lambda \to 0^+$. Consequently, for $\alpha < \alpha_c(\varepsilon)$, $V$ diverges while $q$ remains finite, and the memorization error in Eq.~\eqref{eq: appendix - hinge memorization error} vanishes. For $\alpha > \alpha_c(\varepsilon)$, the divergence of $V$ is lost, and the linear model can no longer perfectly memorize the randomly labelled samples. The critical value $\alpha_c(\varepsilon)$ coincides with the interpolation threshold of linear regression with hinge loss. Unlike in the square-loss case, this threshold depends non-trivially on $\varepsilon$.

Although we cannot derive an explicit expression for $\alpha_c(\varepsilon)$, we can show that it satisfies a self-consistent equation. To this end, we consider the regime $\alpha < \alpha_c(\varepsilon)$ and introduce the same rescaled variables as in the non-linear case. In the limit $\lambda \to 0^+$, the state equations for the linear model reduce to
\begin{equation} \label{eq:app-hinge-perceptron-zero}
\begin{alignedat}{2}
    & m = \frac{\hat{m}}{1 + \hat{V}},
    &\qquad\qquad& \hat{m} = \frac{\alpha}{V} \mathcal{I}_{\hat{m}}(q, \eta), \\[6pt]
    & q = \frac{\hat{m}^2 + \hat{q}}{(1 + \hat{V})^2},
    && \hat{q} = \frac{\alpha}{V^2} \mathcal{I}_{\hat{q}}(q, \eta), \\[6pt]
    & V = \frac{1}{1 + \hat{V}},
    && \hat{V} = \frac{\alpha}{V} \mathcal{I}_{\hat{V}}(q, \eta),
\end{alignedat}
\end{equation}
where $\mathcal I_{\hat m}$, $\mathcal I_{\hat q}$, and $\mathcal I_{\hat V}$ are defined in \eqref{eq: appendix - I hinge integrals lambda zero}. This system can be recast entirely in terms of the overlap order parameters as
\begin{align}
    &m = \alpha \, \mathcal{I}_{\hat m}(q,\eta), \label{eq: appendix - reduced ridgeless m hinge} \\[4pt]
    &q = m^2 + \alpha \, \mathcal{I}_{\hat q}(q,\eta), \\[4pt]
    &V = 1 - \alpha \, \mathcal{I}_{\hat V}(q,\eta). \label{eq: appendix - reduced ridgeless V hinge} 
\end{align}
Now, to determine the interpolation threshold, we study these equation as $\alpha \to  \alpha_c(\varepsilon)^-$. Introducing $\delta \alpha = \alpha_c(\varepsilon) - \alpha$, numerical evidence suggests the asymptotic scalings
\begin{equation} \label{eq:app-perceptron-hinge-scaling-threshold}
    m = \Theta\left( (\delta \alpha)^{-1} \right),
    \qquad
    q = \Theta\left( (\delta \alpha)^{-2} \right),
    \qquad
    V = \Theta( \delta \alpha ).
\end{equation}
At leading order, Eq.~\eqref{eq: appendix - reduced ridgeless m hinge} then yields the following limiting relation for $\eta = m^2 / q$:
\begin{equation} \label{eq:condition_eta_hinge_interpolation}
    \sqrt{\frac{\eta}{1 - \eta}} \underset{\alpha \to \alpha_c(\varepsilon)^-}{\longrightarrow} \frac{\alpha_c(\varepsilon) (1 - \varepsilon)}{\pi}.
\end{equation}
Using this relation in the leading-order expansion of Eq.~\eqref{eq: appendix - reduced ridgeless V hinge}, we finally obtain
\begin{equation} \label{eq: appendix - interpolation hinge}
    1 = \alpha_c(\varepsilon) \left[ \frac{1}{2} -\frac{1-\varepsilon}{\pi} \arctan\!\left( \frac{\alpha_c(\varepsilon)(1-\varepsilon)}{\pi} \right) \right].
\end{equation}
This is precisely the interpolation-threshold equation \eqref{eq:interpolation-hinge-main} for the linear model trained with hinge loss that is stated in the main text.

\begin{figure}
    \centering
    \includegraphics[width=1.0\linewidth]{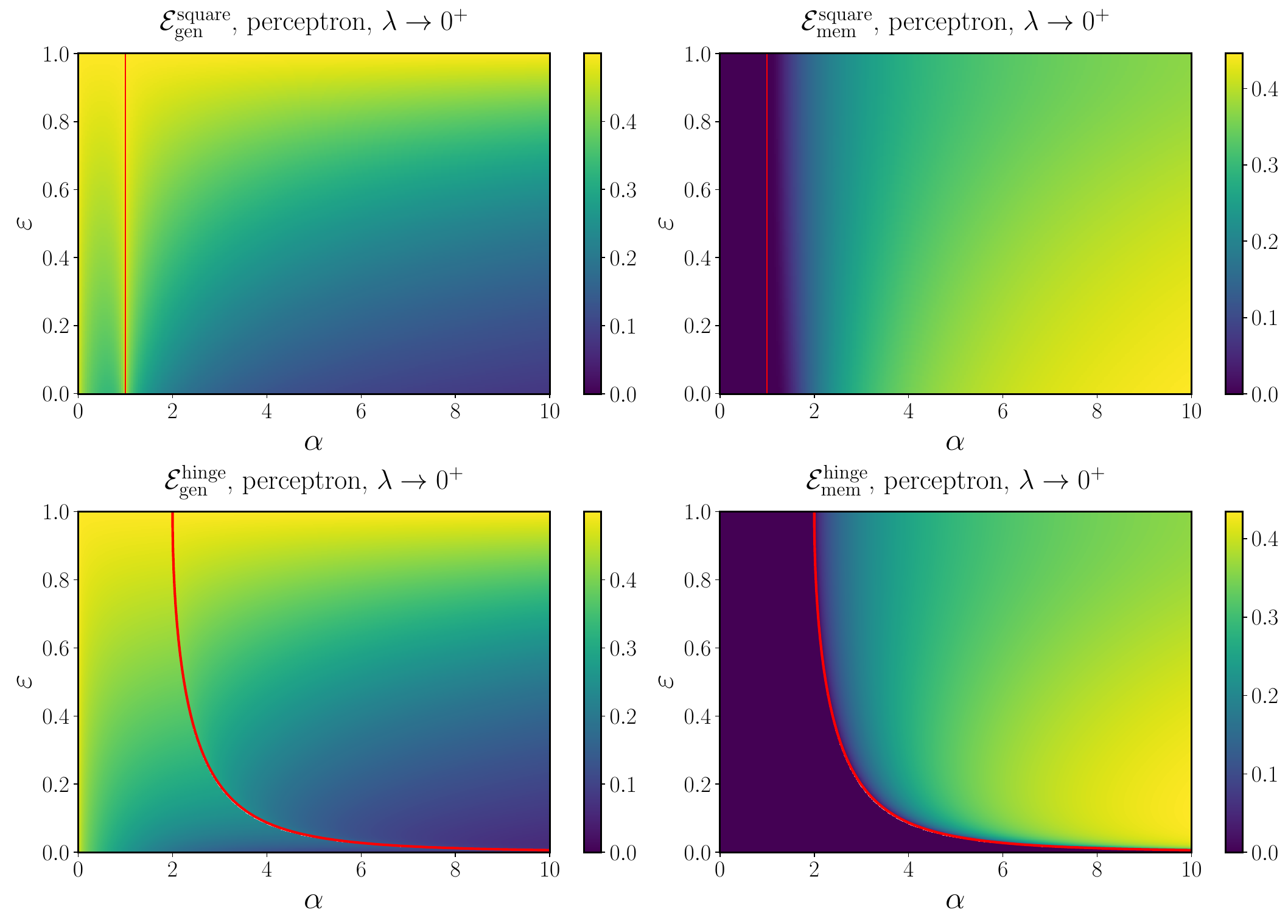}
    \caption{
    Top panels: heatmaps of the generalization error $\cE_{\rm gen}^{\rm square}$ (top left) and memorization error $\cE_{\rm mem}^{\rm square}$ (top right) for the perceptron trained with square loss in the ridgeless limit $\lambda\to 0^+$, as functions of the fraction of facts $\varepsilon$ and the sample complexity $\alpha$. The vertical red line indicates the interpolation threshold $\alpha_c(\varepsilon)=1$. 
    Bottom panels: heatmaps of the generalization error $\cE_{\rm gen}^{\rm hinge}$ (bottom left) and memorization error $\cE_{\rm mem}^{\rm hinge}$ (bottom right) for the perceptron trained with hinge loss in the ridgeless limit $\lambda\to 0^+$, as functions of the fraction of facts $\varepsilon$ and the sample complexity $\alpha$. The red curve is the interpolation threshold $\alpha_c(\varepsilon)$ defined in Eq.~\eqref{eq: appendix - interpolation hinge}. Below this curve, the memorization error is exactly zero, $\cE_{\rm mem}^{\rm hinge}=0$, while above it the perceptron can no longer perfectly memorize the random labels and $\cE_{\rm mem}^{\rm hinge}>0$. The threshold decreases monotonically with $\varepsilon$, interpolating between the known value $\alpha_c(1)=2$ for fully random labels and a diverging threshold as $\varepsilon\to 0^+$. For both losses, we observe an interpolation cusp in the generalization error around the corresponding interpolation threshold. 
    }
    \label{fig:hinge-perceptron-threshold}
\end{figure}

For $\varepsilon = 1$, corresponding to a training set consisting entirely of randomly labelled samples, Eq.~\eqref{eq: appendix - interpolation hinge} gives $\alpha_c(1) = 2$, in agreement with \cite{gaussunivperceptron}. By contrast, its numerical solution indicates that $\alpha_c(\varepsilon)$ diverges as $\varepsilon \to 0$. To determine the corresponding asymptotic behavior, we assume $\alpha_c(\varepsilon) \to+ \infty$ and use the large-argument expansion of the arctangent in Eq.~\eqref{eq: appendix - interpolation hinge}. At leading non-trivial order, this gives
\begin{equation}
    1 \underset{\varepsilon \to 0}{\simeq} \frac{\varepsilon}{2} \alpha_c(\varepsilon) + 1 - \frac{\pi^2}{3 (1 - \varepsilon)^2\,\alpha_c(\varepsilon)^2}.
\end{equation}
After cancelling the constant terms, we find
\begin{equation} \label{eq: appendix - perceptron hinge alpha small eps}
    \alpha_c(\varepsilon) \underset{\varepsilon \to 0}{\simeq} \left(\frac{2\pi^2}{3\varepsilon}\right)^{1/3}.
\end{equation}
Thus, the interpolation threshold diverges as $\varepsilon^{-1/3}$ when the fraction of randomly labelled samples tends to zero, consistently with the initial assumption that $\alpha_c(\varepsilon) \to +\infty$ as $\varepsilon \to 0$.

The bottom panels of Fig.~\ref{fig:hinge-perceptron-threshold} provide numerical support for these predictions. In particular, they show the existence of a threshold $\alpha_c(\varepsilon)$ separating perfect from imperfect memorization in the ridgeless limit, as well as the divergence of $\alpha_c(\varepsilon)$ as $\varepsilon \to 0$. For completeness, the top panels display the generalization and memorization errors attained by the linear model in the ridgeless limit under the square loss, highlighting the interpolation threshold $\alpha_c = 1$ derived previously.

\subsubsection{Large-\texorpdfstring{$\alpha$}{alpha} asymptotics of the SVM generalization error at \texorpdfstring{$\lambda_\mathrm{opt}$}{lambdaopt}}
\label{app:kernel_SVM_large_alpha}
\begin{figure}[h]
    \centering
    \includegraphics[width=0.5\linewidth]{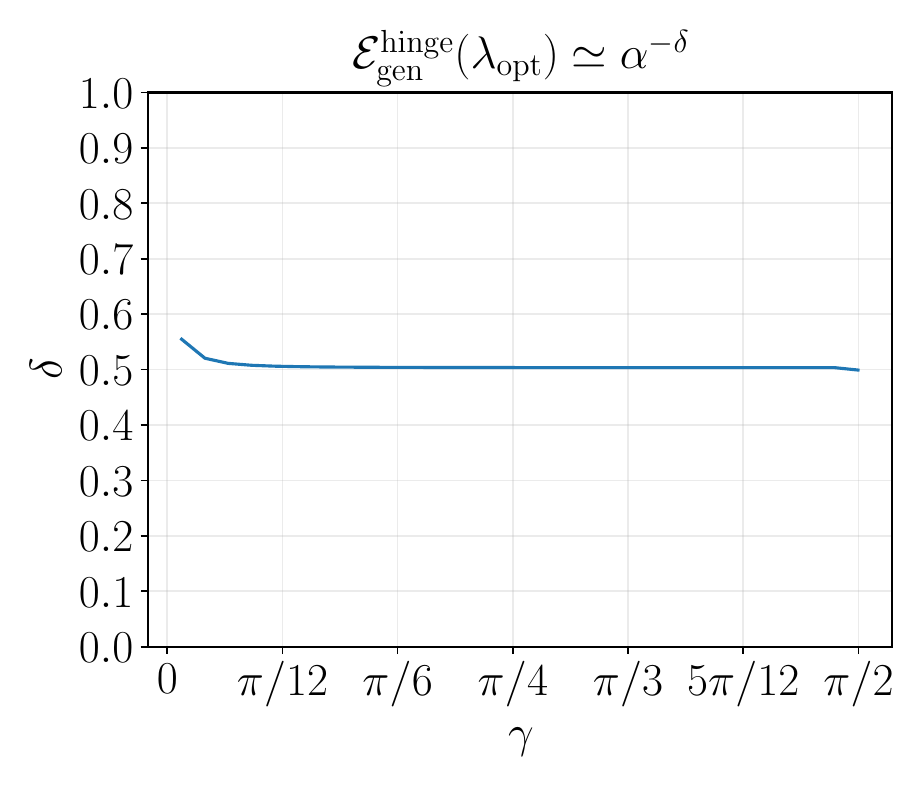}
    \caption{Exponent of the power law decay of the optimal SVM generalization error $\mathcal{E}_\mathrm{gen}^\mathrm{hinge}(\lambda_\mathrm{opt})$ at large $\alpha$ as a function of the angle $\gamma = \arctan(\mu_1 / \mu_\star)$. The fraction of the random facts is fixed to $\varepsilon = 0.5$. The value $\delta$ of the exponent of the power law $\alpha^{-\delta}$ is numerically estimated from a linear fit in log-log scale of the generalization error at the optimal regularization $\mathcal{E}_\mathrm{gen}^\mathrm{hinge}(\lambda_\mathrm{opt})$
    in the range $\alpha \in (10, 10^4)$.}
    \label{fig:SVM slope vs angle large alpha}
\end{figure}

In this subsection, we report the results of the numerical analysis of the large-$\alpha$ behavior of the optimally regularized SVM generalization error $\mathcal{E}_\mathrm{gen}^\mathrm{hinge}$. For each fixed choice of $\alpha$, $\varepsilon$, $\mu_1$ and $\mu_\star$, we solve the state equations \eqref{eq: appendix - SE SVM overlaps}--\eqref{eq: appendix - SE SVM hat variables} for several values of the regularization strength $\lambda$ and evaluate the corresponding generalization error. We then determine $\lambda_\mathrm{opt}$ by numerical cross-validation as the value of $\lambda$ that minimizes $\mathcal{E}_\mathrm{gen}^\mathrm{hinge}$. In Fig.~\ref{fig:SVM slope vs angle large alpha}, for a representative value of $\varepsilon$, we show that the resulting optimally regularized generalization error decays as $\alpha^{-1/2}$ at large $\alpha$.

%% file: Appendices/Appendix_BO-RAF_review.tex
\section{Bayes-optimal generalization error in the RAF model}
\label{app: BO RAF}

This appendix provides the theoretical basis for the results on the Bayes-optimal generalization error in the RAF model presented in Sections~\ref{sec: main equations} and \ref{sec:results}.

Following \cite{aubin2020generalization}, in the high-dimensional limit $n, d \to \infty$ at fixed sample complexity $\alpha = n / d$, for Gaussian inputs and binary labels generated by a generalized linear model with weights $\boldsymbol{w}_\star \sim \mathcal{N}(\boldsymbol{0}, \mathbb{I}_d)$, the Bayes-optimal generalization error is given by
\begin{equation}
\label{eq:app-BO-test-error}
  \cE_{\rm gen}^{\rm BO} = \frac{1}{\pi} \arccos\!\left( \sqrt{q_b} \right),
\end{equation}
where the order parameter $q_b$ denotes the asymptotic overlap between the Bayes-optimal estimator and the teacher vector $\boldsymbol{w}_\star$. In the following, we derive the state equations determining $q_b$ in the RAF model and analyze the large-$\alpha$ behavior of the resulting generalization error.

\subsection{Bayes-optimal state equations}
\label{App:BO-test-error}
For an isotropic Gaussian prior on the teacher weights, the Bayes-optimal overlap satisfies the following system of state equations:
\begin{align}
  q_b &= \frac{\hat{q}_b }{\hat{q}_b +1}, \label{eq: appendix - SE BO q_b}
  \\
  \hat{q}_b &= \alpha \, \mathbb{E}_{\xi \sim \mathcal{N}(0,1)}\!\left[ \int_\mathbb{R} \dd y \, \mathcal{Z}_{{\rm out^\star}}(y,q_b^{1/2}\xi, 1-q_b)\; f_{{\rm out^\star}}(y,q_b^{1/2}\xi, 1-q_b)^2 \right] \label{eq: appendix - SE BO qhat_b}.
\end{align}
Here, $\hat q_b$ is the conjugate variable associated with $q_b$. The functions $\mathcal{Z}_{\mathrm{out}^\star}$ and $f_{\mathrm{out}^\star}$ are defined in Eqs.~\eqref{eq: appendix - output channel partition function} and~\eqref{eq: appendix - Bayes-optimal denoising function}, respectively.

For the RAF data model, the right-hand side of the equation for $\hat{q}_b$ can be evaluated explicitly. Substituting Eqs.~\eqref{app:z_out_denoising RAF} and \eqref{app:f_out_denoising RAF} and carrying out the integration over the labels $y$, we obtain Eq.~\eqref{eq:bo-hatqb} of the main text:
\begin{equation} \label{eq:app-hatqb-BO RAF}
    \hat{q}_b= \frac{4 \alpha (1-\varepsilon)^2}{(2\pi)^{3/2}(1 - q_b)} \int_{\mathbb{R}} \dd \xi \, \frac{\exp\!\left( -\frac{1 + q_b}{2 (1 - q_b)} \, \xi^2 \right)}{ 1 + (1-\varepsilon)\, \mathrm{erf}\!\left( \frac{\sqrt{q_b}\,\xi}{\sqrt{2 (1 - q_b)}} \right)}.
\end{equation}

\begin{figure}
    \centering
    \includegraphics[width=0.6\linewidth]{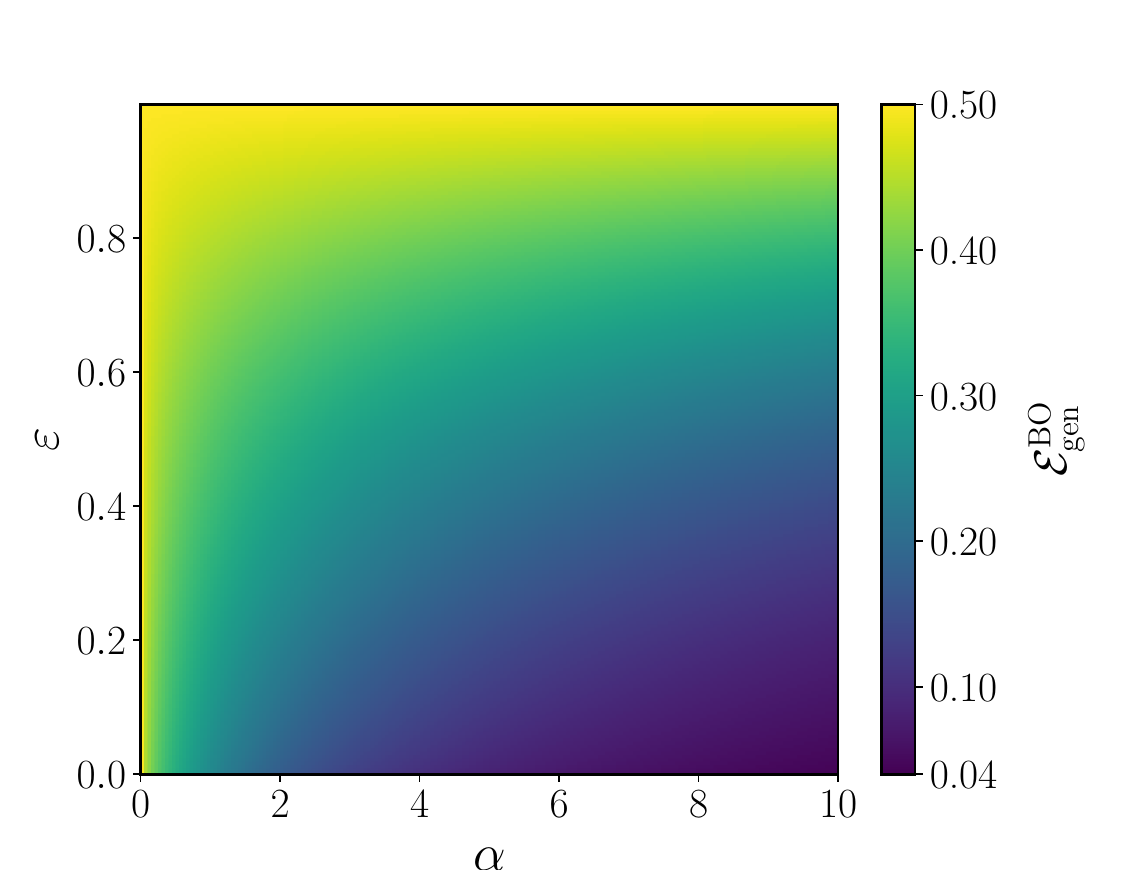}
    \caption{Heatmap of the Bayes-optimal generalization error as a function of the sample complexity $\alpha$ and the fraction of random labels in the training set $\varepsilon$.}
    \label{fig:app-BO-heatmap}
\end{figure}

Fig.~\ref{fig:app-BO-heatmap} shows the Bayes-optimal generalization error across the $(\alpha, \varepsilon)$ plane. At fixed sample complexity $\alpha$, the error increases with the fraction of random labels $\varepsilon$ and approaches the random-guessing value $1/2$ as $\varepsilon \to 1$.

\subsubsection{Large-\texorpdfstring{$\alpha$}{alpha} expansion of the Bayes-optimal generalization error}
\label{app:decay_BO}

For fixed $0 \leq \varepsilon < 1$, we seek the large-$\alpha$ solution of the state equations characterized by $q_b \to 1$. Since $q_b$ quantifies the asymptotic alignment between the Bayes-optimal estimator and the teacher vector, this corresponds to asymptotically perfect recovery of the teacher direction.

Introducing the change of variable $t = \xi / \sqrt{1 - q_b}$, the equation for $\hat{q}_b$ becomes
\begin{equation}
    \hat{q}_b = \frac{4 \alpha (1 - \varepsilon)^2}{(2 \pi)^{3/2} \sqrt{1 - q_b}} \int_{\mathbb{R}} \dd t \, \frac{\exp\!\left( - \frac{1 + q_b}{2} \, t^2 \right)}{ 1 + (1 - \varepsilon)\, \mathrm{erf}\!\left( \frac{\sqrt{q_b} \, t}{\sqrt{2}} \right)}.
\end{equation}
Taking the limit $q_b \to 1$ in the integrand yields
\begin{equation}
    \hat{q}_b \underset{\alpha \to \infty}{\simeq} \frac{4 \alpha (1 - \varepsilon)^2}{(2 \pi)^{3/2} \sqrt{1 - q_b}} \int_{\mathbb{R}} \dd t \, \frac{e^{- t^2}}{ 1 + (1 - \varepsilon)\, \mathrm{erf}\!\left( \frac{t}{\sqrt{2}} \right)} = J(\varepsilon) \, \frac{\alpha}{\sqrt{1 - q_b}},
\end{equation}
where
\begin{equation}
    J(\varepsilon) = \frac{4 (1 - \varepsilon)^2}{(2 \pi)^{3/2}} \int_{\mathbb{R}} \dd t \, \frac{e^{- t^2}}{ 1 + (1 - \varepsilon)\, \mathrm{erf}\!\left( \frac{t}{\sqrt{2}} \right)}.
\end{equation}
The large-$\alpha$ state equations are therefore
\begin{align}
  q_b &= \frac{\hat{q}_b }{\hat{q}_b +1},
  \\
  \hat{q}_b &\underset{\alpha \to \infty}{\simeq} J(\varepsilon) \, \frac{\alpha}{\sqrt{1 - q_b}}.
\end{align}
Solving the system at leading order gives
\begin{equation}
    q_b \underset{\alpha \to \infty}{\simeq} 1 - \frac{1}{\alpha^2 J(\varepsilon)^2}.
\end{equation}
Therefore, the Bayes-optimal generalization error at large-$\alpha$ is given by
\begin{equation}
    \mathcal{E}_\mathrm{gen}^\mathrm{BO}(\alpha, \varepsilon) = \frac{1}{\pi} \arccos\!\left( \sqrt{q_b} \right) \underset{\alpha \to \infty}{\simeq} \frac{1}{\pi J(\varepsilon) \alpha} =  \frac{C_\mathrm{BO}(\varepsilon)}{\alpha},
\end{equation}
where
\begin{equation}
    C_\mathrm{BO}(\varepsilon) = \frac{1}{\pi J(\varepsilon)}.
\end{equation}

We thus conclude that, for every fixed $0 \leq \varepsilon < 1$, the Bayes-optimal generalization error decays as $\alpha^{-1}$.

As a consistency check, for $\varepsilon = 0$ one recovers the pure teacher--student setting. In this case,
\begin{equation}
    J(0) = \frac{1}{\pi}\sqrt{\frac{2}{\pi}} \int_{\mathbb{R}} \dd t \, \frac{e^{- t^2}}{ 1 + \mathrm{erf}\!\left( \frac{t}{\sqrt{2}} \right)},
\end{equation}
and hence
\begin{equation}
    C_\mathrm{BO}(0) = \frac{1}{\sqrt{\frac{2}{\pi}} \int_{\mathbb{R}} \dd t \, \frac{e^{- t^2}}{ 1 + \mathrm{erf}\!\left( \frac{t}{\sqrt{2}} \right)}} \approx 0.4417,
\end{equation}
in agreement with the large-$\alpha$ Bayes-optimal asymptotics reported in~\cite{aubin2020generalization}.

%% file: Appendices/Appendix_Numerics_review.tex
\section{Numerical details and further experiments}
\label{App:E}

In this appendix we collect numerical details, representative kernel families, and additional experiments complementing the results of Secs.~\ref{sec:results} and \ref{sec:real_data_representations}. First, we provide examples of kernel families for which the asymptotic theory is valid, together with the corresponding coefficients $(\mu_0,\mu_1,\mu_\star)$. We also report additional numerical experiments, both on the synthetic RAF model and on CIFAR10-RAF, in order to clarify which aspects of the theory are specific to the Gaussian high-dimensional setting. Throughout this appendix, kernel ridge regression refers to the square-loss estimator of Sec.~\ref{sec:ERM_equations}, and support vector machine refers to the hinge-loss estimator. For the Gaussian-input RAF model, all theoretical curves are obtained by numerically solving the asymptotic state equations stated in Sec.~\ref{sec:ERM_equations} and Appendices \ref{app: KRR RAF} and \ref{app: SVM RAF}. 

\subsection{Kernel families}
\label{appD:kernels}

We begin by listing in Table~\ref{tab:mu-moments-kernels} some representative kernel families captured by our theory, together with the coefficients $(\mu_0,\mu_1,\mu_\star)$ defined in Eq.~\eqref{eq: appendix - mus from kernel}. 
As shown in Sec.~\ref{sec:ERM_equations} and Appendix~\ref{app: eqs_kernel_limit}, for the Gaussian RAF model the asymptotic performance of kernel regression depends on the kernel only through $\mu_1$ and $\mu_\star$, while the constant $\mu_0^2=K(0)$ does not affect the errors.

\begin{figure}[t]
    \centering
    \includegraphics[width=0.6\linewidth]{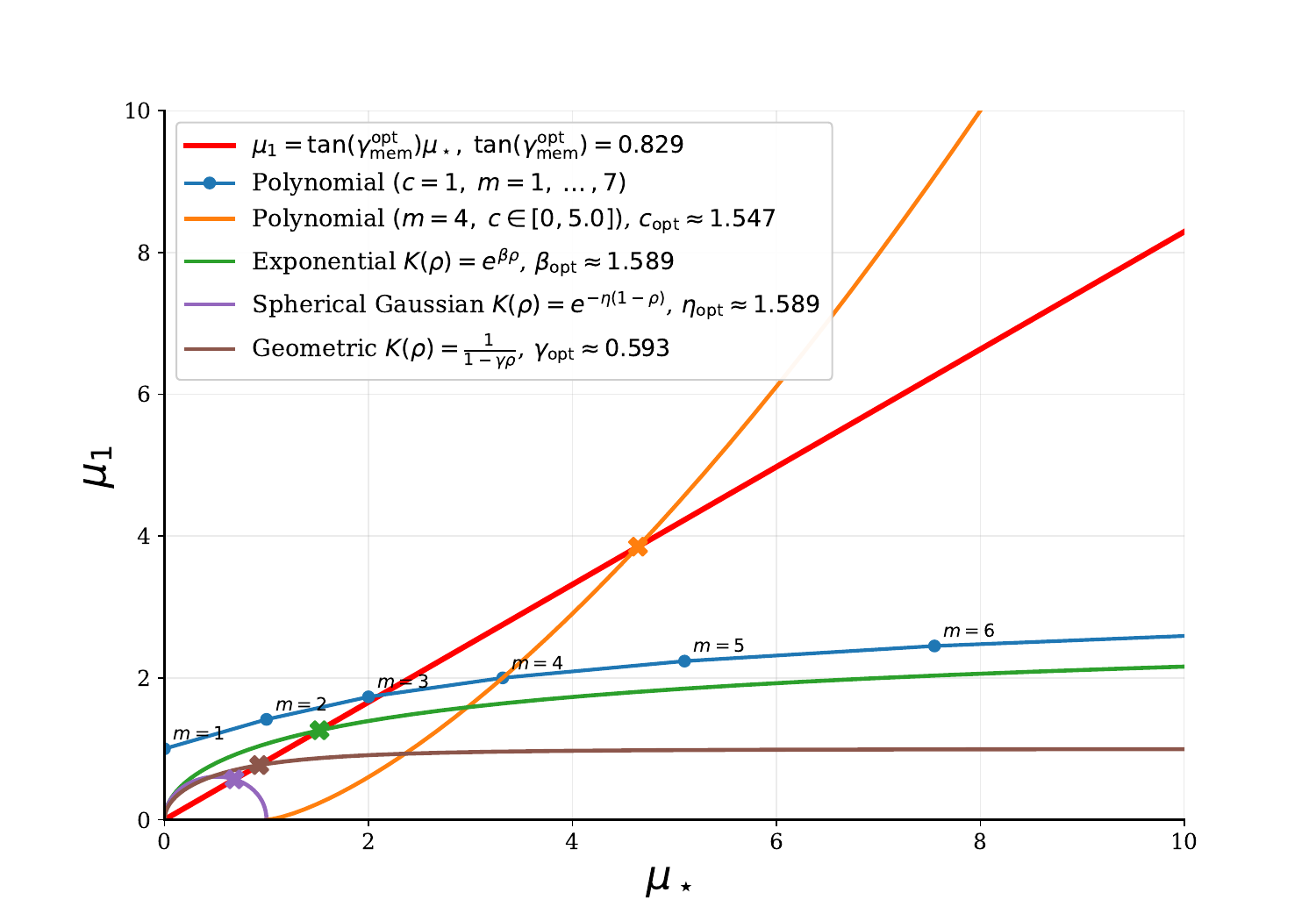}
    \caption{
    Representative kernel families in the $(\mu_\star,\mu_1)$ plane. The red line is the optimal square-loss direction
    $\mu_1=\tan\bigl(\gamma_{\rm mem}^{\rm opt}\bigr)\mu_\star$, with $\gamma_{\rm mem}^{\rm opt}$  for $\varepsilon=0.2$ given by Eq.~\eqref{eq:angle_opt_mem}. 
    Each curve is obtained by varying the internal parameter of a kernel family while plotting the induced pair $(\mu_\star,\mu_1)$. 
    % For polynomial kernels of fixed degree $m$, the curve is parameterized by $c$; for the exponential, spherical Gaussian, and geometric families, it is parameterized respectively by $\beta$, $\eta$, and $\gamma$. 
    Markers indicate the parameter values for which the family intersects the optimal line. This figure makes explicit that distinct kernel families can realize the same asymptotic geometry $(\mu_1,\mu_\star)$, and are therefore predicted by the Gaussian RAF theory to yield the same asymptotic performance.
    }
    \label{fig:AppE-Kernels}
\end{figure}

\paragraph{Conditions for matching the optimal angle.}
For the square loss, Sec.~\ref{subsec:choice_nonlinearity} and Appendix~\ref{app: KRR RAF} show that the optimal interpolating kernels for the generalization error are those determined by the ratio
\begin{equation}
\frac{\mu_1}{\mu_\star}
=
\tan \bigl(\gamma_{\rm mem}^{\rm opt}(\varepsilon)\bigr)
=
\left(\frac{\pi}{2(1-\varepsilon)^2}-1\right)^{-1/2},
\label{eq:appE-optimal-ratio}
\end{equation}
where $\gamma_{\rm mem}^{\rm opt}$ is given in Eq.~\eqref{eq:angle_opt_mem}. 
For a parameterized kernel family, the value of the internal kernel parameters that realizes the optimal angle is therefore determined by imposing Eq.~\eqref{eq:appE-optimal-ratio}. 
Explicitly:
\begin{itemize}
\item \textbf{Polynomial kernel} $K(\rho)=(c+\rho)^m$:
\begin{equation}
\frac{m\,c^{m-1}}{(c+1)^m-c^m-mc^{m-1}}
=
\tan^2\!\bigl(\gamma_{\rm mem}^{\rm opt}(\varepsilon)\bigr).
\label{eq:appE-poly-condition}
\end{equation}
For fixed degree $m$, this is a scalar equation for $c\ge 0$.
\item \textbf{Exponential dot-product kernel} $K(\rho)=e^{\beta\rho}$:
\begin{equation}
\frac{\beta}{e^\beta-1-\beta}
=
\tan^2\!\bigl(\gamma_{\rm mem}^{\rm opt}(\varepsilon)\bigr).
\label{eq:appE-exp-condition}
\end{equation}
\item \textbf{Spherical Gaussian kernel} $K(\rho)=e^{-\eta(1-\rho)}$:
\begin{equation}
\frac{\eta e^{-\eta}}{1-e^{-\eta}(1+\eta)}
=
\tan^2\!\bigl(\gamma_{\rm mem}^{\rm opt}(\varepsilon)\bigr).
\label{eq:appE-gauss-condition}
\end{equation}
\item \textbf{Geometric kernel} $K(\rho)=\frac{1}{1-\gamma\rho}$:
\begin{equation}
\frac{\mu_1}{\mu_\star}
=
\frac{\sqrt{1-\gamma}}{\sqrt{\gamma}}
=
\tan\!\bigl(\gamma_{\rm mem}^{\rm opt}(\varepsilon)\bigr),
\end{equation}
hence the solution is explicit:
\begin{equation}
\gamma_{\rm geo}^{\rm opt}(\varepsilon)
=
\frac{1}{1+\tan^2\!\bigl(\gamma_{\rm mem}^{\rm opt}(\varepsilon)\bigr)}
=
\cos^2\!\bigl(\gamma_{\rm mem}^{\rm opt}(\varepsilon)\bigr).
\label{eq:appE-geo-solution}
\end{equation}
\end{itemize}
A graphical representation of these results is provided in Fig.~\ref{fig:AppE-Kernels}.
%For the sign, erf, and ReLU kernels, the point $(\mu_\star,\mu_1)$ is fixed, so one can only compare their intrinsic angle to the optimal one.

\begin{table}[!ht]
\centering
\small
\renewcommand{\arraystretch}{1.3}
\setlength{\tabcolsep}{6pt}

\begin{tabular}{
>{\raggedright\arraybackslash}p{3.7cm}
>{\raggedright\arraybackslash}p{5.5cm}
>{\raggedright\arraybackslash}p{5.cm}
}
\hline
\textbf{Kernel / activation} & \textbf{$K(\rho)$, $\sigma(u)$} & \textbf{$(\mu_0,\mu_1,\mu_\star)$} \\
\hline

\textbf{Linear (perceptron)}
&
$K(\rho)=\rho$

$\sigma(u)=u$
&
$\begin{aligned}
\mu_0 &= 0,\\
\mu_1 &= 1,\\
\mu_\star &= 0
\end{aligned}$
\\

\hline

\textbf{Arc-sine (sign)}
&
$\displaystyle K_{\mathrm{sign}}(\rho)=\frac{2}{\pi}\arcsin(\rho)$

$\sigma(u)=\mathrm{sign}(u)$
&
$\begin{aligned}
\mu_0 &= 0,\\
\mu_1 &= \sqrt{\frac{2}{\pi}},\\
\mu_\star &= \sqrt{1-\frac{2}{\pi}}
\end{aligned}$
\\

\hline

\textbf{Arc-sine (erf)}
&
$\displaystyle K_{\mathrm{erf}}(\rho)=\frac{2}{\pi}\arcsin\!\Bigl(\frac{2\rho}{3}\Bigr)$

$\sigma(u)=\mathrm{erf}(u)$
&
$\begin{aligned}
\mu_0 &= 0,\\
\mu_1 &= \frac{2}{\sqrt{3\pi}},\\
\mu_\star &= \sqrt{\frac{2}{\pi}\arcsin\!\Bigl(\frac{2}{3}\Bigr)-\frac{4}{3\pi}}
\end{aligned}$
\\

\hline

\textbf{ReLU (arc-cosine)}
&
$\displaystyle
K_{\mathrm{ReLU}}(\rho)=\frac{1}{2\pi}\Bigl(\sqrt{1-\rho^2}+(\pi-\arccos\rho)\rho\Bigr)$

$\sigma(u)=\mathrm{ReLU}(u)$
&
$\begin{aligned}
\mu_0 &= \frac{1}{\sqrt{2\pi}},\\
\mu_1 &= \frac{1}{2},\\
\mu_\star &= \sqrt{\frac{1}{2}\Bigl(\frac{1}{2}-\frac{1}{\pi}\Bigr)}
\end{aligned}$
\\

\hline

\textbf{Polynomial dot-product}
&
$K_{\mathrm{poly}}(\rho)=(c+\rho)^m$

$m\in\mathbb{N},\ c\ge 0$
&
$\begin{aligned}
\mu_0 &= c^{m/2},\\
\mu_1 &= \sqrt{m\,c^{m-1}},\\
\mu_\star &= \sqrt{(c+1)^m-c^m-mc^{m-1}}
\end{aligned}$
\\

\hline

\textbf{Exponential dot-product}
&
$K_{\exp}(\rho)=e^{\beta\rho}, \qquad \beta\ge 0$

$\sigma_\beta(u)=\exp(\sqrt{\beta}\,u-\beta/2)$
&
$\begin{aligned}
\mu_0 &= 1,\\
\mu_1 &= \sqrt{\beta},\\
\mu_\star &= \sqrt{e^\beta-1-\beta}
\end{aligned}$
\\

\hline

\textbf{Spherical Gaussian (RBF, high-$d$ scaling)}
&
$K_{\mathrm{Gauss}}(\rho)=e^{-\eta(1-\rho)}, \qquad \eta\ge 0$

$\sigma_{\mathrm{Gauss}}(u)=e^{-\eta/2}\exp(\sqrt{\eta}\,u-\eta/2)$
&
$\begin{aligned}
\mu_0 &= e^{-\eta/2},\\
\mu_1 &= \sqrt{\eta e^{-\eta}},\\
\mu_\star &= \sqrt{1-e^{-\eta}(1+\eta)}
\end{aligned}$
\\

\hline

\textbf{Geometric dot-product}
&
$K_{\mathrm{geo}}(\rho)=\frac{1}{1-\gamma\rho}, \qquad 0\le\gamma<1$

$\sigma_{\mathrm{geo}}(u)=\sum_{\ell\ge0}\sqrt{\frac{\gamma^\ell}{\ell!}}H_\ell(u)$
&
$\begin{aligned}
\mu_0 &= 1,\\
\mu_1 &= \sqrt{\gamma},\\
\mu_\star &= \sqrt{\frac{\gamma^2}{1-\gamma}}
          = \frac{\gamma}{\sqrt{1-\gamma}}
\end{aligned}$
\\

\hline
\end{tabular}
\caption{
Representative kernels and the corresponding coefficients $(\mu_0,\mu_1,\mu_\star)$. 
For the RAF model with Gaussian inputs, the asymptotic errors depend on the kernel only through $(\mu_1,\mu_\star)$, as established in Sec.~\ref{sec:ERM_equations} and Appendix~\ref{app: eqs_kernel_limit}. 
}
\label{tab:mu-moments-kernels}
\end{table}

\begin{figure}[!b]
    \centering
    \includegraphics[width=1\linewidth]{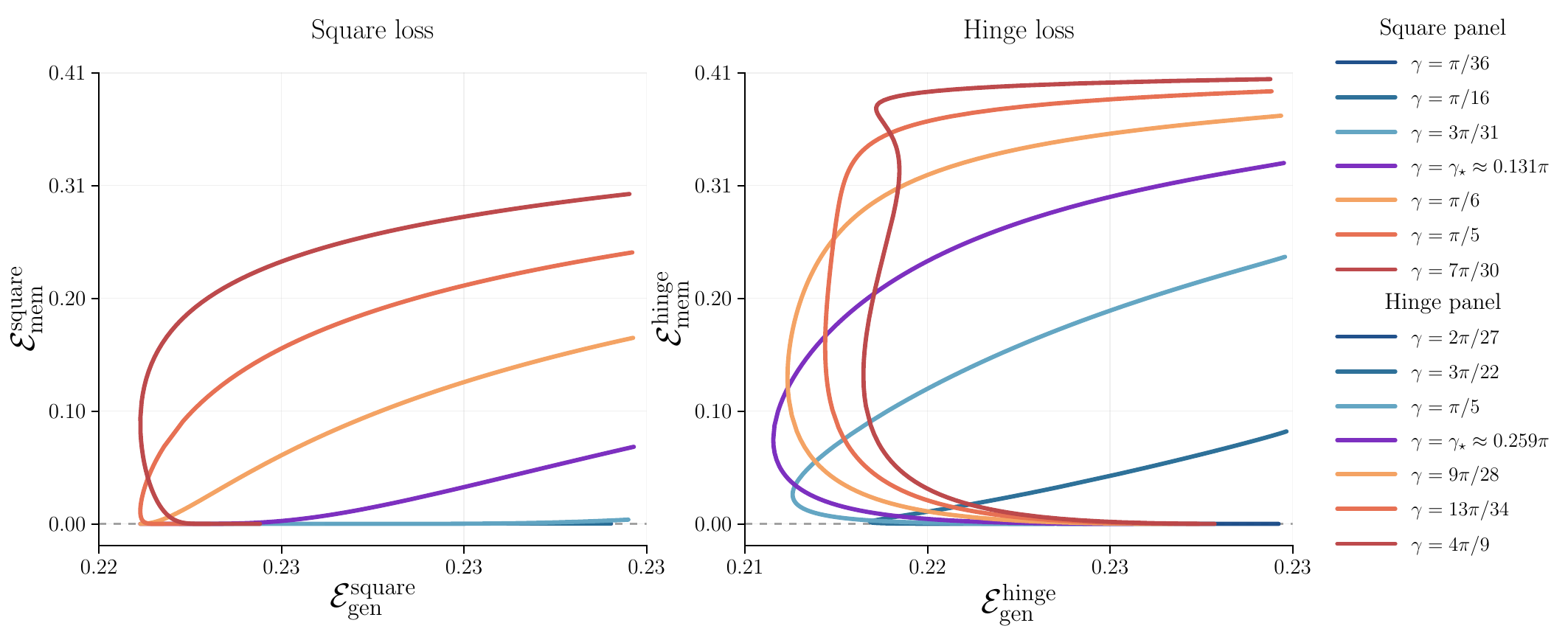}
    \caption{
    Representative trade-off curves $\lambda\mapsto(\cE_{\rm gen}(\lambda),\cE_{\rm mem}(\lambda))$ for a broad range of angles $\gamma=\arctan(\mu_1/\mu_\star)$,
    %at fixed radius $\mu_1^2+\mu_\star^2=9.0$, 
    for $\varepsilon=0.5$ and $\alpha(1-\varepsilon)=4.0$. 
    Left: square loss. Right: hinge loss. 
    The distinguished angles highlighted in the legend in purple correspond to the theoretical optima discussed in Sec.~\ref{subsec:choice_nonlinearity}. For the square loss, the curve with $\gamma=\gamma_{\rm mem}^{\rm opt}$ is the unique one that simultaneously achieves perfect memorization and optimal generalization. For the hinge loss, the family of curves illustrates the trade-off between the two objectives.
    }
    \label{fig:spanning}
\end{figure}

\subsection{Spanning the angle \texorpdfstring{$\gamma$}{gamma}: full phenomenology and additional experiments}
\label{appE:spanning}

In Sec.~\ref{subsec:choice_nonlinearity} we focused on the distinguished angles that optimize either the interpolating regime or the optimal generalization regime. It is also useful to visualize the full phenomenology by spanning a broad range of angles
\begin{equation}
\gamma=\arctan\!\left(\frac{\mu_1}{\mu_\star}\right)\, .
\end{equation}
%at fixed radius $\mu_1^2+\mu_\star^2$. 
Because of the angular invariance established in Appendix~\ref{app: kernel symmetries}, the relevant kernel dependence is entirely controlled by $\gamma$ when either $\lambda\to 0^+$ or $\lambda=\lambda_{\rm opt}$.

Figure~\ref{fig:spanning} shows representative memorization--generalization trade-off curves for a wide range of angles, both for square loss and hinge loss. 
For the square loss, the figure illustrates the sharp transition described analytically in Appendix~\ref{app: reg limits and opt lambda KRR}: the angle $\gamma_{\rm mem}^{\rm opt}$ separates kernels whose optimal regularization, with respect to the generalization error, is strictly positive from kernels whose optimal regularization is already interpolating. 
For the hinge loss, the same phenomenology survives qualitatively, but the optimal angle for optimal generalization error and the angle minimizing the generalization error at zero regularization no longer coincide exactly, in agreement with the discussion in Sec.~\ref{subsec:choice_nonlinearity}.

\begin{figure}[t]
    \centering
    \includegraphics[width=0.98\linewidth]{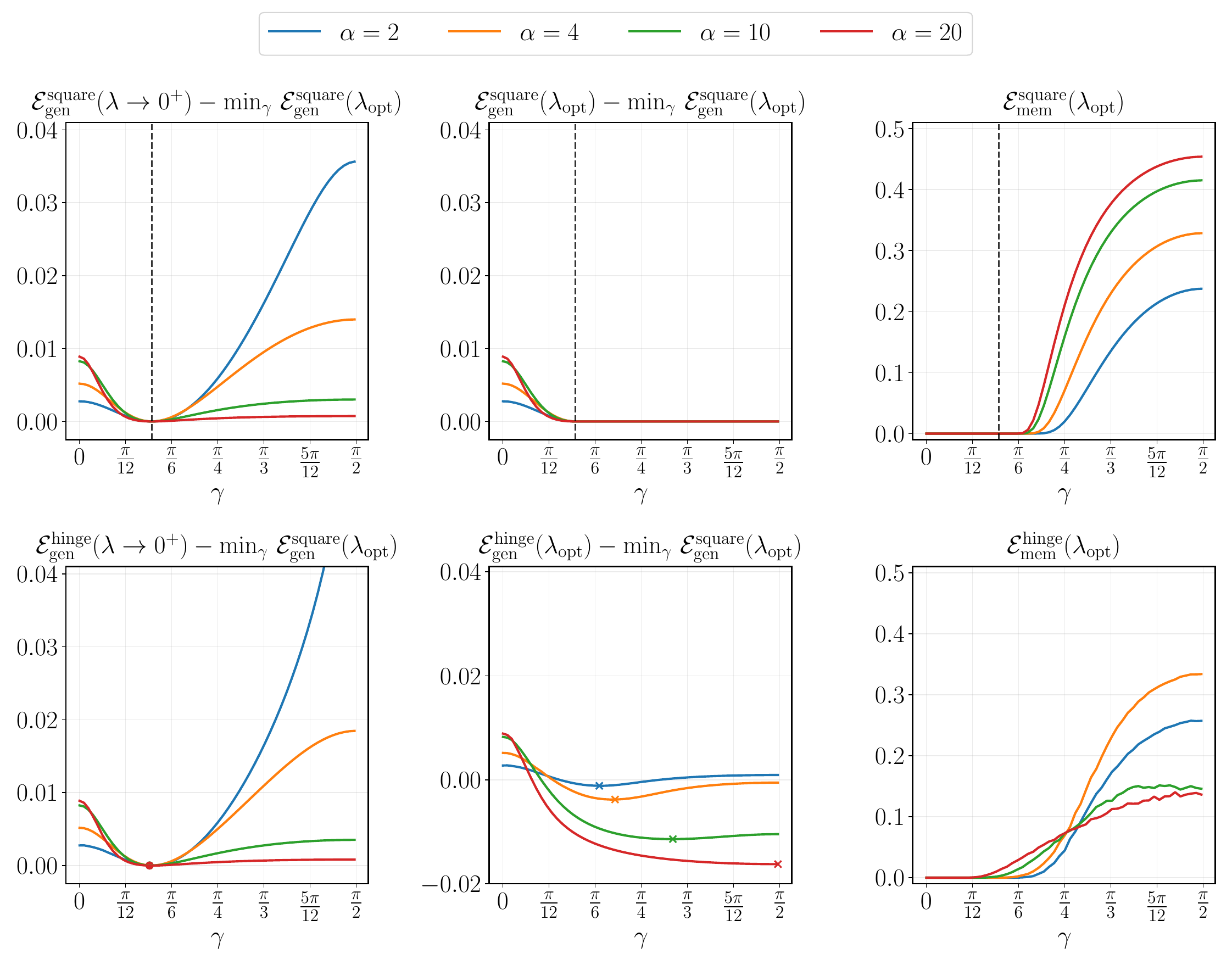}
     \caption{
  \textbf{Generalization and memorization dependence on the angle $\gamma = \arctan\!\left( \mu_1 / \mu_\star \right)$.} KRR (square loss) in the top panels, SVM (hinge loss) in the bottom panels. 
  Left panels: $\cE_{\rm gen}(\lambda \to 0^+) - \min_\gamma \cE_{\rm gen}^{\rm square}(\lambda_\mathrm{opt})$; center panels: $\cE_{\rm gen}(\lambda_{\rm opt}) - \min_\gamma \cE_{\rm gen}^{\rm square}(\lambda_\mathrm{opt})$; right panels: $\cE_{\rm mem}(\lambda_{\rm opt})$. The fraction of facts in the training set is fixed to  $\varepsilon = 0.5$. The sample complexities are  $\alpha \in \{2, 4, 10, 20\}$ in each panel.
  The dashed black vertical line in the upper panels is the optimal angle $\gamma^{\rm opt}_{\rm mem}(\varepsilon)$ in Eq.~\eqref{eq:angle_opt_mem}, for which, for the square loss, both minimum generalization and perfect memorization are simultaneously reached. 
  For the hinge loss (bottom panels), the angle where $\mathcal{E}_\mathrm{gen}^\mathrm{hinge}(\lambda_\mathrm{opt})$ is minimal is marked by a cross, while the angle where $\mathcal{E}_\mathrm{gen}^\mathrm{hinge}(\lambda\to0^+)$ is minimum is marked by a dot. 
  }
  \label{fig: Figure 3 App}
\end{figure}

For completeness, in Fig.~\ref{fig: Figure 3 App} we report analogous curves to the ones shown in Fig.~\ref{fig: Figure 3} of the main text, yet for a different value of the fraction of facts, that is $\eps=0.5$. This figure shows that the phenomenology discussed in the main text does not change in a significant way when changing the fraction of random labels.

\subsection{Numerical agreement for RAF data}
\label{appE:gaussian-numerics}

We now compare the asymptotic theory to direct finite-size kernel experiments on finite size RAF data. 
The goal is to validate two distinct predictions of the theory:
\begin{enumerate}
\item for a fixed kernel family, the parametric trade-off curve $\lambda\mapsto(\cE_{\rm gen}(\lambda),\cE_{\rm mem}(\lambda))$ is accurately captured by the asymptotic equations;
\item two different kernels sharing the same coefficients $(\mu_1,\mu_\star)$ yield the same asymptotic performance and, already at finite size, numerically matching trade-off curves.
\end{enumerate}

\begin{figure}[t]
    \centering
    \includegraphics[width=1.0\linewidth]{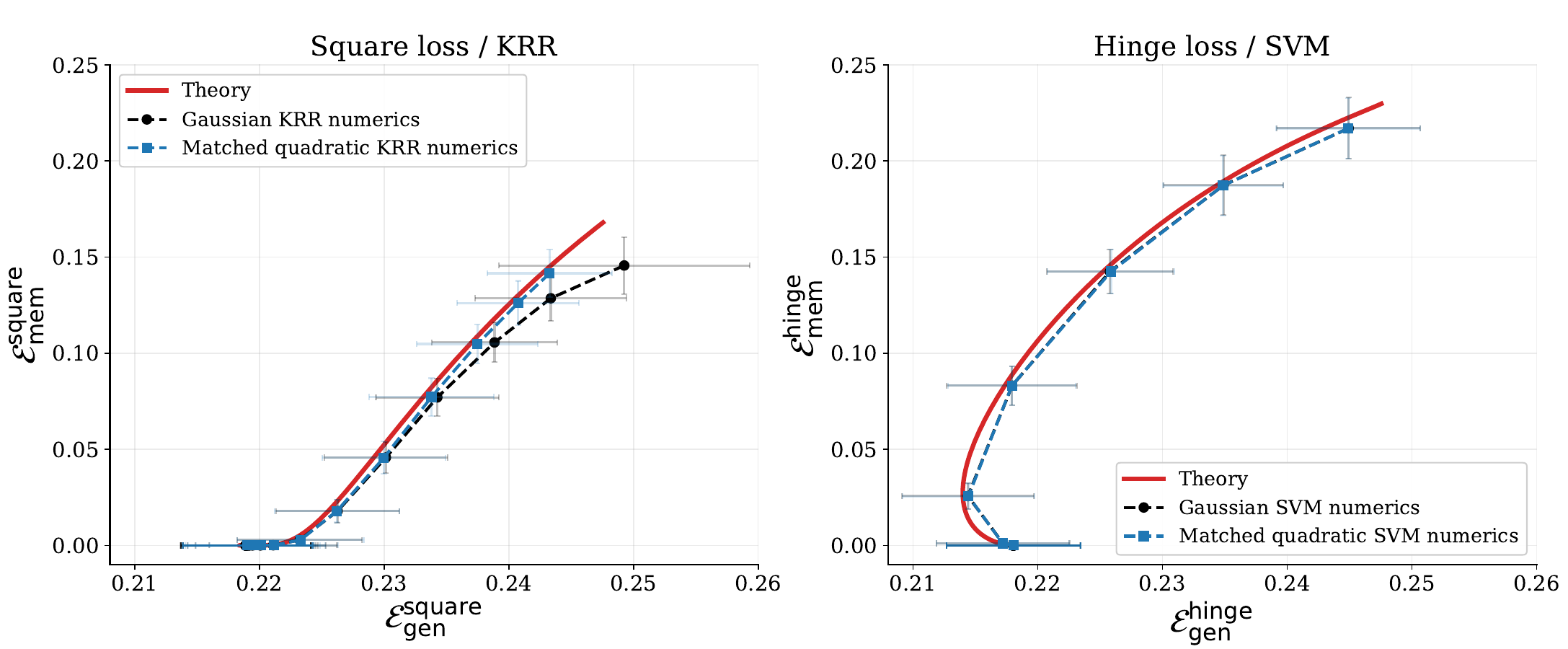}
    \caption{
    Comparison between asymptotic theory for RAF data and direct kernel experiments for the spherical Gaussian kernel $K(\rho)=e^{-\eta(1-\rho)}$, for their respective optimal bandwidths $\eta_{\rm opt}$. In both theory and numerics $\alpha=2.0$, $\eps=0.1$. For the numerics we use $d=3000$ and the error bars are computed over 60 different realizations.
    Left panel: square loss / KRR ($\eta_{\rm opt}^{\rm square}=1.205$). Right panel: hinge loss / SVM ($\eta_{\rm opt}^{\rm hinge}=0.752$). 
    In both cases, the theoretical parametric curves obtained from the state equations closely match the empirical trade-off curves obtained by training the corresponding finite-size kernel estimator. 
    We also compare two distinct kernels with identical coefficients $(\mu_1,\mu_\star)$. 
    The asymptotic theory predicts identical trade-off curves, and the direct kernel experiments confirm this prediction up to finite size corrections.
    }
    \label{fig:appE-gauss-theory-vs-exp}
\end{figure}

\paragraph{Gaussian kernel across bandwidths.}
A convenient family for this comparison is the spherical Gaussian kernel
\begin{equation} \label{eq:gaussian_RBF}
K_{\rm Gauss}(\rho)=e^{-\eta(1-\rho)},
\end{equation}
whose geometry is controlled by
\begin{equation}
\label{eq:mu1_mustar_RBF}
\mu_1=\sqrt{\eta e^{-\eta}},
\qquad
\mu_\star=\sqrt{1-e^{-\eta}(1+\eta)}.
\end{equation}
In the experiments in Fig.~\ref{fig:appE-gauss-theory-vs-exp}, we compare the two representative bandwidths $\eta_{\rm opt}$ for both kernel ridge regression (given by the condition in Eq.~\eqref{eq:appE-gauss-condition}) and support vector machines (found numerically). For the two optimal $\eta_{\rm opt}$, we plot the square-loss and hinge-loss asymptotic performances spanning over the regularization and compare them to the corresponding finite-$n$ KRR and SVM runs. In both cases, we observe a good numerical agreement within finite size corrections of the numerical simulations.

\paragraph{Matching kernels with identical \texorpdfstring{$(\mu_1,\mu_\star)$}{(mu1,mustar)}.}
Another check of the theory is to compare two different kernels that induce the same pair $(\mu_1,\mu_\star)$. 
For the Gaussian RAF model, the asymptotic equations predict that such kernels must yield exactly the same trade-off curves, because the theory depends on the kernel only through these two coefficients.
A simple example consists in matching a spherical Gaussian kernel to a truncated quadratic kernel of the form
\begin{equation}
\label{eq:K_quad}
K_{\rm quad}(\rho)=\mu_1^2\rho+\mu_\star^2\rho^2,
\end{equation}
with $(\mu_1,\mu_\star)$ as in Eq.~\eqref{eq:mu1_mustar_RBF}. The corresponding finite-$n$ numerics is again in Fig.~\ref{fig:appE-gauss-theory-vs-exp} for both KRR and SVM, showing good numerical agreement within their finite-size effects.

\subsection{Matched kernels on CIFAR10-RAF and the hinge loss}
\label{appE:cifar-matching}

In Sec.~\ref{sec:real_data_representations} we compared the RAF-theory phenomenology to the more realistic CIFAR10-RAF construction. A natural stress test is to take two kernels that match in $(\mu_1,\mu_\star)$ and apply them to the same non-Gaussian, structured dataset. 

\begin{figure}[t]
    \centering
    \includegraphics[width=0.95\linewidth]{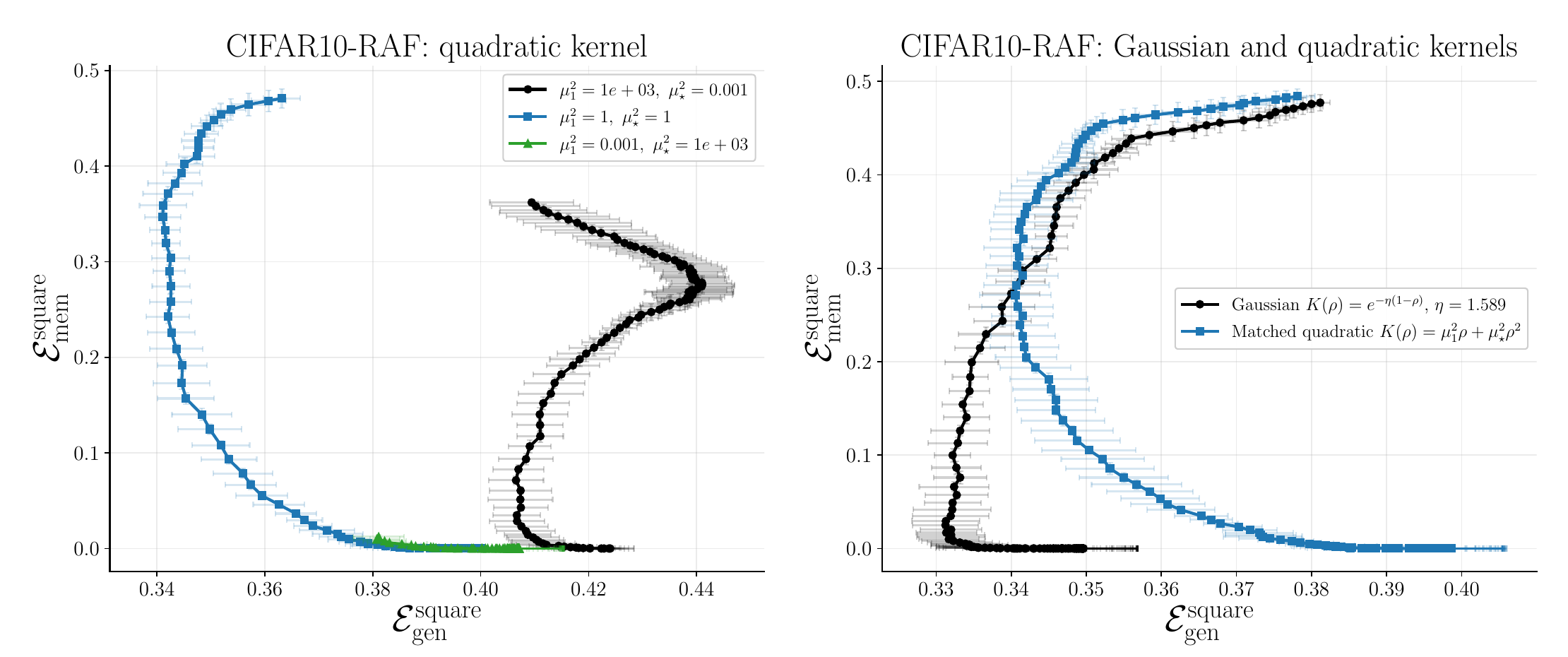}
    \caption{
    CIFAR10-RAF: we compute the memorization and generalization error for KRR with $\alpha=4.0$ and $\eps=0.2$ for both a spherical Gaussian RBF kernel as in Eq.~\eqref{eq:gaussian_RBF}, and a polynomial kernel as in Eq.~\eqref{eq:K_quad} with coefficients matching the Gaussian case. In all cases we average the data over 10 different realizations and the regularization spans over the interval $[10^{-5},10^2]$. 
    Left panel: we show the spans for three representative couples of the parameters $\mu_1$ and $\mu_\star$ in the case of the quadratic kernel. Right panel: comparison between the spherical Gaussian RBF kernel and a quadratic kernel with matching coefficients $\mu_1$, $\mu_\star$ and $\eta=1.589$. 
    %Despite the asymptotic theory matches in the two cases, the CIFAR10-RAF dataset does not yield empirical trade-off curves analogous to those presented in Fig.~\ref{fig:RAF-CIFAR}. 
    This contrasts with the Gaussian RAF prediction and shows that, on structured non-Gaussian data, the pair $(\mu_1,\mu_\star)$ is no longer a complete descriptor of kernel performance. 
    }
    \label{fig:appE-cifar-matching}
\end{figure}

We construct again two kernels, i.e. a spherical Gaussian kernel and a polynomial dot-product kernel, whose induced coefficients $(\mu_1,\mu_\star)$ coincide, as in the previous section. We report the memorization and generalization errors obtained in these two cases in the right panel of Fig.~\ref{fig:appE-cifar-matching}.
When these kernels are trained on the same CIFAR10-RAF dataset, the corresponding KRR memorization-generalization curves are no longer identical, contrary to the theoretical prediction. 
Thus, on structured data, matching $(\mu_1,\mu_\star)$ does not result in equal performances. For completeness, in the left panel of Fig.~\ref{fig:appE-cifar-matching} we report the curves of the errors obtained when spanning over the regularization for the quadratic kernel in Eq.~\eqref{eq:K_quad} when freely choosing the parameters $\mu_1$ and $\mu_\star$, showing a qualitatively similar, yet different, phenomenology to its Gaussian counterpart. 

\begin{figure}[!th]
    \centering
    \includegraphics[width=0.9\linewidth]{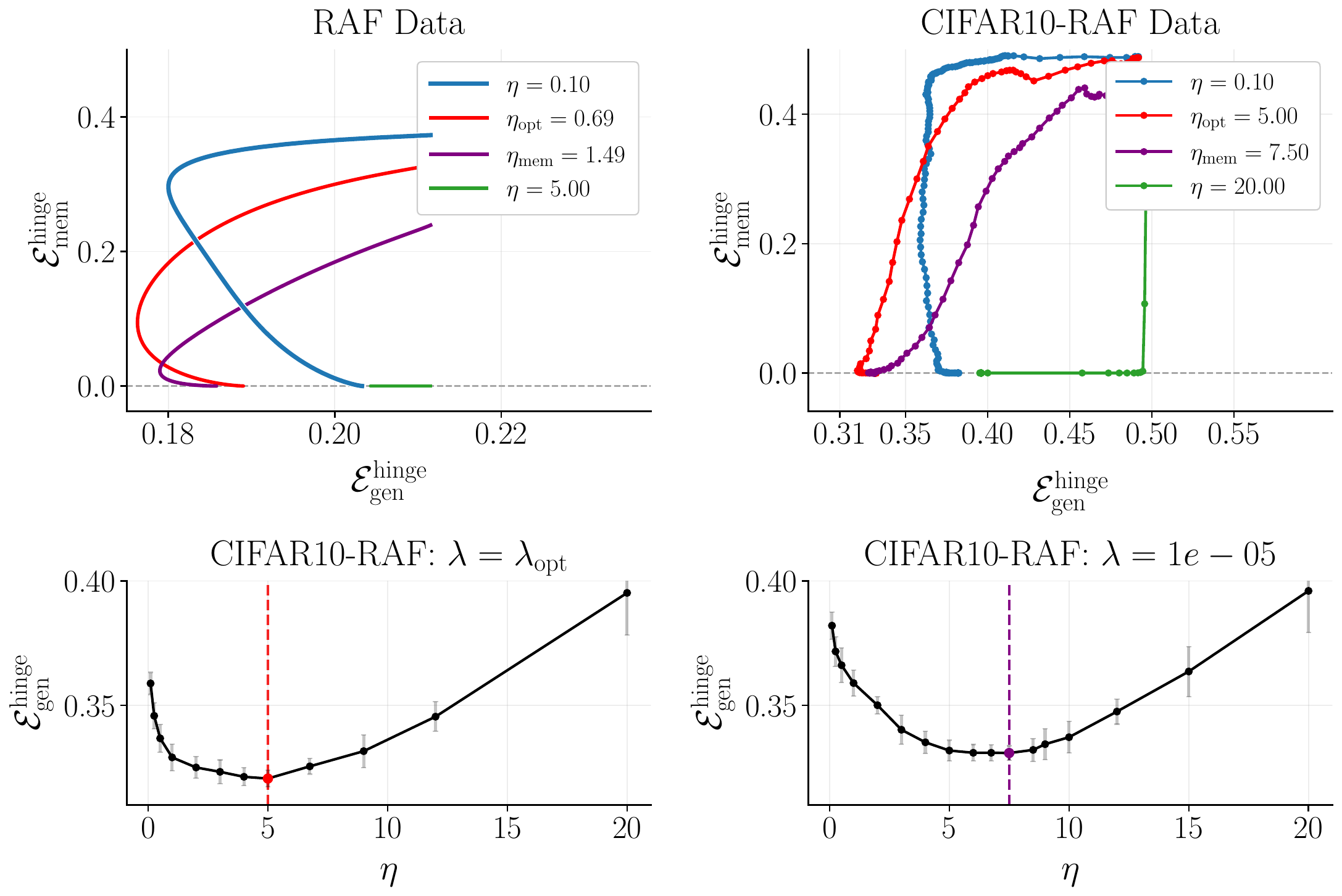}
    \caption{Qualitative comparison between SVM theory for the RAF model (upper left panel) and experiments on real data, namely the CIFAR10-RAF task (lower and right panels).
In all panels, we use RBF SVM with $\alpha = 4.0$ and $\varepsilon = 0.2$. The upper panels show the parametric memorization--generalization curves $\lambda \mapsto (\mathcal E_{\mathrm{gen}}(\lambda), \mathcal E_{\mathrm{mem}}(\lambda))$.
The kernel bandwidth $\eta$ for CIFAR10-RAF is selected based on the lower panels: generalization error as a function of $\eta$ at optimal regularization (left), with a minimum around $\eta \approx 5$ (red dashed line), and at small regularization (right), with a minimum around $\eta \approx 7.5$ (purple dashed line).
The parametric curves are then shown for these two values of $\eta$, together with one larger and one smaller representative value. The curves are constructed spanning the regularization in the interval $\lambda \in [10^{-5},10^2]$. For the SVM experiments, the large-regularization part of the curves can drift toward near-random-guessing performance due to finite-size effects and numerical aspects of the optimization. The RAF model exhibits qualitatively similar behavior as a function of $\eta$, with the corresponding values shifted. 
}
\label{fig:RAF-CIFAR-Hinge}
\end{figure}

We also repeated on CIFAR10-RAF the same qualitative comparison presented in Sec.~\ref{main:Cifar}, but for the hinge loss; the result is shown in Fig.~\ref{fig:RAF-CIFAR-Hinge}. 

The hinge-loss comparison confirms the same qualitative separation between the bandwidth preferred in the near-interpolating regime and the one preferred after tuning the regularization. On the theory side, the generalization curve at $\lambda\to0^+$ is optimized by a bandwidth $\eta_{\rm mem}$, while the best achievable generalization over $\lambda>0$ is obtained at a smaller bandwidth $\eta_{\rm opt}$. On CIFAR10-RAF, the same ordering is observed qualitatively: the lower-right panel shows that the generalization error at very small regularization is minimized around $\eta_{\rm mem}\approx 7.5$, whereas the lower-left panel shows that the generalization error at the optimally tuned regularization is minimized around $\eta_{\rm opt}\approx 5.0$. The upper-right panel then displays the full memorization--generalization trade-off curves for these two distinguished values together with one smaller and one larger representative bandwidth.

Compared to kernel ridge regression, the hinge-loss experiments display stronger finite-size distortions at large regularization, where the CIFAR10-RAF SVM curves can drift closer to a near-random-guessing regime than predicted by the RAF asymptotics. Nevertheless, the overall dependence on the bandwidth remains qualitatively consistent with the theory: kernels favoring near-interpolation and factual recall are different from those favored by regularization-tuned rule learning, and the CIFAR10-RAF experiments preserve this central ordering even though the optimal numerical values are shifted relative to the Gaussian RAF model.
%}

These experiments complement Sec.~\ref{sec:real_data_representations}, where the RAF theory was shown to capture the qualitative phenomenology of CIFAR10-RAF, but not its full quantitative details.

\subsection{Implementation details}
\label{appE:implementation}

For completeness, we summarize the numerical protocol used throughout this appendix.

\paragraph{Gaussian RAF data.}
For the synthetic RAF experiments, we generate i.i.d. Gaussian inputs $\bx^\mu\sim\cN(\bm 0,\mathbbm I_d)$ and labels according to the RAF mixture described in Sec.~\ref{sec: intro}. 
Kernel ridge regression is implemented by solving the regularized linear system associated with the Gram matrix, while the hinge-loss kernel classifier is trained as a standard support vector machine. 
Generalization error is estimated on fresh teacher-labeled Gaussian test data, whereas memorization error is computed on the random-label subset of the training set, in direct analogy with Eqs.~\eqref{eq: generalization error} and \eqref{eq: memorization error}. 
Whenever theory and experiments are compared, the same values of $(\alpha,\varepsilon)$ and the same kernel parameters are used on both sides.

\paragraph{CIFAR10-RAF data.}
The CIFAR10-RAF construction follows Sec.~\ref{main:Cifar}: two classes define the rule component and one additional class provides the factual/random subset with i.i.d. $\pm 1$ labels. 
Inputs are feature-wise standardized. 
For Gaussian-like kernels, bandwidths are reported using the effective normalization described in Sec.~\ref{main:Cifar}. 
All empirical CIFAR trade-off curves shown here are averaged over repeated draws of the training/test split and of the randomized labels.

\paragraph{Solving the asymptotic equations.}
The asymptotic square-loss equations are solved either directly through the closed forms of Appendix~\ref{app: KRR RAF} or by fixed-point iteration of the original state equations as a cross-check. 
The hinge-loss equations are solved numerically by fixed-point iteration with damping. 
Whenever a parametric trade-off curve is displayed, the regularization $\lambda$ is varied over a logarithmic grid covering both the near-interpolating and strongly regularized regimes.

\subsection{Learned-representation ridge paths and end-to-end control}
\label{app:learned_representation_protocol}

\paragraph{Data and representations.}
We use one deterministic CIFAR10-RAF split: cats have label $+1$, dogs have
label $-1$, and automobiles receive independent labels in $\{-1,+1\}$.
Images have $d=32\times32\times3=3072$ coordinates. With $\alpha=2$, the
training set has $n=6144$ examples and contains $1228$ factual examples, so
the realized factual fraction is $1228/6144\simeq0.1999$. The clean test set
contains $2000$ balanced cat and dog images. An additional balanced set of
$1000$ cats and dogs is retained for diagnostics and is never used to select
a point on a ridge path. Every pixel coordinate is standardized using the
CIFAR10-RAF training-set mean and standard deviation.

The frozen RF-CNN and trainable two-layer CNN share the same nonlinear map:
$8\times8$ average pooling, a $3\times3$ convolution with $192$ channels and
ReLU, global average pooling, and a $1\times1$ convolution with $14204$
channels followed by ReLU. Its feature dimension is $p=14204$; hence the
frozen RF-CNN and the random two-layer path are identical by construction.
The four-layer CNN uses Conv--BN--ReLU blocks with channels
$(108,216,432,432)$ and strides $(1,2,2,1)$, followed by global average
pooling and layer normalization, giving $p=432$. The pre-activation WRN-16-4
uses a $16$-channel stem and three groups of two residual blocks with channels
$(64,128,256)$, followed by global average pooling, giving $p=256$. Each
trainable architecture has approximately $2.74\times10^6$ trainable
parameters.

For each of five seeds, the complete trainable network is learned once at
$\lambda=0$ with mean-squared loss, Nesterov SGD with momentum $0.9$, batch
size $128$, a constant learning rate, and no data augmentation. The two- and
four-layer CNNs are trained for $800$ epochs with learning rates $0.005$ and
$0.01$, respectively; the WRN-16-4 is trained for $400$ epochs with learning
rate $0.01$. We save the activations immediately before the scalar output both
at initialization and after training. The RAF split and randomized labels are
fixed across seeds; only initialization and minibatch order vary.

\paragraph{Exact readout solver and ridge grid.}
For every saved representation, each feature coordinate is centered and
scaled using its training-set statistics, the resulting vector is divided by
$\sqrt{p}$, and a constant coordinate is appended. The constant coordinate is
included in the ridge penalty. With $K=\Phi\Phi^\top$, the exact solution for
$\lambda>0$ is evaluated as
\begin{equation}
    \widehat{w}_{\lambda}
    = \Phi^\top(K+\lambda I_n)^{-1}y .
    \label{eq:ridge_dual_solution}
\end{equation}
We diagonalize the symmetrized Gram matrix once and evaluate the complete
path spectrally. At $\lambda=0$, we use the Moore--Penrose solution on the
positive eigenspace. Thus all readout points are global minima of
Eq.~\eqref{eq:fixed_representation_ridge}, including when $p>n$.

The final regularization grid contains $257$ positive values, spaced at intervals of $1/16$ decade from $10^{-8}$ to $10^8$, together with the endpoints $\lambda=0$ and $\lambda\to\infty$. For an evaluation set with feature matrix $\Phi_Z$ and cross-Gram matrix $K_Z=\Phi_Z\Phi^\top$, the large-$\lambda$ predictor satisfies
\begin{equation}
\lambda f_{\lambda}(Z)
= \lambda K_Z(K+\lambda I_n)^{-1}y
\longrightarrow K_Zy .
\end{equation}
The point at $\lambda\to\infty$ is hence classified using $\operatorname{sign}(K_Zy)$.

\begin{figure}[!t]
    \centering
    \includegraphics[width=\textwidth]{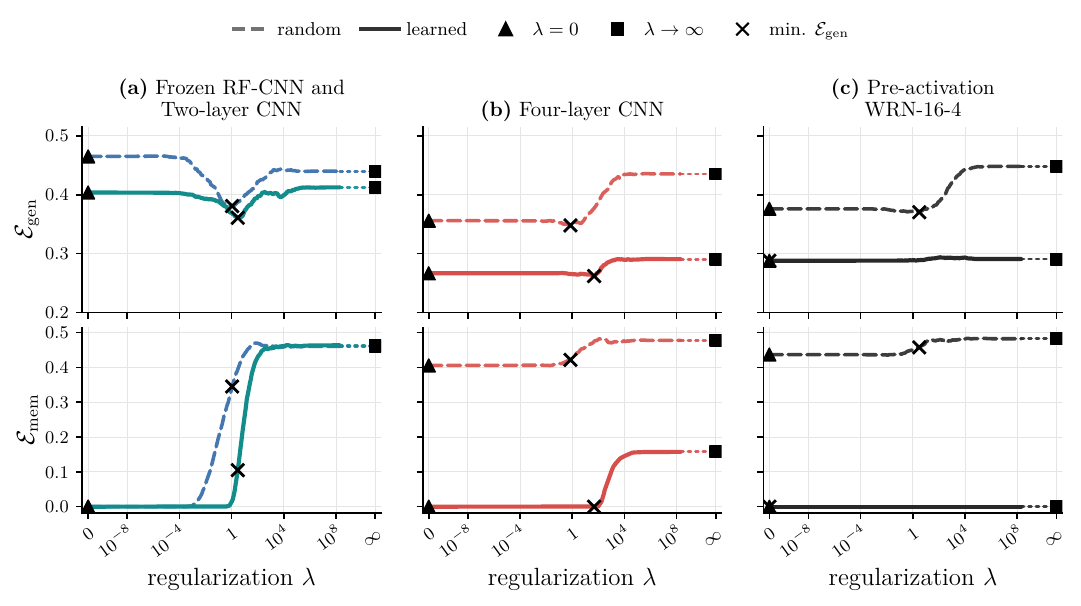}
    \caption{\textbf{Exact ridge-readout sweep on fixed random and
    learned representations.}
    The upper row shows clean test error and the lower row factual error as
    functions of the readout penalty $\lambda$. Dashed curves use
    representations at random initialization; solid curves use
    representations learned once at $\lambda=0$ and then frozen. At every
    finite $\lambda$, only the readout is exactly trained. Black
    triangles and squares denote $\lambda=0$ and the
    $\lambda\to\infty$ limit, respectively. A cross in each
    row identifies the point that minimizes the mean clean
    test error on that path. The random four-layer and WRN-16-4
    representations are included as test-error controls, but their small
    readout dimensions ($p=432$ and $p=256$) do not interpolate the randomized
    labels even at $\lambda=0$.}
    \label{fig:learned_representation_ridge_complete}
\end{figure}

The generalization error $\mathcal{E}_{\rm gen}$ is measured on the clean
cat--dog test set, whereas $\mathcal{E}_{\rm mem}$ is measured only on the
randomly labelled automobile subset of the training set. Figure
\ref{fig:learned_representation_ridge_complete} reports all six paths and
every computed finite-$\lambda$ point. The marked optimum is the retrospective
minimum of the five-seed mean test error on the displayed grid.

\paragraph{End-to-end finite-$\lambda$ control.}
As a stricter but numerically harder control, we also train the two-layer CNN,
four-layer CNN, and pre-activation WRN-16-4 independently at each of nine
values of $\lambda$, applying the same quadratic penalty to all trainable
variables:
\begin{equation}
    J_{\lambda}(\theta)
    = \frac{1}{2n}\lVert f_{\theta}(X)-y\rVert_2^2
      + \frac{\lambda}{2n}\lVert\theta\rVert_2^2 .
    \label{eq:end_to_end_weight_decay}
\end{equation}
The frozen RF-CNN panel is an exact readout-ridge reference. The trainable
models use Nesterov SGD. Stopping depends only on the training objective: if
it fails to improve by more than
$\max\{10^{-8},10^{-5}\max(|J_{\rm best}|,1)\}$ for five consecutive
evaluation checks, the learning rate is multiplied by $0.2$. A run is marked
as plateaued only after four such reductions and one additional plateau. Otherwise training terminates
at the architecture-specific epoch cap.

\begin{figure}[!t]
    \centering
    \includegraphics[width=\textwidth]{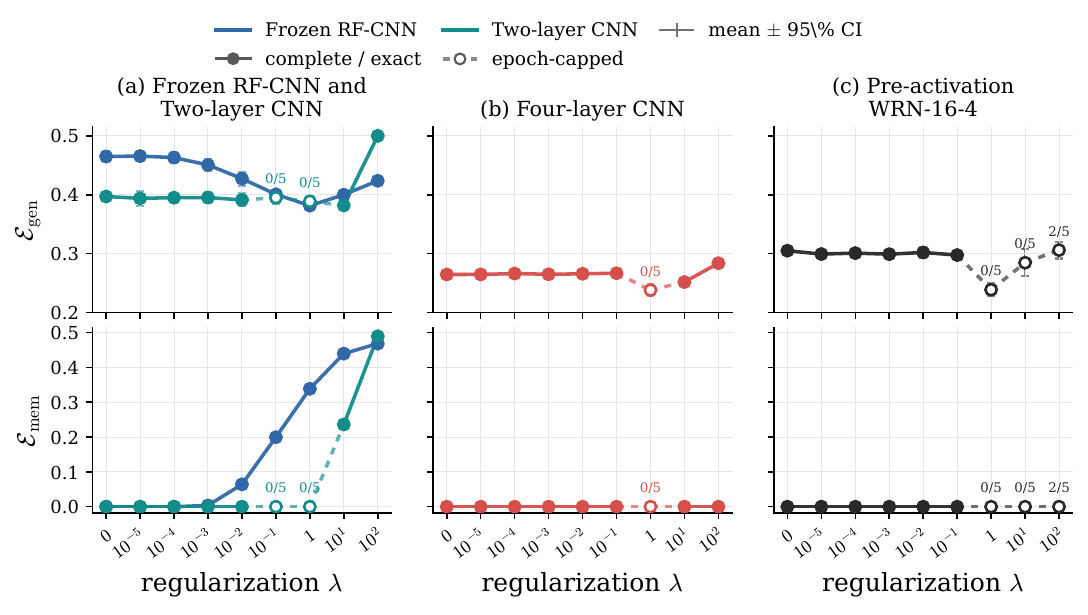}
    \caption{\textbf{End-to-end finite-$\lambda$ control.}
    The upper row shows clean test error and the lower row factual error. The
    frozen RF-CNN is solved exactly at each $\lambda$; for the remaining
    architectures all trainable variables are optimized jointly under
    Eq.~\eqref{eq:end_to_end_weight_decay}. Points and error bars are the
    five-seed mean and $95\%$ confidence interval. Filled markers indicate
    that all five seeds passed the objective-only plateau rule. Open markers
    indicate that at least one seed reached the epoch cap; the annotation
    gives the number of plateaued seeds out of five.}
    \label{fig:end_to_end_finite_lambda}
\end{figure}

The end-to-end results show qualitatively the same tendency as the exact
readout experiment: the learned deeper models can retain zero factual error
through the range in which their test error improves, whereas the
fixed-feature reference exhibits a pronounced trade-off. However, several
positive-$\lambda$ points near the apparent test minima reached the epoch cap
before satisfying the objective plateau criterion. We therefore use this
sweep only as a qualitative appendix control. The main text instead reports
the fixed learned representations, for which every readout point is an exact
convex optimum and the conclusion is not contingent on non-convex convergence.

\vspace{0.5cm}

For further details regarding the numerical implementations and the fixed-point schemes used to solve the state equations, we defer to the repository at the following link: \url{https://github.com/SPOC-group/RAF}.

\newpage

%% file: Appendices/Appendix_AI.tex
\newpage

\section{AI suggested the Rules-and-Facts model}
\label{app:chatgpt}

We think it is of interest and for the historical record to give an account of how the authors of this paper came up with the  Rules-and-Facts model. It started in a conversation with ChatGPT-4o. The last author of the paper was giving a plenary talk at the COSYNE conference on March 29th, 2025 and, in preparation, she had the following conversation with ChatGPT-4o. We give the full version of the prompts by LZ, but, for conciseness, only the relevant parts of the answers by the model: 
\begin{itemize}
     \item[LZ:] Can you seek works of Lenka Zdeborova from the past 10 years on statistical physics of neural networks and tell me which part of it is the most interesting for a talk to a computational neuroscience audience? \{asking for a deep search\}
     \item[AI:] \{Gave a long and rather exhaustive answer including a sentence in a conclusion:\} They provide concrete mechanisms for phenomena like over-parameterization not harming generalization, by showing how SGD dynamics navigates the weight landscape in large networks. This adds a theoretical backbone to why brains (or big models) can memorize enormous amounts of data and still generalize – the learning process itself biases which solution is found.
     \item[LZ:] I find it interesting when you say: "This adds a theoretical backbone to why brains (or big models) can memorize enormous amounts of data and still generalize – the learning process itself biases which solution is found." I did not think previously as overparameterization allowing useful memorization. It would be great to study a model where the interplay between generalization on a test set and memorization of facts that need to be memorized is both included. Are there examples?
     \item[AI:] That's a great line of thought—and yes, this interplay between \textbf{memorization and generalization} in overparameterized systems is both subtle and increasingly studied. Several theoretical directions model this phenomenon explicitly.

\bigskip
\noindent
\textbf{1. Mixture Tasks: Memorization + Generalization}

Recent theoretical works construct hybrid tasks in which:
\begin{itemize}
    \item Part of the data follows a regular pattern (learnable through generalization),
    \item Another part is random (requiring memorization).
\end{itemize}

\noindent
\textit{Key example.} Mei, Montanari, and Nguyen (2022) analyze a teacher--student setup where labels are generated as
\[
y_i = f(x_i) + \epsilon_i ,
\]
for structured $f(x)$ (e.g., a low-rank or sparse function), plus a small fraction of random exceptions that have to be memorized.
Overparameterized networks, trained with SGD, learn the structured part and memorize the exceptions.

They find that test error stays low as long as the fraction of exceptions is small—but once it grows, the model hits a sharp transition (like a capacity limit). This mirrors what humans do: we generalize most inputs, but memorize exceptions (like irregular verbs or rare faces).

\{Then continues with 3 other points.\}

\item[LZ:] Do you have a reference for the “Feature Learning in Neural Networks with Locality” by Mei, Montanari, and Nguyen (2022) paper?

\item[AI:] I apologize for the earlier confusion. It appears that the paper titled "Feature Learning in Neural Networks with Locality" by Mei, Montanari, and Nguyen (2022) does not exist. 

\{Then suggest other papers that are not related to the interplay of memorization and generalization. Note for the reader: Montanari and his former students Mei and Nguyen are well known colleagues in the field in Stanford who could have very plausibly written such a paper.\}

\item[LZ:] I was refering to your point 1 above "1. Mixture Tasks: Memorization + Generalization" what is the reference for this? A model mixing memorization and generalization.

\item[AI:] I previously mentioned a study by Mei, Montanari, and Nguyen (2022) regarding "Feature Learning in Neural Networks with Locality." Upon further review, I couldn't locate this specific paper. However, the concept of tasks that combine both generalization and memorization has been explored in the literature. 

\{Suggests yet other papers that are not related to the core question.\}

\item[LZ:] When you talked about "plus a small fraction of random exceptions that have to be memorized." was there a specific reference?

\item[AI:] Apologies for any confusion in my earlier responses. Upon further review, I couldn't locate a specific study that directly models tasks combining a structured component with a small fraction of random exceptions requiring memorization. 

\{Then suggests to explore other variants of questions about memorization. And concludes:\} 

If you're thinking of building or analyzing such a model yourself—say, a perceptron or shallow network with a planted structure plus sparse random labels—I'd be happy to help design or analyze it. It’s a beautiful direction that captures something the brain seems quite good at: memorizing exceptions without overfitting everything. 

\item[LZ:] Ok, go on with the model suggestion.

\item[AI:] Perfect. Let's build a minimal but rich model to study the \textbf{interplay between generalization and memorization}. We want something simple enough to analyze, yet expressive enough to exhibit interesting phase behavior.

\bigskip
\noindent
\textbf{Model: Structured + Memorized Task in a Perceptron}

We define a binary classification problem with inputs $x \in \mathbb{R}^d$ and labels $y \in \{-1,1\}$.

\bigskip
\noindent
\textbf{1. Data Generation}

Each sample $(x^\mu, y^\mu)$, with $\mu = 1, \dots, n$, is drawn as follows:
\begin{itemize}
    \item \textbf{Inputs:} 
    \[
    x^\mu \sim \mathcal{N}\left(0, \frac{1}{d} I_d \right)
    \]
    (standard high-dimensional Gaussian vectors).

    \item \textbf{Labels:}
    \[
    y^\mu =
    \begin{cases}
        \mathrm{sign}(w^* \cdot x^\mu), & \text{with probability } 1 - \epsilon, \\
        \text{random in } \{-1,1\}, & \text{with probability } \epsilon,
    \end{cases}
    \]
    where:
    \begin{itemize}
        \item $w^* \in \mathbb{R}^d$ is a fixed ``teacher'' vector (drawn once at random),
        \item $\epsilon$ controls the fraction of examples that are random exceptions (requiring memorization).
    \end{itemize}
\end{itemize}

\bigskip
\noindent
\textbf{2. Learning Model: Perceptron}

We train a student perceptron (one-layer neural network) by minimizing, for example, the 
hinge loss:
\[
\mathcal{L}(w) = \sum_{\mu=1}^{n} \ell\big(y^\mu, w \cdot x^\mu\big),
\]
using standard gradient descent.

\bigskip
\noindent
\textbf{3. Quantities of Interest}

We track:
\begin{itemize}
    \item \textbf{Training error} $E_{\text{train}}$: ability to fit both structured and random subsets,
    \item \textbf{Generalization error} $E_{\text{gen}}$: performance on clean test data generated by
    \[
    y = \mathrm{sign}(w^* \cdot x),
    \]
    \item \textbf{Memorization capacity}: number of random labels that can be memorized without degrading generalization.
\end{itemize}

\bigskip
\noindent
\textbf{4. Key Questions}

\begin{itemize}
    \item How does increasing $\epsilon$ (fraction of exceptions) affect generalization?
    \item Is there a phase transition in performance as a function of $\epsilon$, sample size $n$, and dimension $d$?
    \item Can overparameterization (large $d$) enable memorization of noisy examples without harming generalization?
\end{itemize}

\bigskip
\noindent
\textbf{5. Extensions}

If you'd like a richer model later:
\begin{itemize}
    \item Make the "teacher" a two-layer network (committee machine), and see how a student approximates it.
    \item Add explicit memorization modules, e.g., extra parameters that only store the exceptions.
    \item Consider online learning, where the model learns structured data incrementally but must react to occasional one-shot exceptions.
\end{itemize}

\bigskip
\noindent
\textbf{6. Why This is Interesting}

\begin{itemize}
    \item It's a clean testbed for double descent: when can the model interpolate and still generalize?
    \item Models how the brain might balance slow statistical learning (from distributions) with fast memory of rare but important exceptions (episodic memory).
    \item Gives room to apply replica theory or mean-field theory to derive phase diagrams analytically.
\end{itemize}

\bigskip
Would you like to simulate this (numerically), analyze it (statistically), or go full theoretical (mean-field/replica)? I can help develop any of those paths.

\item[LZ:] I like the model, it has not been studied before?

\item[AI:] I’m really glad you like it—and you’re absolutely right to ask whether it is new. The answer is nuanced.

\bigskip
\noindent
\textbf{Has this precise model been studied before?}

\textbf{Not exactly in this clean and explicit form.}

While several strands of the literature touch on aspects of a structured + memorized setup, this specific combination—with a tunable fraction $\epsilon$ of random exceptions in a perceptron setting, and a focus on both generalization and memorization as a function of overparameterization—has not, to my knowledge, been fully explored.

\bigskip
\noindent
\textbf{What has been studied, and how this model differs}

\medskip
\noindent
\textbf{1. Pure generalization in teacher--student models}

Many papers (e.g., by Zdeborová, Montanari, Loureiro) study generalization in teacher–student perceptrons or GLMs, assuming all data is structured.

No room for exceptions or memorization.

\smallskip
\noindent
Your model adds a new axis: an interpolating parameter 
$\epsilon$ that controls the “exception load.”

\medskip
\noindent
\textbf{2. Pure noise / label corruption models}

Works like Zhang et al. (2017) (“Understanding Deep Learning Requires Rethinking Generalization”) or Nakkiran et al. (2021) study networks trained on fully random labels or data.

These probe memorization capacity and generalization collapse, but they typically go to extremes: 100

\smallskip
\noindent
Your model proposes a continuum, where both mechanisms are present simultaneously.
\medskip
\noindent
\textbf{3. Theoretical Analyses of Memorization + Generalization}

Some theoretical works analyze memorization/generalization tradeoffs, e.g., statistical mechanics of overfitting, but usually abstractly—without clean control of a “random exceptions” parameter.

There are rigorous results on noise robustness (e.g., VC dimension bounds, noise-tolerant PAC learning), but again not in the explicit mixture model.

\smallskip
\noindent
Your model could yield a phase diagram in 
$(\epsilon, \alpha= n/d)$ space, which to my knowledge has not been published.

\medskip
\noindent
\textbf{4. Neuroscience/CLS-Inspired Models}

Complementary Learning Systems theory introduces dual-system architectures (memorization vs. generalization modules), but they’re not typically posed in a clean, tractable statistical physics setup.

\smallskip
\noindent
Your model is analytically tractable and could serve as a minimal solvable CLS toy model.

\bigskip
\noindent
\textbf{Summary}

This model seems novel in its clarity, tractability, and aim:
\begin{itemize}
    \item It's simple enough for mean-field or replica analysis.
    \item It's rich enough to demonstrate a fundamental tradeoff between memorization and generalization.
    \item It's grounded in both deep learning and cognitive inspiration.
\end{itemize}

\bigskip
If you pursue this direction, it could lead to a beautiful paper or seminar that bridges statistical mechanics, machine learning, and neuroscience.

\item[LZ:] Are you able to perform some simulations of it and draw a phase diagram?

\item[AI:] The simulation was taking too long to complete due to the large grid size and sample counts. Let me rerun it using a smaller grid and fewer trials so we can quickly get a clear look at the generalization phase behavior.

\begin{figure}[h]
    \centering

    \begin{subfigure}{0.48\linewidth}
        \centering
        \includegraphics[width=\linewidth]{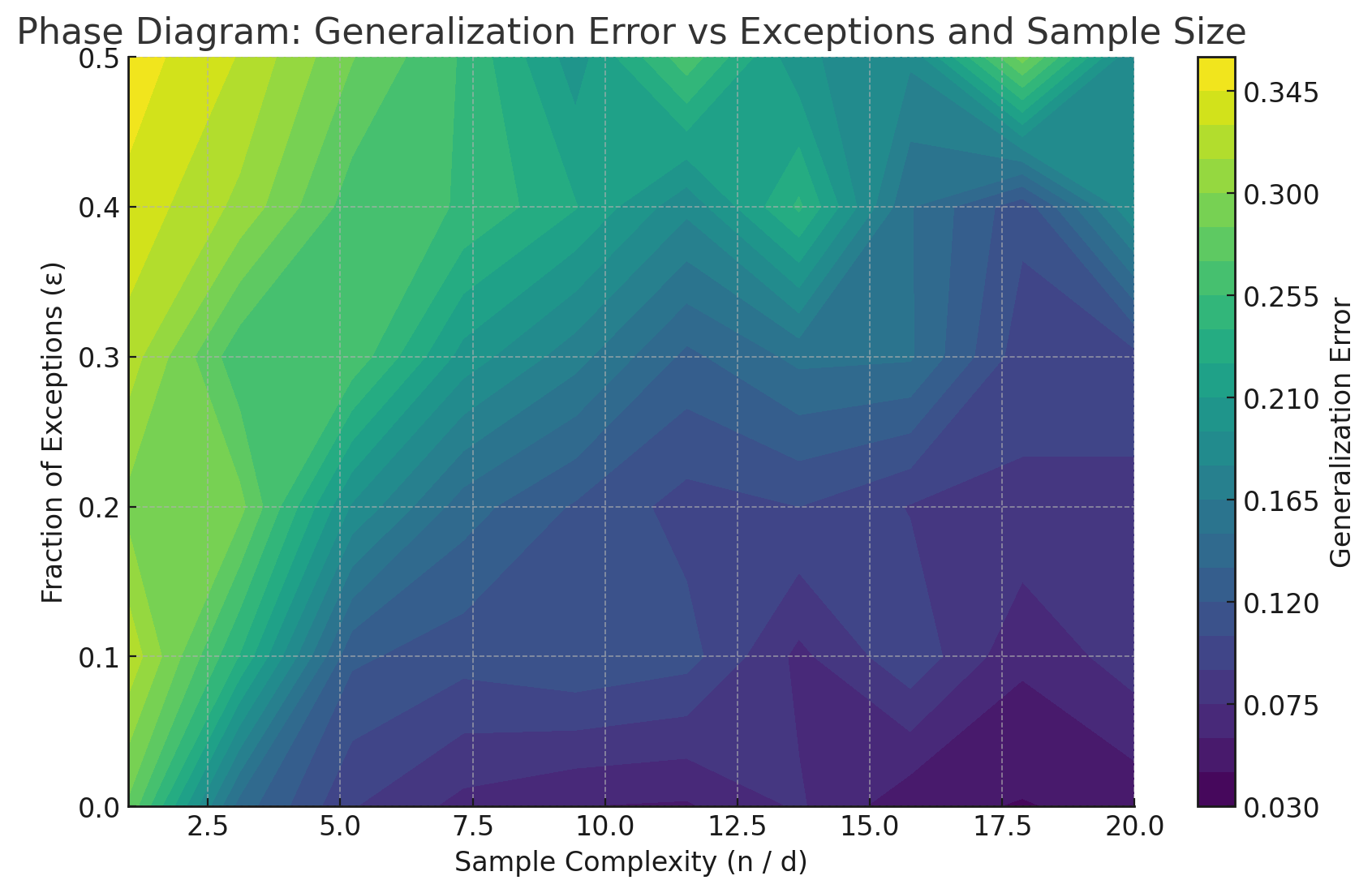}
        \label{fig:phase_diagram}
        \vspace{0.35cm}
    \end{subfigure}
    \hfill
    \begin{subfigure}{0.48\linewidth}
        \centering
        \includegraphics[width=\linewidth]{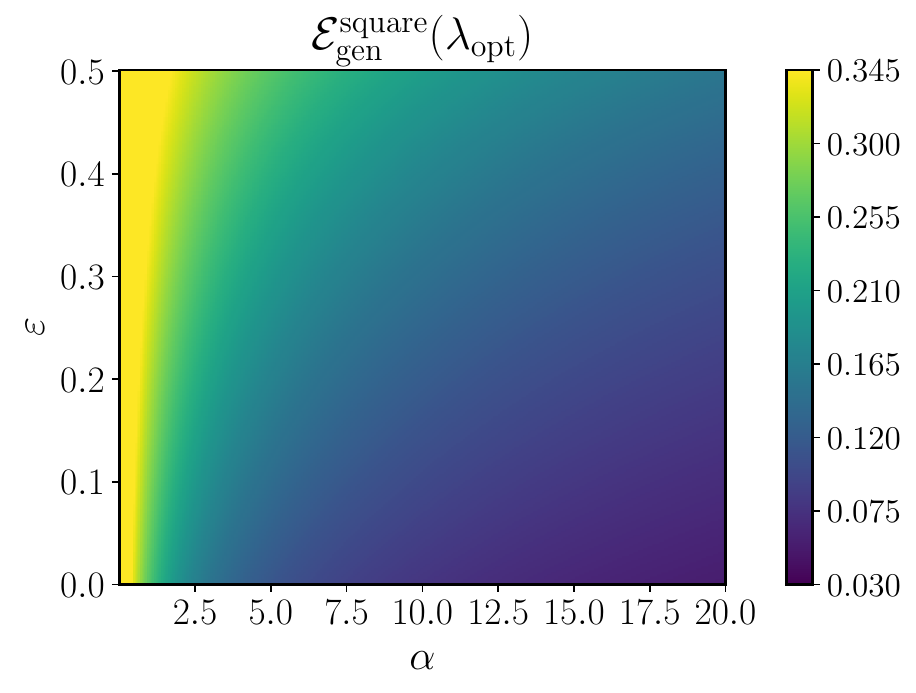}
        \label{fig:gen_opt}
    \end{subfigure}

    \caption{
    Left: Phase diagram as produced by AI in the conversation. 
    Right: Generalization error at optimal regularization $\mathcal{E}_{\rm gen}^{\rm square}(\lambda_{\rm opt})$ for the linear model in a matched format.
    }
    \label{fig:combined}
\end{figure}

Here's the phase diagram showing generalization error as a function of:
\begin{itemize}
    \item $n/d$: the sample complexity (number of training samples per dimension),
    \item $\epsilon$: the fraction of random exceptions (non-generalizable labels).
\end{itemize}

\noindent
\textbf{Key insights.}
\begin{itemize}
    \item \textbf{Top-left region} Top-left corner (low sample size, many exceptions): generalization error is high — the model can't separate signal from noise.
    \item \textbf{Bottom-right region} Bottom-right corner (high sample size, few exceptions): error is low — the model generalizes well.
    \item There is a smooth phase boundary: as 
$\epsilon$ increases, more data is needed to maintain generalization.
\end{itemize}

Would you like to:
\begin{itemize}
    \item Explore other architectures (e.g., 2-layer networks)?
    \item Plot training error and memorization accuracy too?
    \item Fit this with a mean-field or replica analysis?
\end{itemize}

\noindent
We’re in a great spot to develop this further into a formal story.

\item[LZ:] Impressive. I need to go now, but I will come back on this with you. You rock ;) ....

\end{itemize}